\RequirePackage{fix-cm}
\documentclass[twocolumn]{svjour3}          
\smartqed  

\usepackage{graphicx}
\usepackage[numbers,sort&compress]{natbib}
\usepackage{xcolor}
\usepackage{hyperref}
\hypersetup{
    colorlinks,
    linkcolor={black!50!black},
    citecolor={blue!50!black},
    urlcolor={blue!80!black}
}

\usepackage{lineno}
\usepackage{soul}
\usepackage{lscape}
\usepackage{booktabs}
\usepackage{amsmath,mathtools}
\usepackage[intoc]{nomencl}
\usepackage{url}
\usepackage{doi}
\usepackage[boxruled,vlined,linesnumbered]{algorithm2e}
\usepackage{amsfonts}
\usepackage{relsize}
\usepackage{commath}
\usepackage{float}
\usepackage{textcomp}
\usepackage{array}
\usepackage[nottoc,notlot,notlof]{tocbibind}
\usepackage{setspace}
\usepackage{mathptmx}

\usepackage{subcaption}
\graphicspath{{plots/}}

\newcommand{\Bm}{\mathbf{m}}
\newcommand{\Bg}{\mathbf{g}}
\newcommand{\Bf}{\mathbf{f}}
\newcommand{\Bd}{\mathbf{d}}
\newcommand{\Bdhat}{\tilde{\mathbf{d}}}

\newcommand{\Na}{N_\text{a}}
\newcommand{\Nhm}{N_\text{hm}}
\newcommand{\Npred}{N_\text{pred}}

\newcommand{\Nf}{N_\text{f}}
\newcommand{\R}{\mathbb{R}}
\newcommand{\Nqoi}{N_\text{QoI}}
\newcommand{\Nt}{N_\text{t}}
\newcommand{\Bdpred}{\mathbf{d}_\text{pred}}
\newcommand{\Bdobs}{\mathbf{d}_\text{obs}}
\newcommand{\Bepsilon}{\boldsymbol{\epsilon}}
\newcommand{\Bxi}{\boldsymbol{\xi}}
\newcommand{\Bxipca}{\boldsymbol{\xi}^{\text{PCA}}}
\newcommand{\Bxirae}{\boldsymbol{\xi}^{\text{RAE}}}
\newcommand{\CD}{C_{\text{D}}}

\newcommand{\Cpd}{C_{\text{pd}}}

\newcommand{\Bdhm}{\mathbf{d}_\text{hm}}
\newcommand{\Bdpca}{\mathbf{d}^\text{PCA}}
\newcommand{\Bdprior}{\bar{\mathbf{d}}_\text{prior}}

\newcommand{\Nr}{N_\text{r}}
\newcommand{\Nl}{N_\text{l}}
\newcommand{\Nd}{N_\text{d}}
\newcommand{\Nh}{N_\text{h}}

\newcommand{\Bit}{\mathbf{i}_t}
\newcommand{\Bft}{\mathbf{f}_t}
\newcommand{\Bot}{\mathbf{o}_t}

\newcommand{\Bx}{\mathbf{x}}
\newcommand{\Bh}{\mathbf{h}}
\newcommand{\Bb}{\mathbf{b}}

\newcommand{\Bc}{\mathbf{c}}
\newcommand{\Bchat}{\Tilde{\mathbf{c}}}

\begin{document}

\title{Data-Space Inversion Using a Recurrent Autoencoder for Time-Series Parameterization}
\author{Su Jiang \and Louis J. Durlofsky}

\institute{Su Jiang \at
              Department of Energy Resources Engineering, \\
              Stanford University, Stanford CA 94305, USA \\
              \email{sujiang@stanford.edu}           
           \and
           Louis J. Durlofsky \at
              Department of Energy Resources Engineering, \\
              Stanford University, Stanford CA 94305, USA \\
              \email{lou@stanford.edu}
}

\date{Received: date / Accepted: date}
\maketitle

\begin{abstract}
Data-space inversion (DSI) and related procedures represent a family of methods applicable for data assimilation in subsurface flow settings. These methods differ from usual model-based techniques in that they provide only posterior predictions for quantities (time series) of interest, not posterior models with calibrated parameters. DSI methods require a large number ($O(500-1000)$) of flow simulations to first be performed on prior geological realizations. Given observed data, posterior predictions for time series of interest, such as well injection or production rates, can then be generated directly. DSI operates in a Bayesian setting and provides posterior samples of the data vector. In this work we develop and evaluate a new approach for data parameterization in DSI. Parameterization is useful in DSI as it reduces the number of variables to determine in the inversion, and it maintains the physical character of the data variables. The new parameterization uses a recurrent autoencoder (RAE) for dimension reduction, and a long-short-term memory (LSTM) recurrent neural network architecture to represent flow-rate time series. The RAE-based parameterization is combined with an ensemble smoother with multiple data assimilation (ESMDA) for posterior data sample generation. Results are presented for two- and three-phase flow in a 2D channelized system and a 3D multi-Gaussian model. The new DSI RAE procedure, along with several existing DSI treatments, are assessed through detailed comparison to reference rejection sampling (RS) results. The new DSI methodology is shown to consistently outperform existing approaches, in terms of statistical (P$_{10}$-P$_{90}$ interval and Mahalanobis distance) agreement with RS results. The method is also shown to accurately capture derived quantities, which are computed from variables considered directly in DSI. This requires correlation and covariance between variables to be properly captured, and accuracy in these relationships is demonstrated. The RAE-based parameterization developed here is clearly useful in DSI, and it may also find application in other subsurface flow problems.
\keywords{Data assimilation \and History matching \and Reservoir simulation \and Data-space inversion \and Time-series parameterization \and Deep-learning}
\end{abstract}

\section{Introduction}
Model calibration is typically required before subsurface flow simulation tools can be used for prediction and optimization. This is usually accomplished through application of formal inversion procedures in which key flow parameters, such as porosity and permeability in all grid blocks in the model, are determined based on observations and a prior concept of the geological scenario. With large and geologically complex models, however, this inversion can be challenging. This is because many flow simulations may be required, which can be expensive for highly refined models, and because it can be difficult to retain geological realism in the posterior (calibrated) models.

A complementary set of procedures, referred to as data-space inversion (DSI) or prediction-focused methods, circumvents some of the challenges associated with traditional model-space approaches. These methods provide selected posterior flow information, such as time-varying flow rates at wells, using only prior flow simulation results and observed data. Data-space approaches have several advantages relative to model-based methods, including the ability to treat prior realizations drawn from different scenarios, or models defined on different simulation grids. They are also able to assimilate new observed data, or vary data-error or model-error parameters, without performing additional flow simulations. These methods also have disadvantages relative to traditional model-based procedures, most notably the fact that they provide only posterior predictions for a particular set of flow quantities, not posterior models. Thus model-space and data-space techniques would be used to address different engineering questions. Our goal in this work is to improve the quality of DSI predictions through the incorporation of a new deep-learning-based data parameterization method, referred to as a recurrent autoencoder (RAE).

The literature on model-space data assimilation (model calibration) in the context of subsurface flow is very extensive. Our discussion here will focus on methods that apply parameterization, since an emphasis of this work is on the use of RAE-based data parameterization in DSI. Model parameterization enables the properties of interest (e.g., permeability values in all $N_b$ grid blocks in the model) to be represented in terms of a low-dimensional variable of dimension $N_l$, with $N_l<<N_b$. This simplifies the inversion procedure since many fewer parameters must be determined, and geological realism is retained, at least to some degree, through the parameterization. Principal component analysis (PCA) methods, directly applicable for multi-Gaussian models, were originally applied by Oliver~\cite{oliver1996multiple}. More general PCA-based methods include kernel PCA~\cite{sarma2008kernel}, optimi\-zation-based PCA~\cite{vo2014new}, grouping with sparse PCA~\cite{golmohammadi2015group}, and PCA with distance transform~\cite{hakim2017distance}. A tensor-based higher-order singular value decomposition procedure, which can also be viewed as a generalization of PCA, was applied by Afra and Gildin~\cite{afra2016tensor}. Recently, a method involving the use of convolutional neural networks with PCA (CNN-PCA) was developed for parameterizing non-Gaussian systems, such as channelized models~\cite{liu2019deep}. A number of parameterizations involving deep-learning-based autoencoder procedures have also been introduced~\cite{canchumuni2019history, canchumuni2019towards, laloy2017inversion, mo2020integration}. Finally, generative adversarial networks (GANs) combined with CNNs have been applied to parameterize geological media~\cite{chan2019parametric, mosser2017reconstruction, laloy2018training, laloy2019gradient}. Challenges still exist, however, in applying these methods for model-based inversion for complex 3D systems.

The DSI procedure developed in this work builds on the methodology introduced by Sun and Durlofsky~\cite{Sun2017} and Sun et al.~\cite{SunCG}, and recent developments presented by Lima et al.~\cite{lima2019data}. Data variables, which in the cases considered here comprise a concatenation of the time-varying water injection and phase production rates in all wells in the model, are generated by performing flow simulation on prior geological models. These prior geological models are realizations consistent with the underlying geological scenario and conditioned to hard data at wells, but they do not provide flow predictions in agreement with observations. The data variables for the prior realizations are assembled into data vectors, and posterior predictions (data vectors) are constructed from the prior data variables and observations using DSI. The method operates within a Bayesian framework, and posterior samples can be generated using the randomized maximum likelihood (RML) method, originally developed for model-based inversion by Kitanidis~\cite{kitanidis1986parameter} and Oliver~\cite{oliver1996multiple}. Ensemble-based sampling procedures, specifically ensemble smoother with multiple data assimilation~\cite{lima2019data}, can also be used to generate posterior data samples. The DSI framework is flexible, and has recently been extended by Sun and Durlofsky~\cite{sun2019data} to predict spatial variables (CO$_2$ saturation in the top layer of the model at a particular time in a CO$_2$ storage problem) and by Jiang et al.~\cite{jiang2019data} for use in oil production optimization under uncertainty.

Methods that share important similarities with DSI have been developed by a number of investigators. Scheidt et al.~\cite{scheidt2015prediction} proposed a prediction-focused approach based on the statistical relationship between historical and predicted data. Satija and Caers~\cite{satija2015direct} applied canonical functional component analysis to build a linear relationship in the low-dimensi\-onal space. This framework was applied to groundwater tracer-flow and oil reservoir problems~\cite{satija2017direct, hermans2016direct}. These methods do, however, rely on a degree of linearity in the reduced space. Jeong et al.~\cite{jeong2018learning} introduced a learning-based forecast-focused method by applying dimension reduction with artificial neural network (ANN). They applied ANN and support vector regression in the latent space to build the statistical relationship. More recently, Lima et al.~\cite{lima2019data} proposed a DSI implementation that uses ESMDA for posterior sampling along with localization. In this approach, data variables are updated directly. ESMDA enables fast and efficient posterior sampling for cases with a large number of data variables, though in the absence of parameterization or truncation, unphysical behavior can occasionally occur (such as negative water production rates).  

Parameterization can be highly useful in both model-space and data-space methods as it reduces the number of variables that must be determined, while acting to maintain the statistical structure inherent in prior (model or data) realizations. In existing DSI methods~\cite{SunCG}, PCA with histogram transformation is applied. Histogram transformation preserves the marginal distribution of data variables, but not the correlations between data variables. In many cases this is not a major concern, but when `derived' quantities, involving combinations of DSI data variables, are of interest this treatment can lead to inaccuracy (an example of a derived quantity is the total liquid production rate over the entire system). To address this issue, in this work we incorporate a recurrent autoencoder into our existing DSI procedure. The RAE applies an autoencoder for dimensionality reduction~\cite{kramer1991nonlinear}, along with a long-short-term memory (LSTM) recurrent neural network architecture~\cite{hochreiter1997long}, to represent time series corresponding to the time-varying flow rates at wells. Consistent with its use by Sagheer and Kotb~\cite{sagheer2019unsupervised} for the prediction of multivariate time series, LSTM is used here for both encoding and decoding. Unlike PCA with histogram transformation, the RAE is able to maintain complex correlations in well data. The RAE is combined with ESMDA (operating in reduced space) to generate posterior DSI flow predictions. 

This paper proceeds as follows. In Section~\ref{sec:dsi_method}, we present the standard DSI method~\cite{SunCG} and discuss the application of ESMDA for DSI~\cite{lima2019data}. We also describe the use of data parameterization in combination with ESMDA. In Section~\ref{sec:rae}, the RAE procedure and its use in DSI is described in detail. Next, in Section~\ref{sec:result_2d}, the new DSI framework, involving RAE with ESMDA, is applied to a 2D oil-water channelized system. We evaluate posterior statistics and correlations for various quantities of interest through comparisons to reference (computationally expensive) rejection sampling results. We further evaluate the performance of RAE with ESMDA by applying the method to a 3D system containing oil, water and gas. Concluding remarks and suggestions for future work in this area are provided in Section~\ref{sec:conclusion}. 
\section{Data-Space Inversion Methodology}\label{sec:dsi_method}

In this section, we describe the basic data-space inversion (DSI) procedure developed by Sun and Durlofsky~\cite{Sun2017} and Sun et al.~\cite{SunCG}, and the DSI implementation based on ensemble smoother with multiple data assimilation (ESMDA) proposed by Lima et al.~\cite{lima2019data}. We then discuss the combination of data parameterization and ESMDA within the DSI framework. 

\subsection{Basic DSI formulation}

As discussed in the Introduction, traditional data assimilation procedures generate posterior model parameters conditioned to observations. These calibrated (history matched) models can then be used for flow forecasting. DSI methods, by contrast, require that a substantial number of prior flow simulations be performed (typically 500--1000). Then, using these prior simulation results in conjunction with observed data, they generate posterior forecasts for quantities of interest, such as phase production rates, directly. A Bayesian framework is adopted, which enables the construction of the posterior distribution of data variables. 

In the DSI process, we first generate an ensemble of prior geological realizations $\Bm_i$, $i = 1, \ldots, \Nr$, where $\Nr$ is the total number of prior realizations. These realizations can be based on one or more geological scenarios (defined, e.g., by a training image or multi-Gaussian correlation structure) and are constructed using geological software. The models are conditioned to (hard) property data at wells when available. Next, flow simulation is performed on each realization under the appropriate well settings and boundary conditions. The resulting data vectors containing the quantities of interest are denoted by $\Bd_i$, $i = 1, \ldots, \Nr$. The simulation procedure, which is a nonlinear forward process that maps model parameters $\Bm_i$ to data variables $\Bd_i$, is denoted by $\Bg$; i.e.,
\begin{equation}
    \Bd_i = \Bg(\Bm_i).
\end{equation}
In this paper, the data vector $\Bd \in \R^{\Nf \times 1}$ is a concatenation of all well injection and phase production rates at each simulation time step. Thus $\Nf = \Nqoi \times \Nt$, where $\Nqoi$ is the number of quantities of interest, and $\Nt$ is the number of time steps in the full simulation period. 

The prior data vectors $\Bd_i$ contain data variables from both the historical (history match, denoted hm) period over which observations are collected, $(\Bdhm)_i \in \R^{\Nhm \times 1}$, and over the prediction period, $(\Bdpred)_i \in \R^{\Npred \times 1}$, where $\Nhm$ and $\Npred$ denote the number of data in the historical period and prediction periods. The data vector can thus be written as $\Bd_i = [(\Bdhm)_i^T, (\Bdpred)_i^T]^T$. The observed data is denoted $\Bdobs$, with $\Bdobs \in \R^{\Nhm \times 1}$. In the examples in this paper, the `true' model is a particular realization $\Bm_{\rm true}$ that is not included in the original set of $\Nr$ realizations. The `true' data vector is then denoted $\Bd_{\rm true}$ ($\Bd_{\rm true}=\Bg(\Bm_{\rm true})$). In our setting the observed data $\Bdobs$ is always taken to have some amount of measurement error associated with it, so it differs from $\Bd_{\rm true}$. Thus we have
\begin{equation}
    \Bdobs = \Bd_{\rm true} + \Bepsilon = H\Bd_{\rm true} + \Bepsilon,
\end{equation}
where $H \in \R^{\Nhm \times \Nf}$ denotes a selection matrix that extracts the data corresponding to $\Bdhm$ from $\Bd$, $\Bepsilon \in \R^{\Nhm \times 1}$ represents measurement error, which is here taken to be random noise sampled from a Gaussian distribution with zero mean and covariance $\CD$.

In the DSI framework, posterior samples of data variables are generated using a Bayesian framework. The posterior probability density function (PDF) of data variables $\Bd$ conditioned to dynamic flow observations $\Bdobs$ is given by
\begin{equation}
    p(\Bd|\Bdobs) = \frac{p(\Bdobs | \Bd)p(\Bd)}{p(\Bdobs)} \propto p(\Bdobs|\Bd) p(\Bd),
\end{equation}
where $p(\Bdobs | \Bd)$ denotes the conditional PDF of $\Bdobs$ given $\Bd$, and $p(\Bd)$ denotes the prior PDF of data vector $\Bd$. Note that we use $\Bd$ here rather than $\Bd_i$ since we are referring to random variables associated with both prior and posterior distributions of $\Bd$. When the data vector $\Bd$ follows a Gaussian distribution, the posterior PDF is (see the discussion in~\cite{Sun2017})
\begin{equation}\label{eq:post_pdf}
    \begin{split}
         p(\Bd|\Bdobs) \propto & \ \text{exp} \Big(-\frac{1}{2}(H\Bd - \Bdobs) ^T\CD^{-1}(H\Bd- \Bdobs) \\
         &-\frac{1}{2}(\Bd - \Bdprior)^T\Cpd^{-1}(\Bd - \Bdprior) \Big),
    \end{split}
\end{equation}
where $\Bdprior$ and $\Cpd$ are the mean and covariance of the prior data variables $\Bd_i$, $i = 1, \ldots, \Nr$. 

The data vector $\Bd$ is generally non-Gaussian, however, due to the high degree of nonlinearity associated with multiphase subsurface flow problems. This is the case even if the geological model is multi-Gaussian. The posterior distribution in this case is much more complicated and cannot be expressed directly as in equation~\eqref{eq:post_pdf}. As described in~\cite{Sun2017}, the strong correlations between data variables also render $\Cpd$ low rank and not invertible. 

We thus apply a parameterization method to represent the high-dimensional variable in terms of a low-dimensional latent variable $\Bxi \in \R^{\Nl \times 1}$, where $\Nl$ is the dimension of the latent variable. We use $\Bd \approx \Bdhat = \Bf(\Bxi)$ to express the mapping process from $\Bxi$ to $\Bdhat$, where $\Bdhat \in \R^{\Nf \times 1}$ represents the parameterized data vector. The correlations between data variables and the physical character of the flow response should, ideally, be retained in the parameterized representation.

Sun et al.~\cite{SunCG} proposed the use of a PCA-based parameterization followed by histogram transformation (HT) to represent the data variable. This treatment, applied separately for each component in $\Bd$, preserves the marginal distribution for each variable but not the joint distribution. The detailed procedure will be discussed in the next section. With this parameterization, the posterior PDF of the latent variable $\Bxi$ can be written as
\begin{equation}\label{eq:post_pdf_xi}
    \begin{split}
        &p(\Bxi|\Bdobs) \propto \\
        &\text{exp} \left(-\frac{1}{2}(H\Bf(\Bxi) - \Bdobs) ^T\CD^{-1}(H\Bf(\Bxi)- \Bdobs) -\frac{1}{2}\Bxi^T\Bxi \right).
    \end{split}
\end{equation}
Since the parameterization process is nonlinear, a sampling procedure is required to generate posterior results.

The randomized maximum likelihood method (RML) was used by Sun et al.~\cite{SunCG} to provide posterior samples of $\Bdhat = \Bf(\Bxi_{\text{rml}})$; i.e., 
\begin{equation}
    \begin{split}
        \Bxi_{\text{rml}} =  \underset{\Bxi}{ \operatorname{argmin}} \{ & \frac{1}{2}(H\Bf(\Bxi)- \Bdobs^*) ^T\CD^{-1}(H\Bf(\Bxi) - \Bdobs^*) \\
        & + \frac{1}{2}(\Bxi - \Bxi^*)^T(\Bxi - \Bxi^*) \},
    \end{split}
\end{equation}
where $\Bdobs^*$ is a realization of the observed data (sampled from $N(\Bdobs, \CD)$), and $\Bxi^*$ is a prior realization of $\Bxi$, sampled from $N(\mathbf{0}, I)$. The regularization-type term $(\Bxi - \Bxi^*)^T(\Bxi - \Bxi^*)$ enters once the data vector is parameterized using PCA. Multiple posterior samples of $\Bxi_{\text{rml}}$, and thus of $\Bdhat$, are generated by solving the minimization problem for different random samples of $\Bdobs^*$ and $\Bxi^*$. Given the posterior $\Bdhat$ vectors, the required statistics for all quantities of interest can be generated.

\subsection{ESMDA in DSI}
The DSI framework is flexible, and different methods for parameterization and for sampling from the posterior distribution can be implemented. Lima et al.~\cite{lima2019data} recently introduced a DSI implementation with ESMDA, which we now review. We then discuss how this sampling procedure can be combined with data parameterization methods.  

\subsubsection{DSI Implementation with ESMDA}
ESMDA, developed by Emerick and Reynolds~\cite{emerick2013ensemble}, is a popular approach for model-based history matching. ESMDA generates posterior samples by assimilating data multiple times over an ensemble with inflated error covariance. Within the DSI framework, the posterior samples of data variables $\Bd$, conditioned to the observations, can be generated directly (posterior models are not required). 

The update equation for $\Bd$ is given by Lima et al.~\cite{lima2019data}:
\begin{equation}\label{eq:esmda}
\begin{split}
    &\mathbf{d}_i^{k+1} = \\
    &\mathbf{d}_i^{k} + C_{\text{d},  \text{d}_{\text{hm}}}^k \left(C_{\text{d}_{\text{hm}}}^k + \alpha_k\CD \right)^{-1}(\Bdobs+\sqrt{\alpha_k}\mathbf{e}_i^k - (\Bdhm)_i^k),
\end{split}
\end{equation}
for $i = 1, \ldots, \Nr$, and $k = 1, \ldots, \Na$, where $\Na$ is the number of iterations in the data assimilation process, and $\alpha_k$ represents the inflation coefficient at iteration $k$. The $\alpha_k$ are required to satisfy the constraint $\sum_{k=1}^{\Na}\frac{1}{\alpha_k} = 1$ to generate the correct posterior for the linear-Gaussian case~\citep{emerick2013ensemble}. The vector $\mathbf{e}$ in equation~\eqref{eq:esmda} denotes random noise sampled from $N(\mathbf{0}, \CD)$. The portion of the data vector from the historical period ($\Bdhm$) is extracted from the updated $\Bd$ from the previous iteration. The cross-covariance matrix $C_{\text{d}, \text{d}_{\text{hm}}} \in \R^{\Nf \times \Nhm}$ represents the covariance between data variables $\Bd$ and $\Bdhm$, and the auto-covariance matrix $C_{\text{d}_{\text{hm}}} \in \R^{\Nhm \times \Nhm}$ denotes the covariance of data variables $\Bdhm$. Distance-based localization was incorporated by Lima et al.~\cite{lima2019data}. This localization, applied based on the physical distance between wells and a specified critical length, acts to eliminate the effects of long-distance interactions (and thus correlations) between well data. Localization is often required with ensemble-based methods to avoid ensemble collapse, which can occur in problems with a large amount of observed data.

Lima et al.~\cite{lima2019data} found this ESMDA sampling method to outperform RML for the DSI problems considered. They did not apply any parameterization in their DSI implementation, but instead used the data vectors directly. Truncation of the posterior results was, however, applied to ensure that the well rates remained physical. This could entail, for example, mapping a negative rate to be zero. In Section~\ref{sec:result_2d}, we will compare a number of DSI treatments including ESMDA with and without parameterization.

\subsubsection{ESMDA Combined with Parameterization}\label{sec:esmda_ksi}
We now introduce the mapping described earlier, $\Bd \approx \Bdhat = \Bf(\Bxi)$, into the ESMDA DSI framework. The resulting update equation, which is analogous to equation~\eqref{eq:esmda} above, is given by
\begin{equation}\label{eq:esmda_ksi}
\begin{split}
    &\Bxi_i^{k+1} = \\
    &\Bxi_i^{k} + C_{\xi, \text{d}_{\text{hm}}}^k(C_{\text{d}_{\text{hm}}}^k + \alpha_k\CD)^{-1}(\Bdobs+\sqrt{\alpha_k}\mathbf{e}_i^k - (\Bdhm)_i^k),
\end{split}
\end{equation}
for $i = 1, \ldots, \Nr$ and $k = 1, \ldots, \Na$. Here $C_{\xi, \text{d}_{\text{hm}}}$ denotes the cross-covariance matrix between latent variables $\Bxi$ and historical data variables $\Bdhm$. At each iteration $k$ we apply 
\begin{equation}
    (\Bdhm)_i^k = H\Bf(\Bxi_i^k).
\end{equation}
The matrices $C_{\xi, \text{d}_{\text{hm}}}^k$ and $C_{\text{d}_{\text{hm}}}^k$ are then evaluated using the updated ensemble of $\Bxi^k$ and $\Bdhm^k$ vectors. After $\Na$ iterations are completed, the posterior results for $\Bd$ ($\Bd \approx \Bdhat=\Bf(\Bxi)$) are generated from the posterior results for $\Bxi$. 
\section{Data Parameterization Using Recurrent Autoencoder}
\label{sec:rae}

In this section, we introduce a new deep-learning-based data parameterization approach that applies a recurrent autoencoder (RAE). This nonlinear parameterization procedure represents time-series data variables $\Bd$ in terms of latent space variables $\Bxi$. To motivate this discussion, we first describe an existing DSI parameterization that combines principal component analysis (PCA) with histogram transformation. This approach preserves key marginal distributions and is thus often effective, but it does not in general maintain correlations between data-variable components. This is the motivation for using the RAE in combination with DSI.

\subsection{PCA with Histogram Transformation}
\label{sec:PCA_HT}

We now briefly review the DSI procedure described by Sun et al.~\cite{SunCG}, which employs PCA for dimension reduction followed by histogram transformation (HT). To construct the PCA basis matrix, we perform singular value decomposition on the data matrix $D \in \R^{\Nf \times \Nr}$, constructed as
\begin{equation}
    D = \frac{1}{\sqrt{\Nr - 1}}[\Bd_1 - \Bdprior \ \  \Bd_2 - \Bdprior \ \  \cdots  \ \  \Bd_{\Nr} - \Bdprior], 
\end{equation}
where $\Bdprior \in \R^{\Nf \times 1}$ is the mean of the prior data vectors. We define the basis matrix $\Phi \in \R^{\Nf \times \Nl}$. The value for $\Nl$ (latent space dimension, where $\Nl<\Nr$) can be determined based on an `energy' criterion. Realizations of $\Bd$, referred to here as $\Bdpca$, can now be generated through application of
\begin{equation}\label{eq:reconstruct_pca}
    \Bdpca = \Phi \Bxipca + \Bdprior,
\end{equation}
where $\Bxipca \in \R^{\Nl \times 1}$ contains uncorrelated standard normal (latent) variables. 

PCA entails a linear mapping from standard-normal $\Bxipca$ to data variables $\Bdpca$. Thus $\Bdpca$ are Gaussian-distributed, with mean $\Bdprior$ and covariance $\Phi\Phi^T$. This linear Gaussian parameterization can generate unphysical behavior when the data are non-Gaussian, as they indeed are in two-phase subsurface flow problems. For example, negative flow rates may be observed for quantities with mean value close to zero. Histogram transformation (HT) is thus applied as a post-processing step to avoid unphysical effects. The goal of HT is to map the initial Gaussian distribution of $\Bdpca$ to a distribution consistent with the prior distribution of $\Bd$. The cumulative distribution function (CDF) is used for this mapping. The transformed data vector $\Bdhat$ is expressed as
\begin{equation}
    \Bdhat = h_T(\Bdpca) = f_T^{-1}(f_I(\Bdpca)),
\end{equation}
where $f_I(\Bdpca)$ represents the initial CDF of $\Bdpca$ (Gaussian distributed), $f_T(\Bd)$ represents the target CDF of the prior ensemble (non-Gaussian distributed), and $h_T$ represents the histogram transformation process. As shown in~\cite{SunCG}, the post-processing step enables DSI to generate improved posterior predictions. 

The histogram transformation is applied separately to each component in $\Bdpca$. This ensures that the marginal distributions are captured, but it does not maintain correlations in space (between different wells) or in time. For example, in some cases the problem physics dictate that well rates should increase monotonically in time, which can be viewed as requiring that the time-correlations in $\Bd$ be of a particular form. As we will see, this behavior may not be retained following application of the PCA and HT procedures. This motivates the development of the RAE procedure, which we now describe.

\subsection{Recurrent Autoencoder (RAE)}

DSI predictions in the current setting are essentially multivariate time series forecasts. Such problems can be challenging to parameterize. Here we address this problem using a recurrent autoencoder, which is an autoencoder based on recurrent neural networks (RNNs). An autoencoder~\citep{hinton2006reducing} is a nonlinear generalization of PCA. It is comprised of two components: an encoder, which transforms high-dimensional input data to a low-dimensional latent space, and a decoder, which maps from the latent space to the high-dimensional space. Various deep neural networks can be applied in the encoder and decoder components. For the time-series data in the DSI framework, we apply RNN in both the encoder and decoder to capture the temporal dynamics.

\subsubsection{Long-Short-Term Memory (LSTM) Recurrent Network}

Long-short-term memory~\citep{hochreiter1997long} is an efficient method within the class of RNNs. It enables neural networks to learn both short-term and long-term dependencies within the data. LSTM preserves the correlations between different quantities and different time steps when reconstructing data and generating new samples. 

\begin{figure}[!htpb]
    \centering
    \includegraphics[trim = 10 90 10 90, clip, width =0.47 \textwidth]{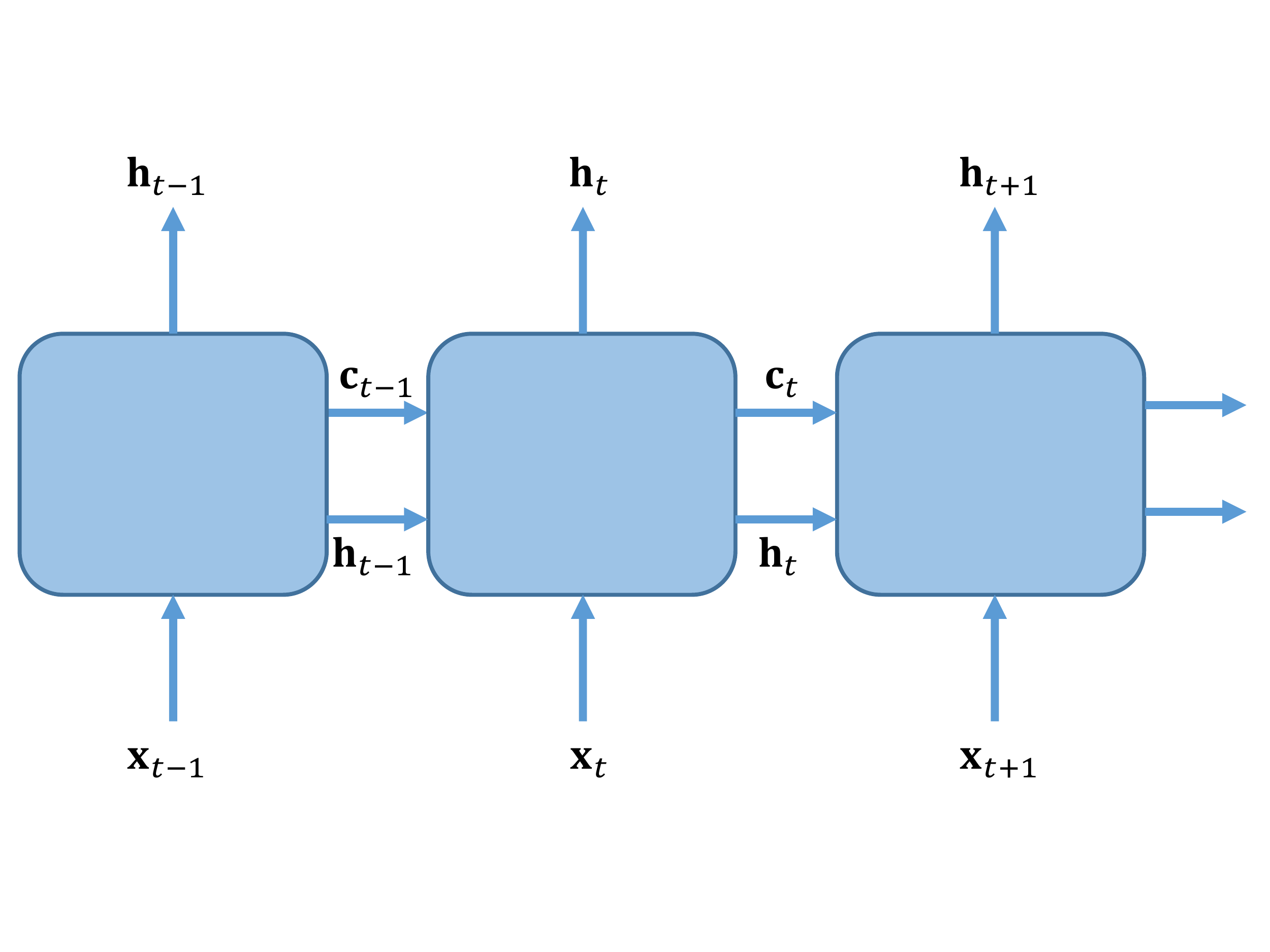}
    \caption{Chain structure of LSTM}\label{fig:lstm}
\end{figure}

LSTM has a chain structure of repeating neural network cells, as shown in Fig.~\ref{fig:lstm}. At each time step $t$, the neural network cell receives input $\Bx_t \in \R^{\Nd \times 1}$, the hidden (output) state from the previous cell $\Bh_{t-1}\in \R^{\Nh \times 1}$, and the previous cell state $\Bc_{t-1} \in \R^{\Nh \times 1}$, where $\Nd$ is the length of input data vector and $\Nh$ is the number of hidden units. The cells have four interacting layers to receive input features at time $t$ and to `remember' information for long and short periods of time. 

Three gates, which control the information flow, are used in each cell. These include the forget gate (controlled by $\Bft \in \R^{\Nh \times 1}$), the input gate (controlled by $\Bit \in \R^{\Nh \times 1}$), and the output gate (controlled by $\Bot \in \R^{\Nh \times 1}$). The information to retain or forget from the previous cell state $\Bc_{t-1}$ is controlled by the forget gate $\Bft$. The portion of the input features to be written into the long-term cell state $\Bh_{t}$ is controlled by the input gate $\Bit$. The output $\Bh_t$ from the current cell state $\Bc_{t}$ is controlled by the output gate $\Bot$. The values of these quantities are based on the previous hidden state $\Bh_{t-1}$ and input $\Bx_t$. The expressions for the different gates $\Bft$, $\Bit$, $\Bot$ are
\begin{equation}
    \begin{split}
        &\Bft = \sigma(W_f[\Bh_{t-1}, \Bx_t] + \Bb_f), \\
        &\Bit = \sigma(W_i[\Bh_{t-1}, \Bx_t] + \Bb_i), \\
        &\Bot = \sigma(W_o[\Bh_{t-1}, \Bx_t] + \Bb_o), \\    
    \end{split}
\end{equation}
where $\sigma(\cdot)$ represents the sigmoid function. The proposed cell state $\Bchat_t$ is calculated as 
\begin{equation}
    \Bchat_t = \text{tanh}(W_x[\Bh_{t-1}, \Bx_t] + \Bb_c).
\end{equation}
In the above equations, matrices $W \in \R^{\Nh \times (\Nh + \Nd)}$ and vectors $\Bb \in \R^{\Nh \times 1}$ represent the weight and bias terms respectively. These are shared across all of the recurrent cells in the LSTM framework. 

The cell state $\Bc_{t}$ is updated by the forget gate $\Bft$ and the input gate $\Bit$ via
\begin{equation}
    \Bc_t = \Bft \circ \Bc_{t-1} + \Bit \circ \Bchat_t,
\end{equation}
where the symbol $\circ$ means the element-wise product. The output $\Bh_t$ at time step $t$ is determined from the cell state $\Bc_t$ controlled by the output gate $\Bot$ as follows
\begin{equation}
    \Bh_t = \Bot \circ \text{tanh}(\Bc_t).
\end{equation}
The use of the LSTM framework acts to preserve key dependencies in the time-series data. The recurrent architecture used to incorporate LSTM in the DSI parameterization will be discussed in the next section.

\subsubsection{RAE Architecture}
Autoencoders have been widely used in different types of deep neural networks for natural language processing~\cite{serban2017multiresolution, li2015hierarchical} and signal processing~\cite{marchi2017deep, mousavi2019unsupervised}. LSTM was used for encoding and decoding in the context of multivariate time series by Sagheer and Kotb~\cite{sagheer2019unsupervised}. Here we focus on parameterizing data vectors and combining the latent variables with data assimilation to generate posterior prediction results. The parameterization process is referred to as a recurrent autoencoder because it  combines an autoencoder with LSTM (which is a recurrent neural network).

\begin{figure}[htbp!]
    \centering
    \includegraphics[trim = 10 95 10 95, clip, width = 0.47 \textwidth]{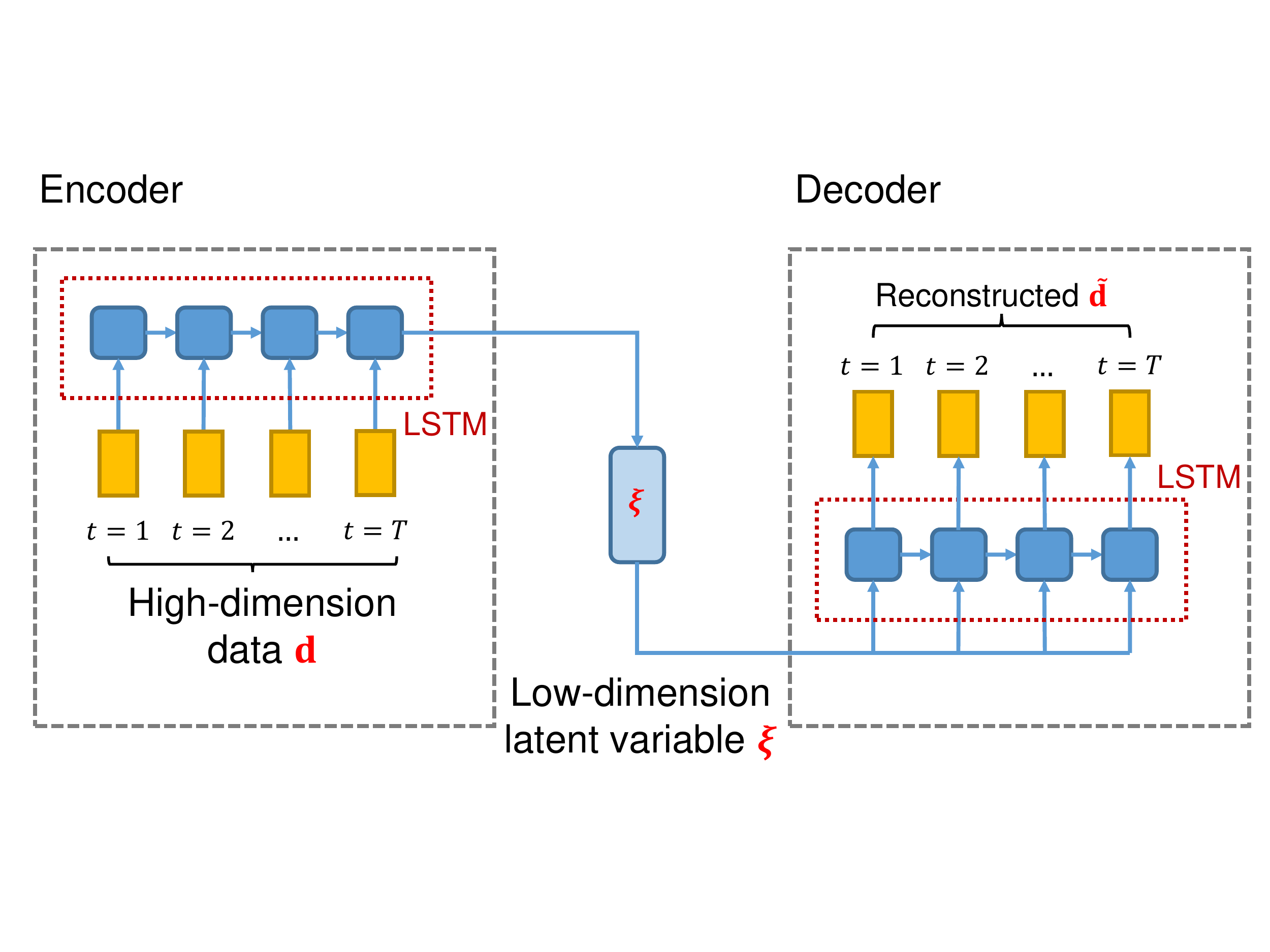}
    \caption{Recurrent autoencoder: Encoder contains one LSTM layer and decoder contains three LSTM layers (see Table~\ref{tab:arc_encoder_decoder})}\label{fig:RAE}
\end{figure}

The structure of the RAE is shown in Fig.~\ref{fig:RAE}. The encoder portion compresses the high-dimensional (full-order) time series data $\Bd$ to a low-dimensional latent variable $\Bxirae$. We apply LSTM in the encoder for the nonlinear mapping $f_e(\Bd; W_e): \R^{\Nf \times 1} \rightarrow \R^{\Nl \times 1}$, where $W_e$ denotes the weights in the encoder. The latent variables $\Bxirae$ are calculated as
\begin{equation}\label{eq:encoder}
    \Bxirae = f_e(\Bd; W_e).
\end{equation}
The input for the LSTM cells in the encoder is the data vector $\Bd \in \R^{\Nf \times 1}$, where $\Nf = \Nqoi \times \Nt$. Following the architecture in the previous section, the vector $\Bx_t \in \R^{\Nqoi \times 1}$ input to the LSTM cells includes the values of all data quantities at time step $t \ (t = 1, 2, \cdots, \Nt)$. The output variables $\Bxirae \in \R^{\Nl \times 1}$ are mapped from the hidden states $\Bh_t$ in all time steps in the LSTM cells through an additional fully connected layer.

\begin{table}[!htb] 
\caption{Architecture of Recurrent Autoencoder}
\centering
    \begin{tabular}{ c c c }
    \hline
     Net & Layer & Output Size \\
    \hline
     Encoder & Input & $(\Nqoi, \Nt)$\\
      & LSTM (units $=\Nh$)  & $(\Nh, \Nt)$\\
      & Flatten & $(\Nh \times \Nt, 1)$\\
      & Dense & $(\Nl, 1)$\\
      Decoder & Input & $(\Nl, 1)$\\
      & RepeatVector & $(\Nl, \Nt)$\\
      & LSTM (units $=\Nh$)  & $(\Nh, \Nt)$\\
      & LSTM (units $=\Nh$)  & $(\Nh, \Nt)$\\
      & LSTM (units $=\Nh$)  & $(\Nh, \Nt)$\\
      & Dense (activation function: $\text{tanh}$) & $(\Nqoi, \Nt)$\\
     \hline
     \label{tab:arc_encoder_decoder}
\end{tabular}
\end{table}

The decoder portion reconstructs high-dimensional data $\Bdhat$ using the latent-space variable $\Bxirae$. The decoding process is represented as $f_d(\Bxirae; W_d): \R^{\Nl \times 1} \rightarrow \R^{\Nf \times 1}$, where $W_d$ denotes the weights in the decoder. The mapping function is written as 
\begin{equation}\label{eq:decoder}
    \Bdhat = f_d(\Bxirae; W_d).
\end{equation}
LSTM is applied in the decoder to generate the multivariate time-series data. We stack three LSTM layers in the decoder to allow for a more complicated network structure in depth, which enables better complex representation of the latent features and results in more physical time-series data. The input data $\Bx_t$ for time step $t$ for the first LSTM layer is the latent-space vector $\Bxirae$. At every time step, we again input $\Bxirae$, which enables the network to learn from both previous time steps and latent features simultaneously. The output (hidden) states $\Bh_t$ are used as the input $\Bx_t$ for the next LSTM layer. We apply a fully connected layer after the stacked LSTM to transform the output $\Bh_t$ of the last LSTM layer to $\Bdhat_{t} \in \R^{\Nqoi \times 1}$ (note that $\Nqoi=\Nd$). 

In the parameterization process, we normalize the input data vector in the range $[-1, 1]$. We apply the activation function $\text{tanh}$ in the decoder to map the reconstructed data vector back to the same range $[-1, 1]$ to avoid unphysical rates. The architecture of the RAE is shown in Table~\ref{tab:arc_encoder_decoder}.

In the RAE training process, we minimize a loss function that measures the dissimilarity between the actual prior simulated data realizations and their reconstructed versions; i.e.,
\begin{equation}
    L_{RAE} = \frac{1}{\Nr}\sum_{i=1}^{\Nr}||\Bd_i - \Bdhat_i||_2^2,
\end{equation}
where $\Nr$ is the number of data vectors (time series) used for training. This loss is the mean square error between the prior data and the reconstructed data. In this training we determine the elements associated with $W_e$ in equation~\eqref{eq:encoder} and $W_d$ in equation~\eqref{eq:decoder}. The minimization is accomplished using the adaptive moment estimation (ADAM) algorithm~\citep{kingma2014adam}.

\subsection{RAE with ESMDA in DSI}
In the DSI framework, we apply RAE for data parameterization and ESMDA for posterior sampling. With these treatments, we are able to consider non-Gaussian prior and posterior latent-space distributions. We now describe the overall framework -- the specific steps are provided in Algorithm~\ref{alg:rae_esmda}.

We first generate an ensemble of prior geological models $\Bm_i$, $i = 1, \ldots, \Nr$, conditioned to any available hard data. Flow simulation is then performed on each of these models, which provides the prior data realizations $\Bd_i$, $i = 1, \ldots, \Nr$. These $\Bd_i$ comprise the training data for RAE. We then train the networks and apply the trained encoder to generate the prior latent variables, $(\Bxirae_{\text{prior}})_i$, $i = 1, \ldots, \Nr$, through application of equation~\eqref{eq:encoder}. Then, given historical data, we apply ESMDA to generate posterior predictions. As discussed in Section~\ref{sec:esmda_ksi}, for each iteration $k$, we represent the historical data via
\begin{equation}\label{eq:dhm_rae}
    (\Bdhm)_i^k = H f_d(\Bxi_i^k; W_d),
\end{equation}
and we update the latent variable $\Bxi_i^k$ by applying equation~\eqref{eq:esmda_ksi}. After multiple data assimilation steps, we apply the decoder to generate the posterior data variables. 

\begin{algorithm}[!ht]{
\caption{DSI framework using RAE and ESMDA}\label{alg:rae_esmda}
\SetAlgoVlined
Perform flow simulation on ensemble of prior models $\Bm_i$ to generate $\Bd_i$, for $i = 1, \ldots, \Nr$ \\
Train the RAE networks with the ensemble of prior data $\Bd_i$, $i = 1, \ldots, \Nr$ \\
Generate prior latent variables $(\Bxirae_{\text{prior}})_i$, $i = 1, \ldots, \Nr$, from the trained encoder $(\Bxirae_{\text{prior}})_i = f_e(\Bd_i; W_e)$ \\
\For{$k \gets 1$ to $\Na$}
{Generate ensemble of historical data $(\Bdhm)_i^k$, $i = 1, \ldots, \Nr$, using $\Bxi_i^{k}$ and equation~\eqref{eq:dhm_rae}, where we use the prior ensemble of latent variables $(\Bxirae_{\text{prior}})_i$ as $\Bxi_i^{1}$ \\ 
Calculate the covariance matrices $C_{\xi, \text{d}_{\text{hm}}}^k$ and $C_{\text{d}_{\text{hm}}}^k$ \\
Update latent data variables $\Bxi_i^{k+1}$, $i = 1, \ldots, \Nr$, applying equation~\eqref{eq:esmda_ksi}\\
}
Generate posterior predictions $(\Bd_{\text{post}})_i = f_d((\Bxirae_{\text{post}})_i; W_d)$, where $(\Bxirae_{\text{post}})_i = \Bxi_i^{\Na+1}$, $i = 1, \ldots, \Nr$, denotes the updated latent variable after data assimilation
}
\end{algorithm}

\subsection{Prior Reconstruction with PCA and RAE}

We have completed our description of the enhanced DSI framework. Before presenting data assimilation results, it is useful to evaluate the RAE parameterization itself. We now compare this procedure to the use of PCA and histogram transformation, referred to as PCA+HT, for the reconstruction of prior data realizations. We consider the 2D channelized system described in Section~\ref{sec:result_2d}. In this case, we fix wellbore pressures and then assemble the simulated well rate data, for eight quantities of interest, into the data vectors. The data include time-varying water injection rates (WIR) for two injectors, and water production rates (WPR) and oil production rates (OPR) for three producers. The number of time steps $\Nt$ for each quantity is 100. Thus each data vector $\Bd_i$ contains 800 elements. We simulate $\Nr = 800$ geomodels and thus generate 800 data realizations for training. There are about 500,000 RAE parameters to determine in the training process. Training requires about 10~minutes using a NVIDA Tesla V100 GPU. For both PCA+HT and RAE, we set the dimension of the latent space to be $\Nl=31$.

\begin{figure*}[!htb]
\centering
\begin{minipage}{.32\linewidth}\centering
\includegraphics[width=\linewidth]{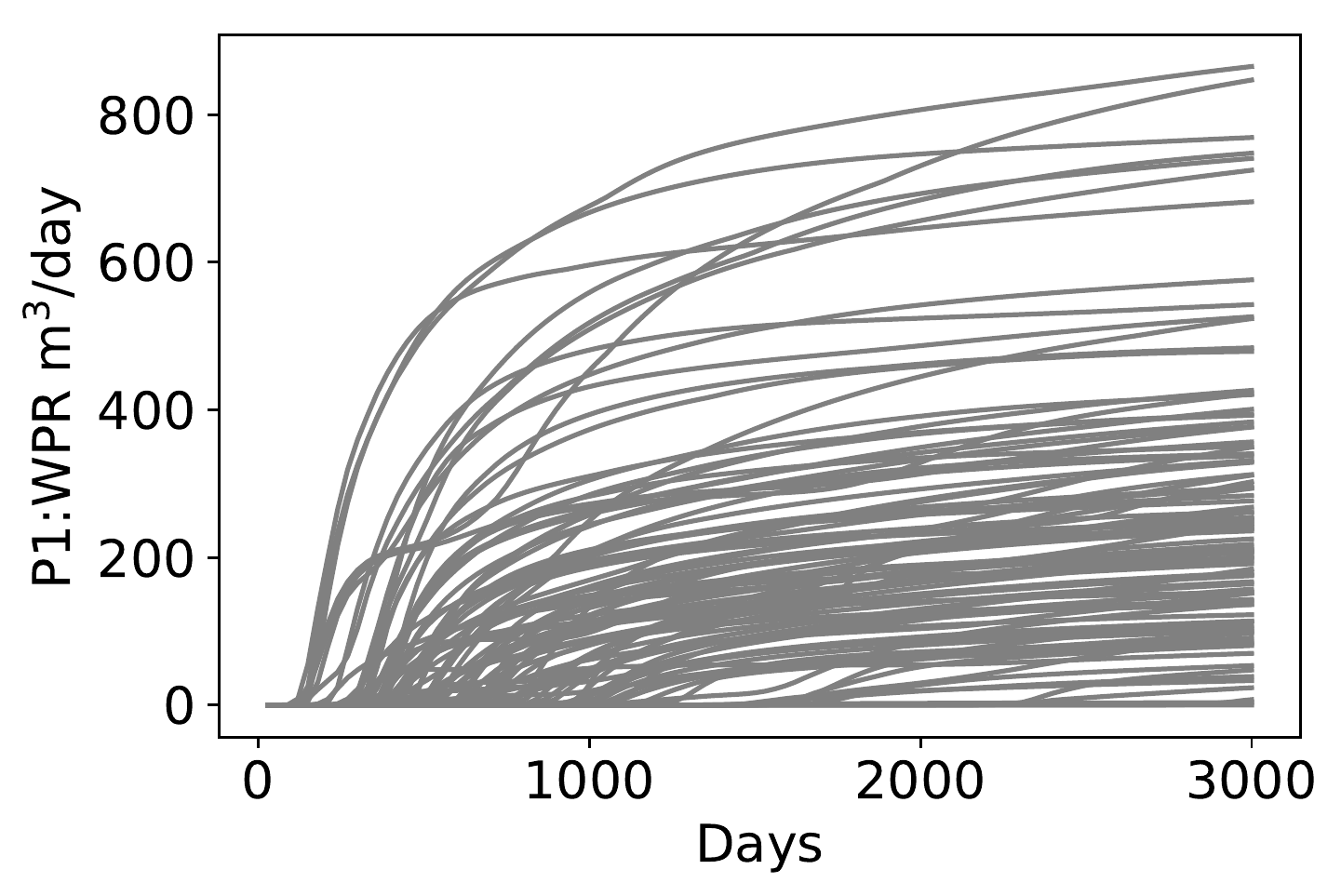}\subcaption{P1 WPR -- Prior}
\end{minipage}
\begin{minipage}{.32\linewidth}\centering
\includegraphics[width=\linewidth]{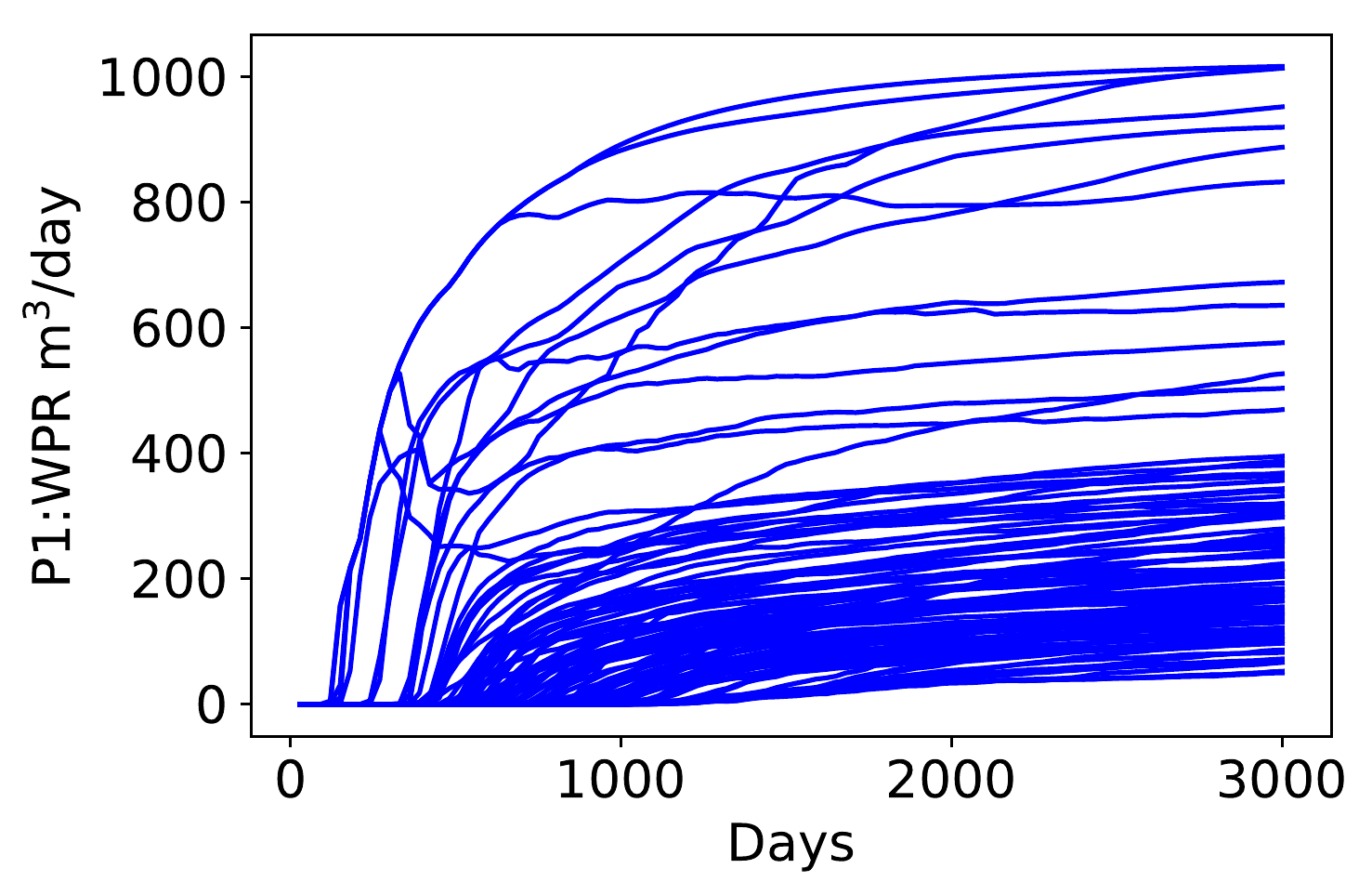}\subcaption{P1 WPR -- PCA+HT}
\end{minipage}
\begin{minipage}{.32\linewidth}\centering
\includegraphics[width=\linewidth]{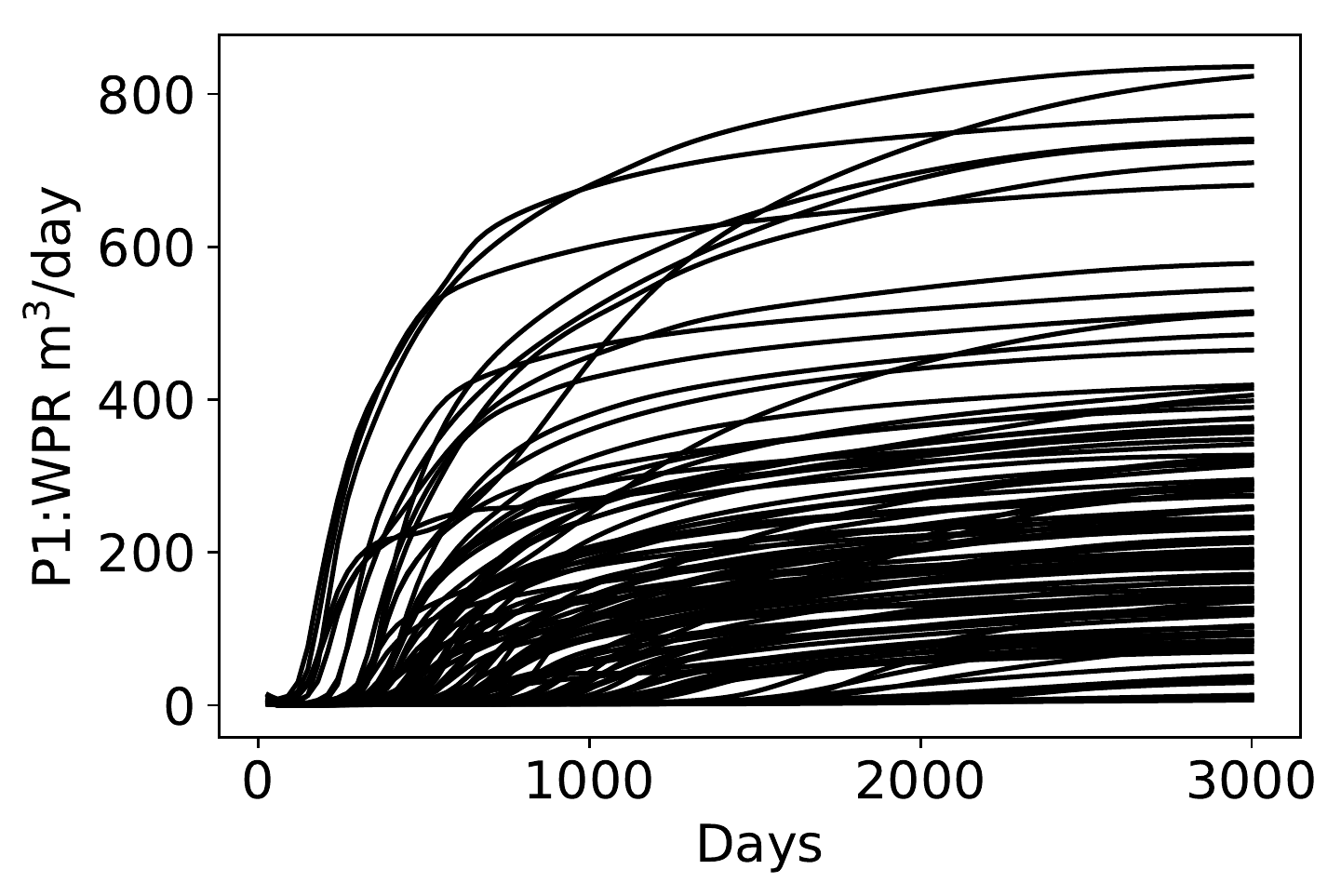}\subcaption{P1 WPR -- RAE}
\end{minipage}

\begin{minipage}{.32\linewidth}\centering
\includegraphics[width=\linewidth]{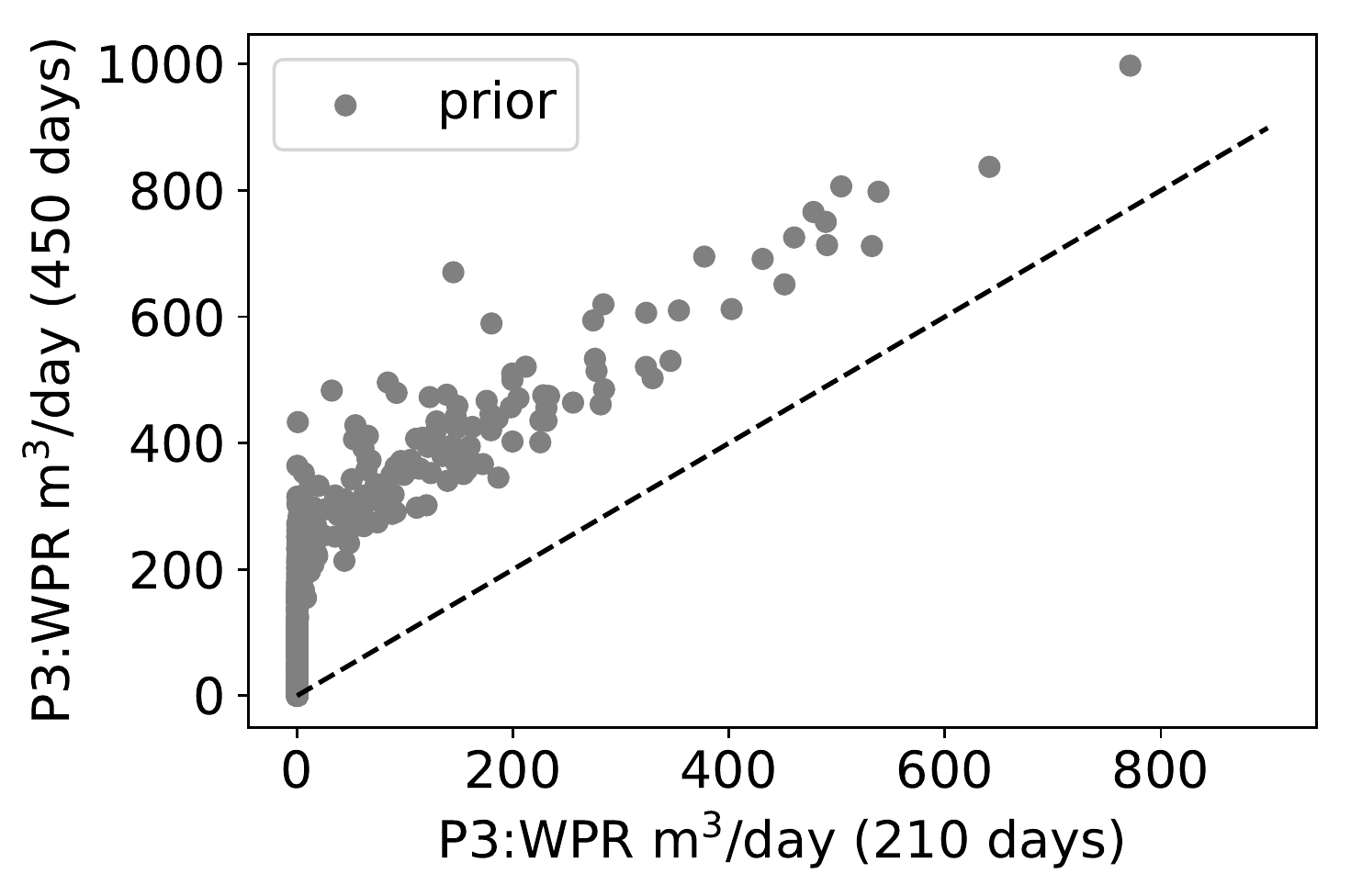} \subcaption{P3 WPR -- Prior}
\end{minipage}
\begin{minipage}{.32\linewidth}\centering
\includegraphics[width=\linewidth]{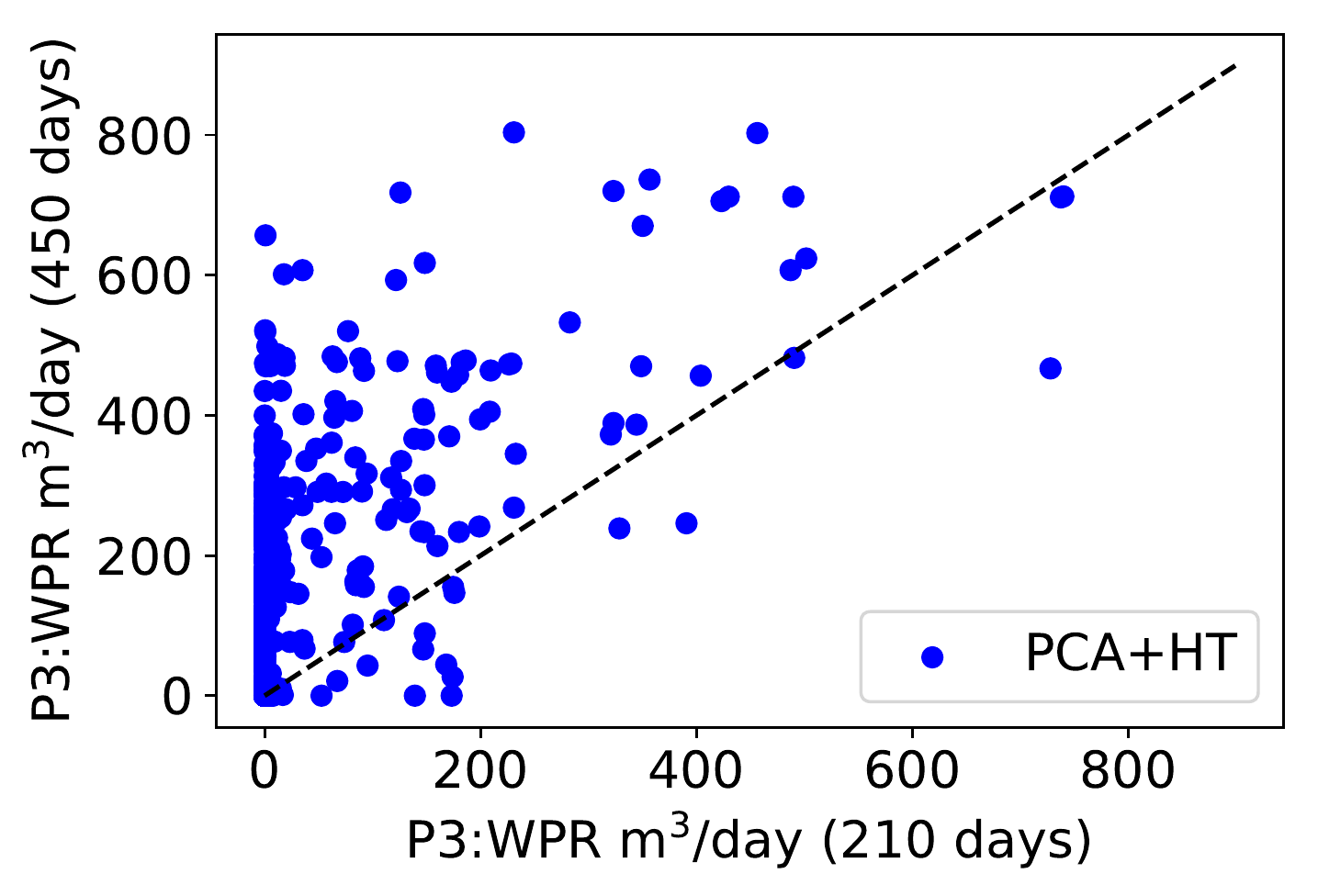} \subcaption{P3 WPR -- PCA+HT}
\end{minipage}
\begin{minipage}{.32\linewidth}\centering
\includegraphics[width=\linewidth]{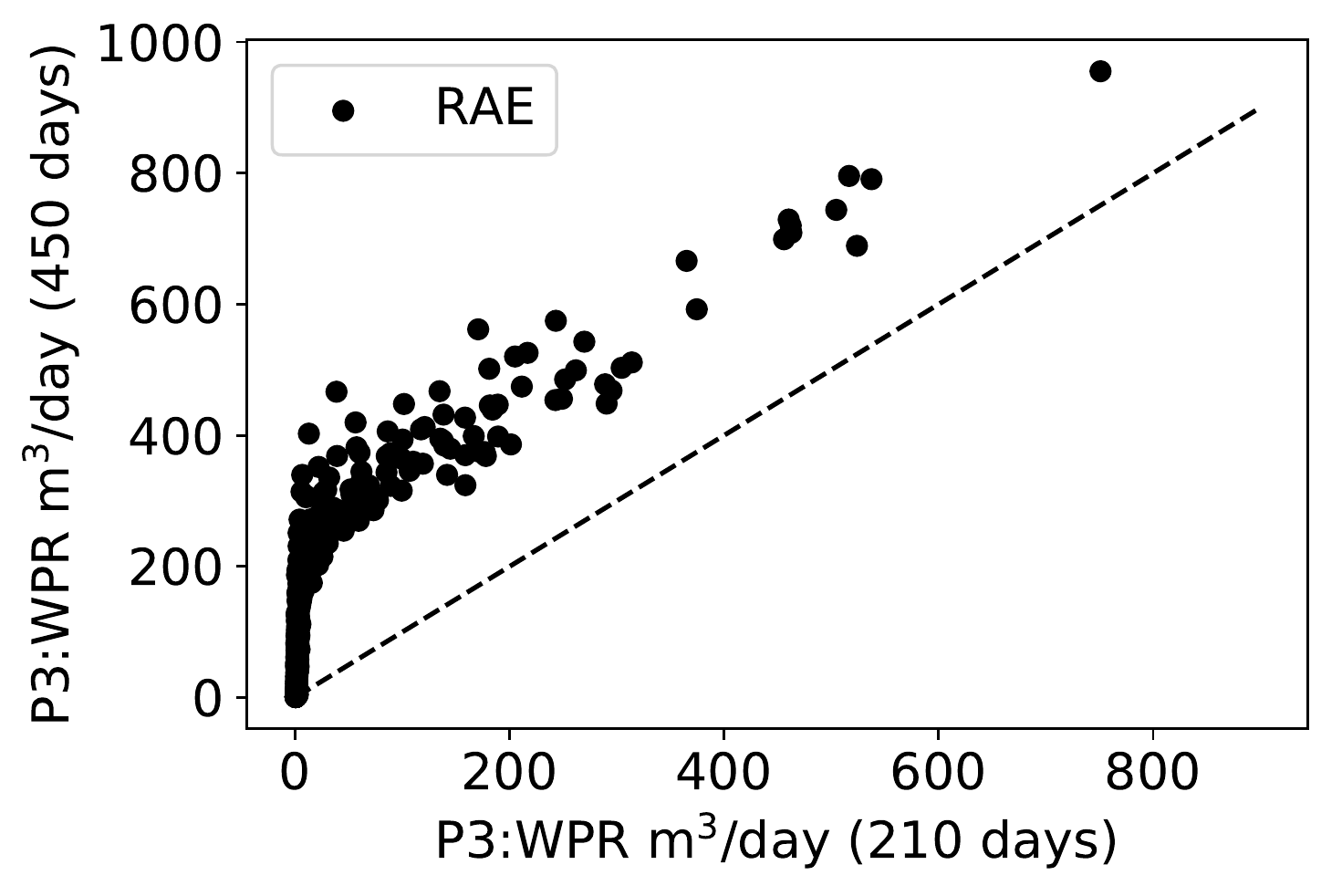} \subcaption{P3 WPR -- RAE}
\end{minipage}
\caption{Prior and reconstructed times series for P1 WPR using PCA+HT and RAE (top row), and cross-plots for P3 WPR at 210 and 450~days (bottom row). Physically, all points in the cross-plots should fall above the 45$^\circ$ line}\label{fig:prior_rec}
\end{figure*}

The upper row of Fig.~\ref{fig:prior_rec} shows the reference prior simulation results (left) along with reconstructed results using PCA+HT (center) and RAE (right), for water production rate in well P1. Results for 100 prior realizations and reconstructed time series are shown in both cases. The prior results indicate that, following water breakthrough, P1 water rate increases monotonically in all data realizations (this is related to the fact that water saturation increases monotonically at production-well blocks, and wells operate at fixed pressure). As discussed earlier, PCA+HT may not preserve time correlations between different time steps, and as a result some of reconstructed results for P1 water rate display nonmonotonic behavior. This unphysical effect is not observed in the RAE curves, which visually coincide more closely to the reference prior simulation results.

The lower row in Fig.~\ref{fig:prior_rec} shows cross plots for water production rates in another well (P3) at two different times -- 210 and 450~days. Because water rate again increases monotonically in time for this well, all points in the reference prior simulation results (left) would be expected to fall above the dashed 45$^\circ$ line, and this is indeed observed. For the reconstructed PCA+HT results (center), it is apparent that the prior distribution is not reproduced, and some points fall below the 45$^\circ$ line. The RAE results are much more consistent with the prior, with none of the points in the reconstruction lying below the 45$^\circ$ line. The results in Fig.~\ref{fig:prior_rec} thus illustrate the advantages of the RAE parameterization compared to the simpler PCA+HT approach. In the next section, we consider the use of these and other methods within the full DSI framework.

\section{History Matching Results using DSI}\label{sec:result_2d}

In this section, we present history matching results using the DSI framework for a 2D channelized system and a 3D Gaussian model. The posterior results from DSI, using a range of treatments, are compared to reference rejection sampling (RS) results. We assess the quality of the various DSI results, including correlations, for several quantities of interest.

\subsection{Model Setup}
\label{sec:model_2d}

\begin{figure*}[!ht]
\centering
\begin{minipage}{.4\linewidth}\centering
\includegraphics[trim = 100 250 150 250, clip, width=\linewidth]{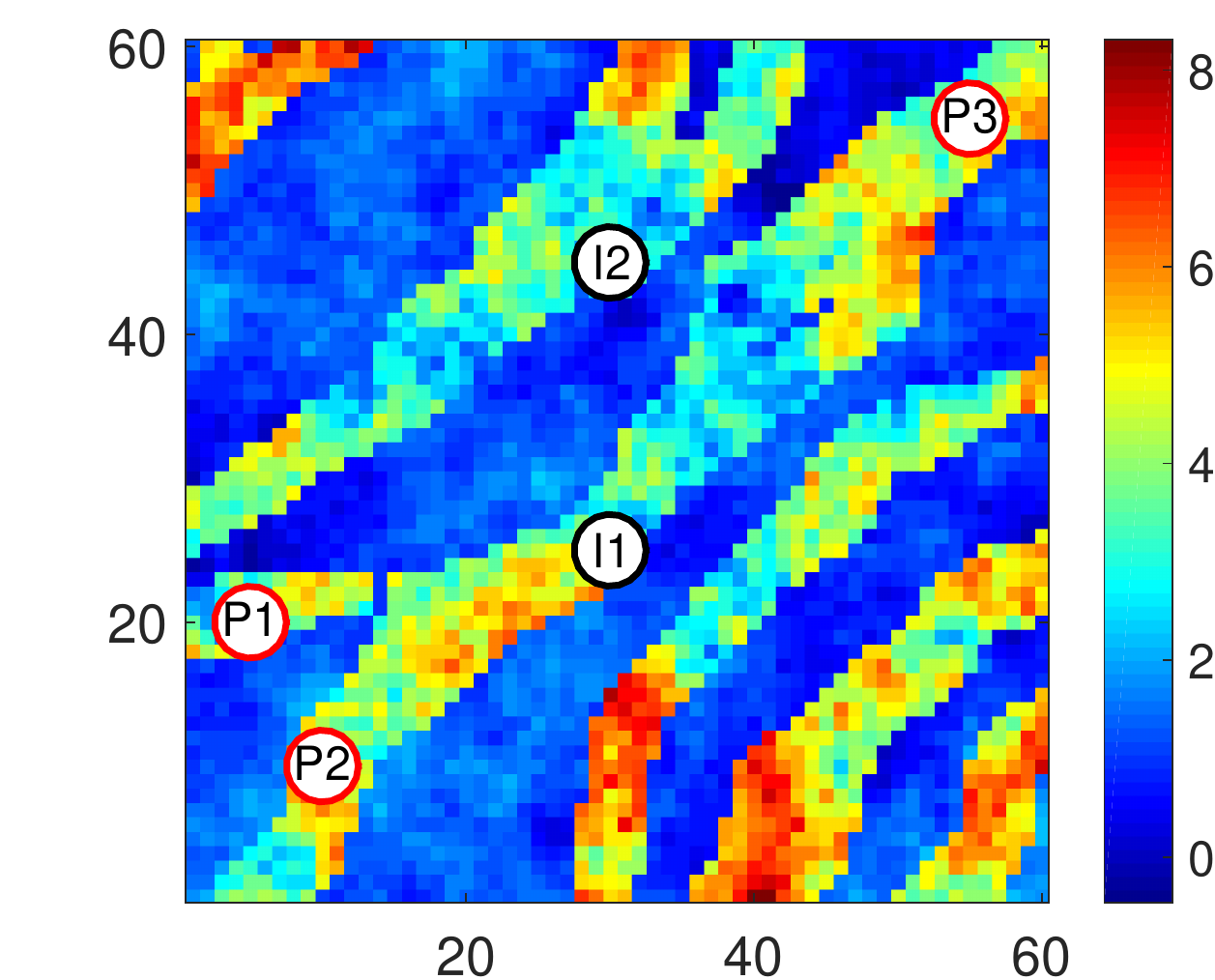}
\subcaption{`True' model}
\end{minipage}
\hspace{.05\linewidth}
\begin{minipage}{.4\linewidth}\centering
\includegraphics[trim = 100 250 150 250, clip, width=\linewidth]{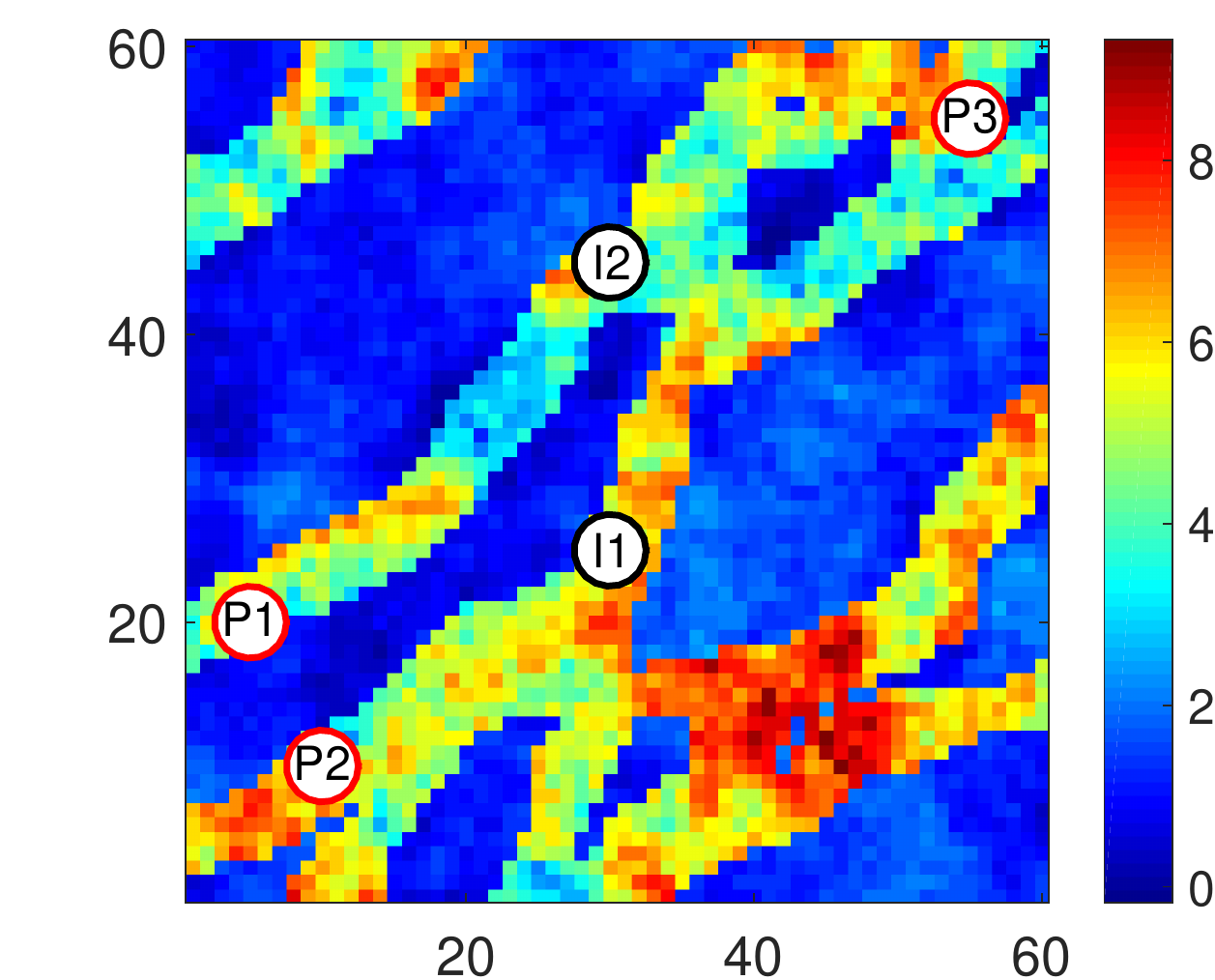}
\subcaption{Prior model~1}
\end{minipage}
\begin{minipage}{.4\linewidth}\centering
\includegraphics[trim = 100 250 150 250, clip, width=\linewidth]{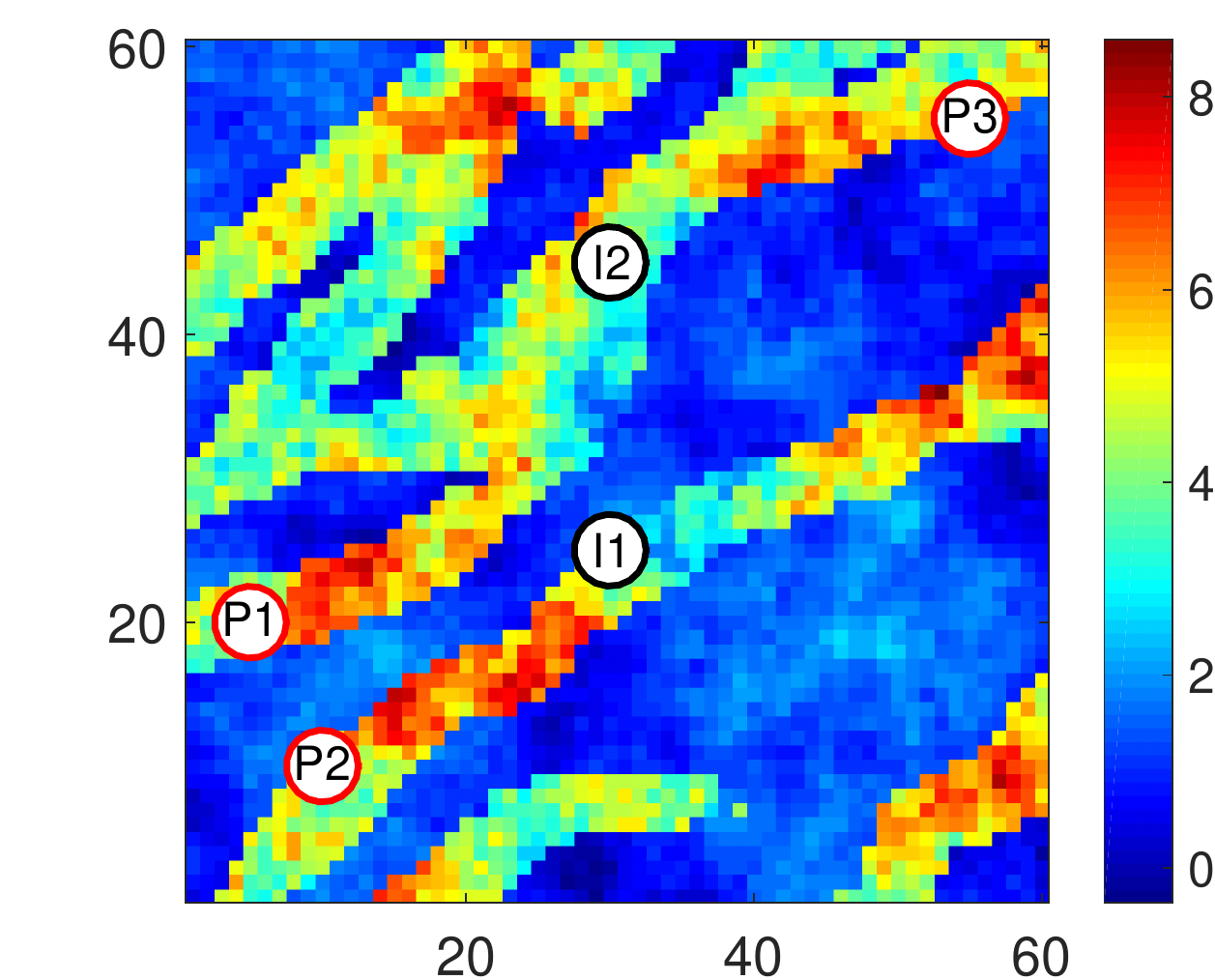}
\subcaption{Prior model~2}
\end{minipage}
\hspace{.05\linewidth}
\begin{minipage}{.4\linewidth}\centering
\includegraphics[trim = 100 250 150 250, clip, width=\linewidth]{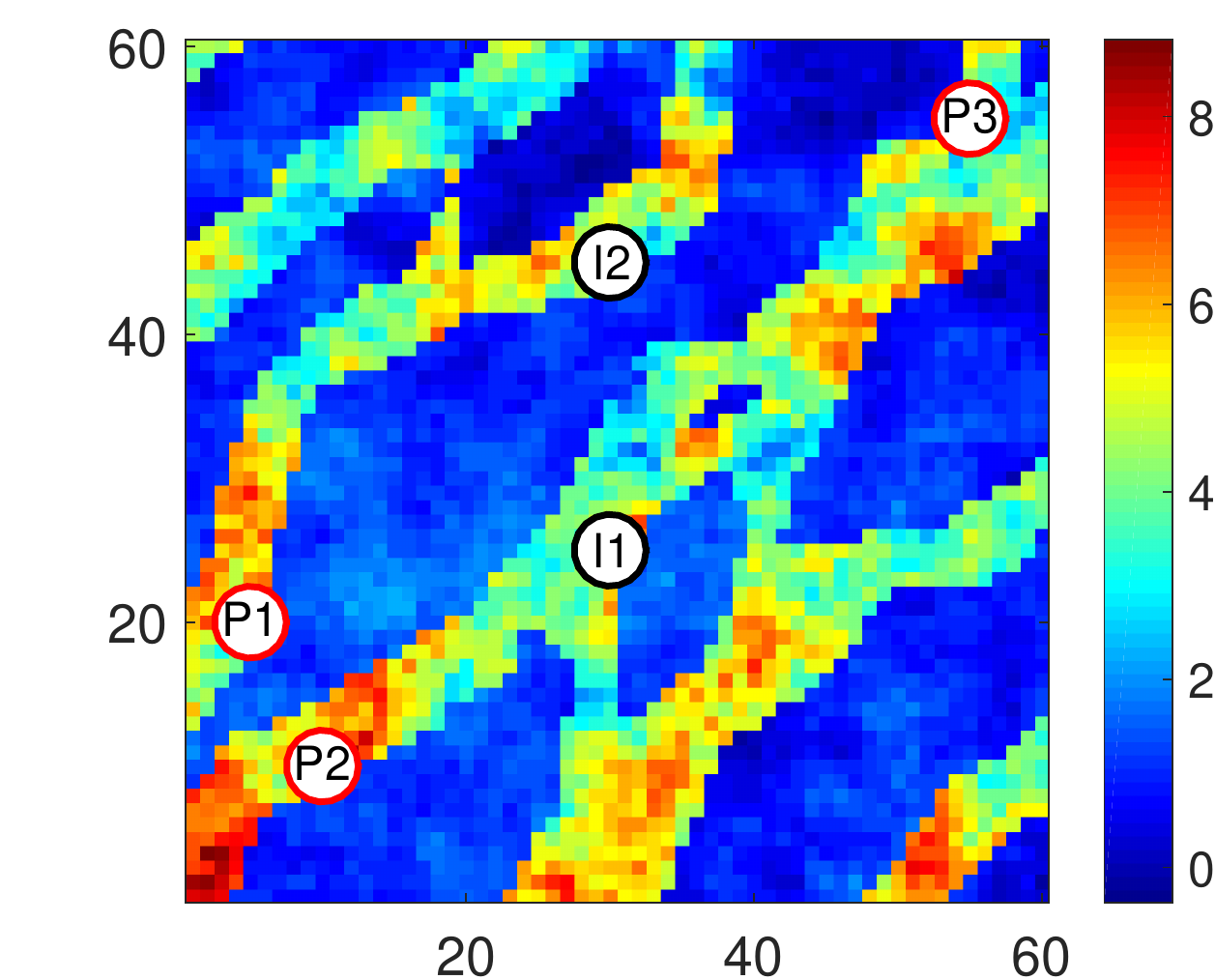}
\subcaption{Prior model~3}
\end{minipage}
\caption{Log-permeability realizations of 'true' and prior channelized models, all conditioned to hard data at well locations}\label{fig:perm}
\end{figure*}

The 2D synthetic bimodal channelized system considered here is the same as the system treated in earlier DSI studies~\citep{Sun2017,jiang2019data}. The channelized realizations are defined on a $60 \times 60$ grid. The size of each grid block is $25~\text{m} \times 25~\text{m} \times 10~\text{m}$. We use $\Nr = 800$ prior realizations. Figure~\ref{fig:perm} displays four prior realizations of log-permeability. Porosity is constant at 0.2 in all realizations. The channelized model includes sand and mud facies, with permeability variation in each facies -- thus the log-permeability is bimodal. The model shown in Fig.~\ref{fig:perm}(a) is taken to be the `true' case in most of the results that follow. There are five wells in the model (two injectors and three producers). These are all drilled in the sand, as is evident in Fig.~\ref{fig:perm}. The realizations are conditioned to the hard data from all five wells.

\begin{figure}[!ht]
\centering
\includegraphics[width = 0.47\textwidth]{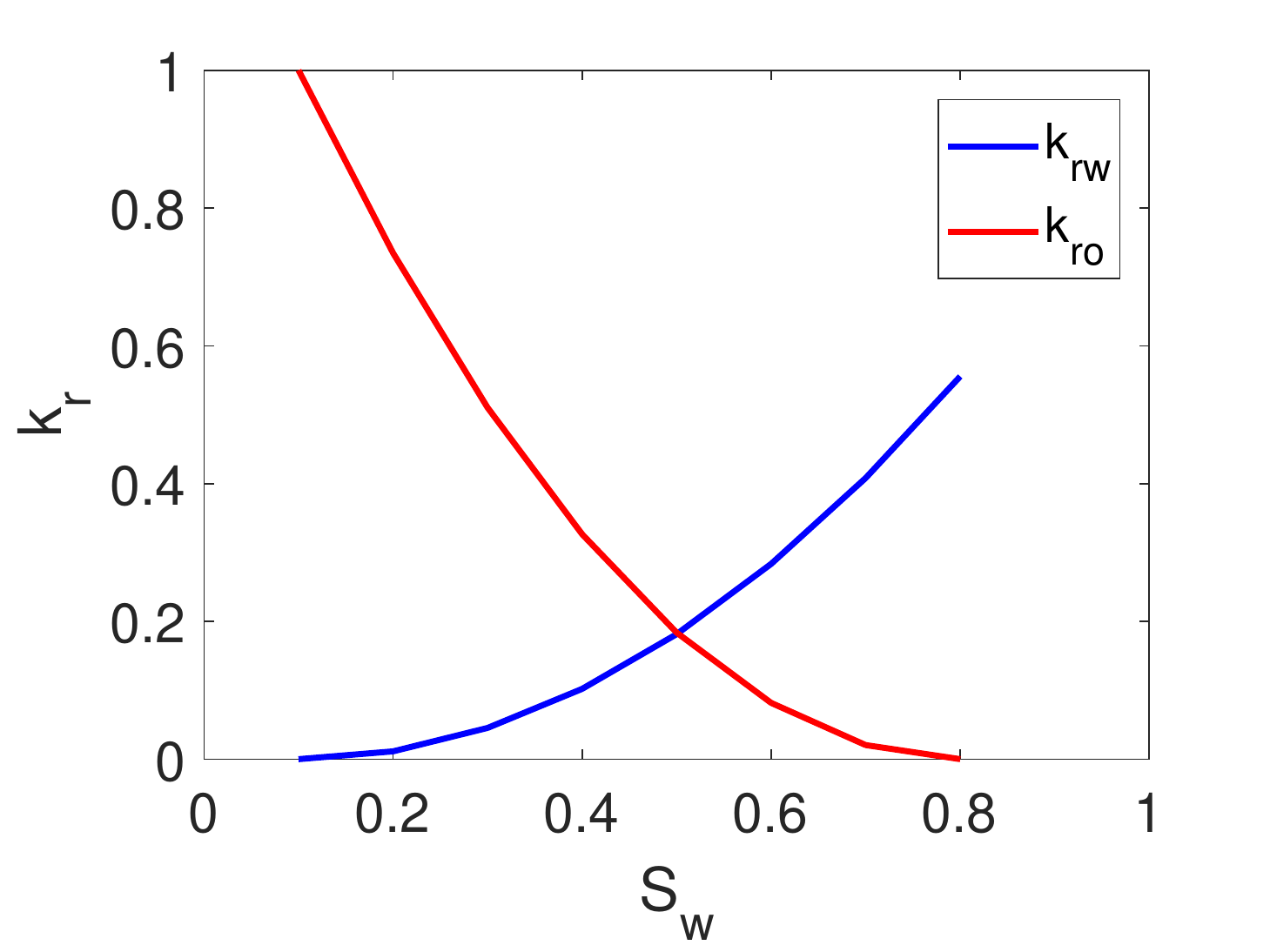}
\caption{Oil and water relative permeability curves}\label{fig:relaPerm}
\end{figure}

We consider a two-phase flow scenario. Water is injected into a formation containing a nonaqueous phase liquid (NAPL), referred to as oil. This setup could represent either the production of oil via water injection or a contaminant remediation operation. The initial in-situ phase saturations for oil and water are 0.9 and 0.1, respectively. The viscosities for oil and water are 1.16~cp and 0.31~cp. Figure~\ref{fig:relaPerm} displays the nonlinear relative permeability curves for the two phases. Capillary effects are neglected.

The five wells operate under specified (and constant-in-time) wellbore pressure. Wellbore pressure is typically specified at a particular depth (e.g., the uppermost location at which the well is open to flow), and this pressure is referred to as bottom-hole pressure (BHP). The wellbore pressure at other locations is impacted by gravitational effects and the (mixture) density of fluid in the wellbore, and must be computed by the simulator. The current example is however 2D, so the fixed wellbore pressures correspond to BHPs. This is not the case in the second example (which is 3D).

The well BHPs are specified as 550~bar for I1, 600~bar for I2, and 200~bar for all three producers. The flow simulations are run for 3000~days, with data collected every 30~days, which means $\Nt = 100$. Only a small subset of these data is used as observed data for history matching, as noted below. Simulation results are generated using AD-GPRS -- Stanford's Automatic Differentiation General Purpose Research Simulator~\cite{zhou2012parallel}. The simulation results include the water injection rate (WIR) for injectors I1 and I2, and the water production rate (WPR) and the oil production rate (OPR) for producers P1, P2 and P3. Thus we have eight quantities (time series) of interest; i.e., $\Nqoi=8$.

\subsection{DSI Results for Primary Quantities}

In this section we evaluate DSI posterior results, using a range of DSI treatments, for primary quantities of interest. By primary quantities we mean data variables directly included in the prior data vectors $\Bd_i$, $i=1, \ldots, \Nr$. These correspond to well-by-well water injection rate (WIR), water production rate (WPR), and oil production rate (OPR). To enable comparison with a rigorous rejection sampling method, discussed below, we consider a small amount of observed data -- specifically WIR in I1 and I2 and WPR and OPR in P3, at 180, 360 and 540~days. The number of observations $\Nhm$ is thus 12. We set the standard deviation of measurement error to be $10\%$ for all quantities.

Rejection sampling is a rigorous approach for sampling the posterior distribution that is used here to provide reference results~\cite{oliver2008inverse}. The method is very computationally intensive, however, and may become intractable when a large amount of observed data is considered (this is why we use $\Nhm=12$). In the RS process applied here, we use flow simulation results for $10^6$ realizations of the 2D channelized system (these results are from Sun and Durlofsky~\cite{Sun2017}). Models (and associated data variables) are rejected or accepted, with a probability that depends on the data mismatch relative to the true model. With the amount of data and data error considered here, we accept 117 models (out of the $10^6$ simulated), which are then used to represent the posterior results.

We will compare DSI posterior results to RS using four different DSI formulations. These entail (1) the direct application of ESMDA on the data vectors, without any parameterization, as applied by Lima et al.~\cite{lima2019data} (we refer to this method simply as ESMDA), (2) use of PCA with histogram transformation for parameterization and RML for posterior sampling (PCA+HT+RML), as applied by Sun et al.~\cite{SunCG}, (3) use of PCA with histogram transformation for parameterization and ESMDA for posterior sampling (PCA+HT+ ESMDA), and (4) application of our new DSI framework that uses RAE for parameterization and ESMDA for posterior sampling (RAE+ESMDA). 

\begin{figure*}[!ht]
\centering
\begin{minipage}{.4\linewidth}\centering
\includegraphics[width=\linewidth]{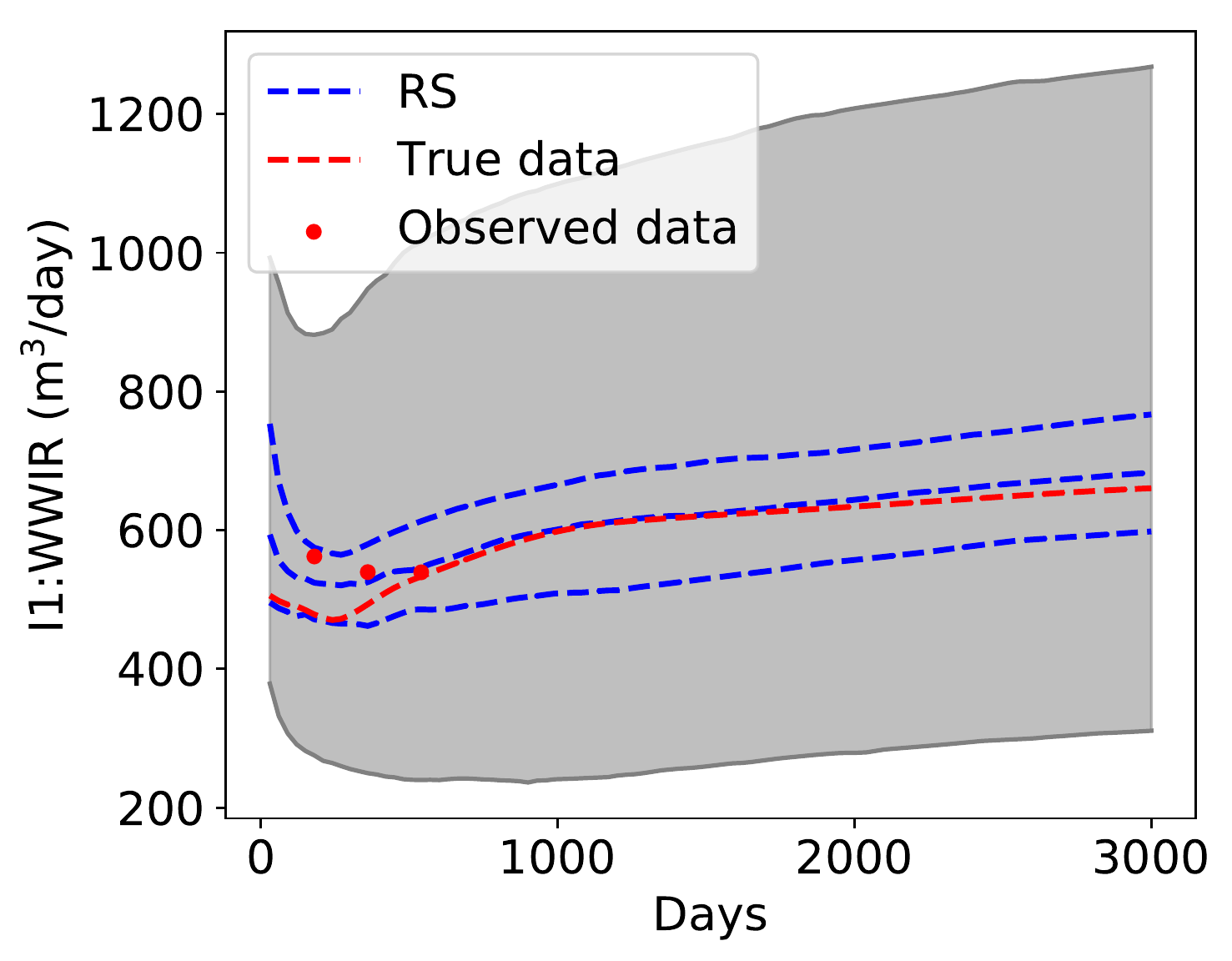}
\subcaption{I1 WIR}
\end{minipage}
\hspace{.05\linewidth}
\begin{minipage}{.4\linewidth}\centering
\includegraphics[width=\linewidth]{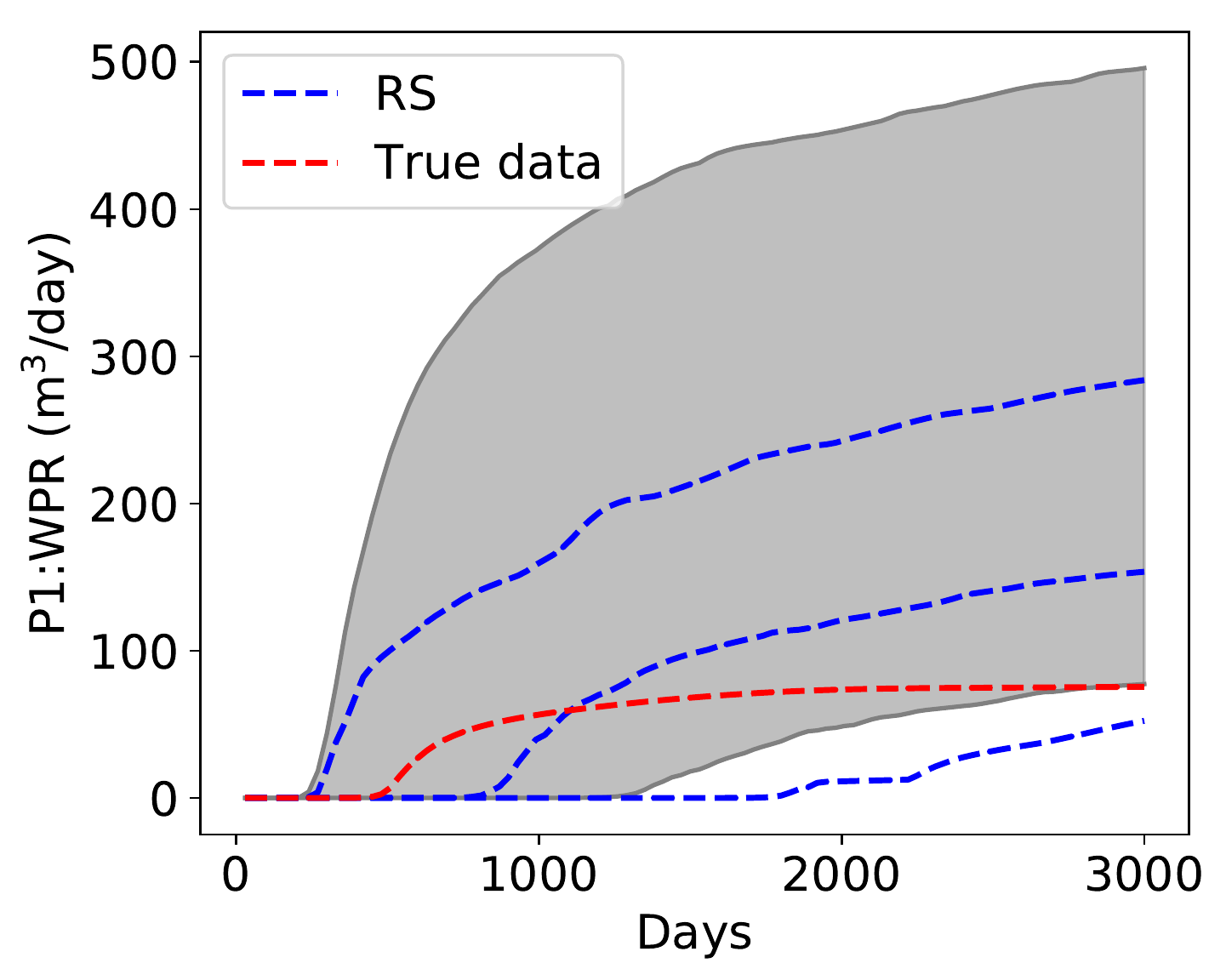}
\subcaption{P1 WPR}
\end{minipage}
\begin{minipage}{.4\linewidth}\centering
\includegraphics[width=\linewidth]{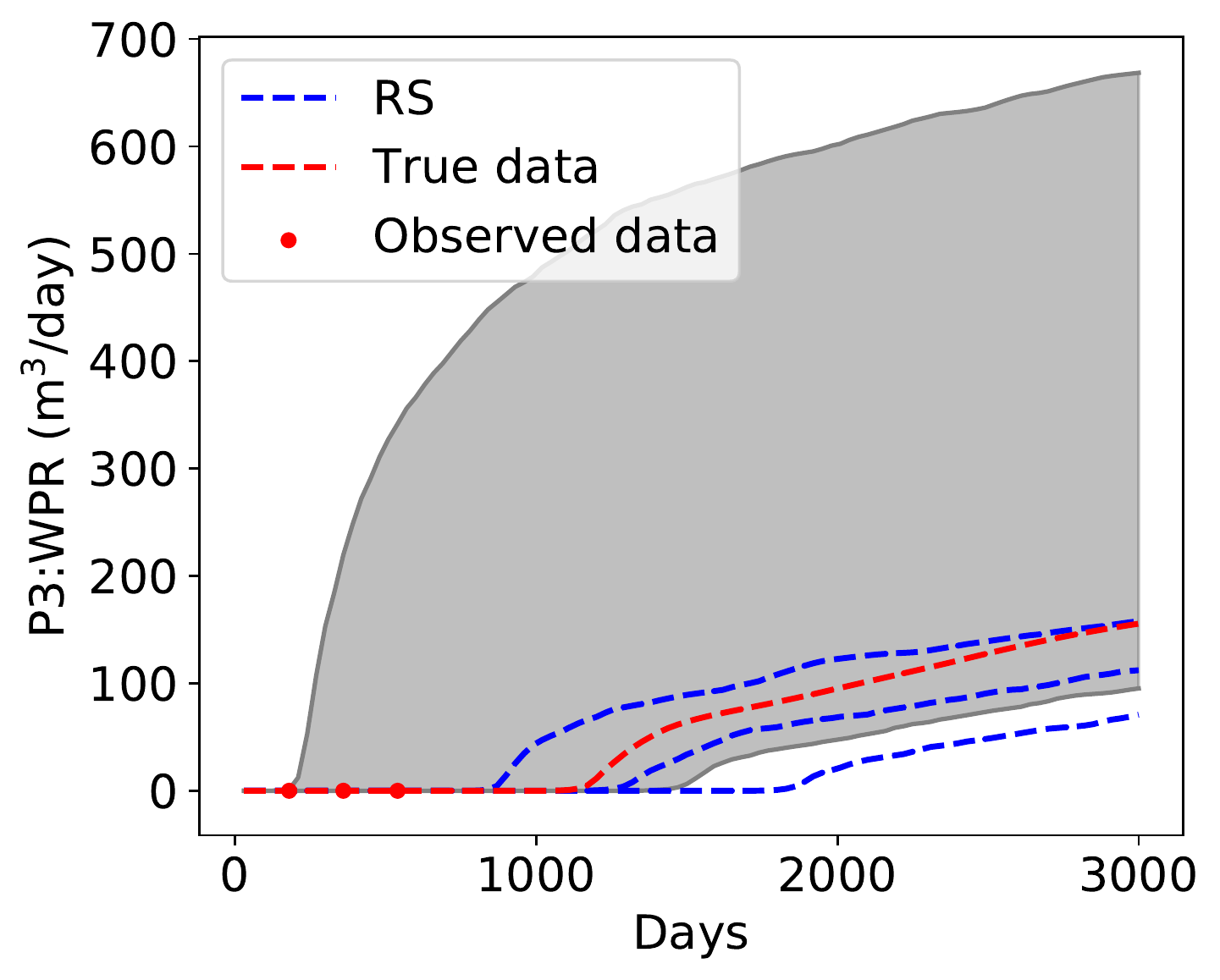}
\subcaption{P3 WPR}
\end{minipage}
\hspace{.05\linewidth}
\begin{minipage}{.4\linewidth}\centering
\includegraphics[width=\linewidth]{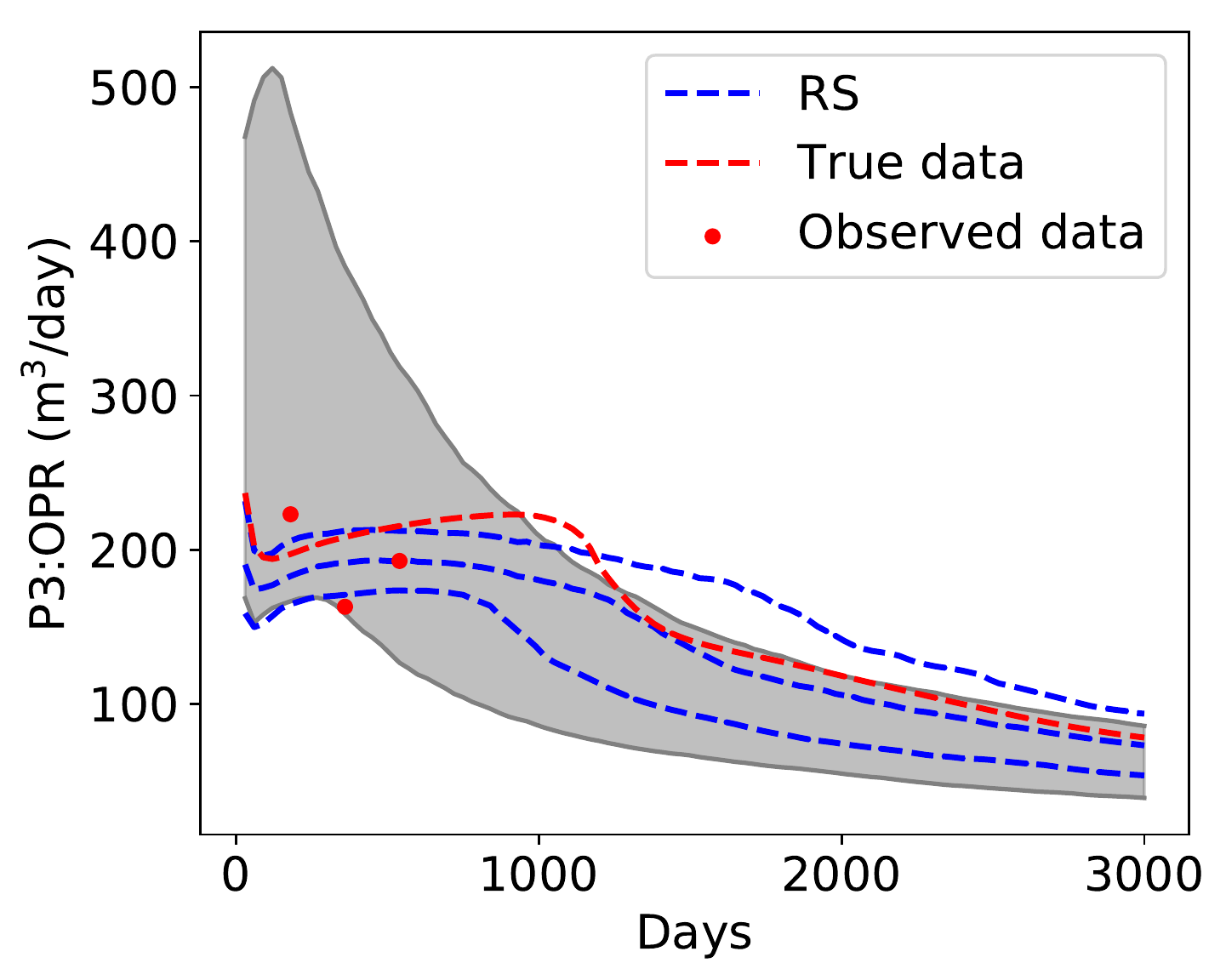}
\subcaption{P3 OPR}
\end{minipage}
\caption{Prior simulation results and reference RS posterior results. The gray shaded area shows the P$_{10}$-P$_{90}$ range of the prior simulation results}\label{fig:prior_rs_2d}
\end{figure*}

We first display, in Fig.~\ref{fig:prior_rs_2d}, prior simulation results along with RS posterior results for I1 WIR, P1 WPR, and P3 WPR and OPR. The large gray shaded regions in each plot indicate the P$_{10}$-P$_{90}$ interval of the prior data. By P$_{10}$ we mean the tenth percentile value of the data at that time step (and similarly for P$_{90}$ and P$_{50}$). Note that different models may correspond to the P$_{10}$ result at different time steps. The true data (simulation results for the true model) are depicted by the red dashed lines. The red points show the observed data, generated by adding measurement error to the true data during the history match period. The P$_{10}$, P$_{50}$ and P$_{90}$ posterior results from RS are indicated by the lower, middle and upper blue dashed lines, respectively. A large amount of uncertainty reduction is observed in the RS results (i.e., the area spanned by the blue curves is much less than that in the gray region). Note that, even though there are no observations for P1 WPR (Fig.~\ref{fig:prior_rs_2d}(b)), uncertainty reduction is still accomplished due to correlations between different components of the data vector.

\begin{figure*}[!htpb]
\centering
\begin{minipage}{.4\linewidth}\centering
\includegraphics[width=\linewidth]{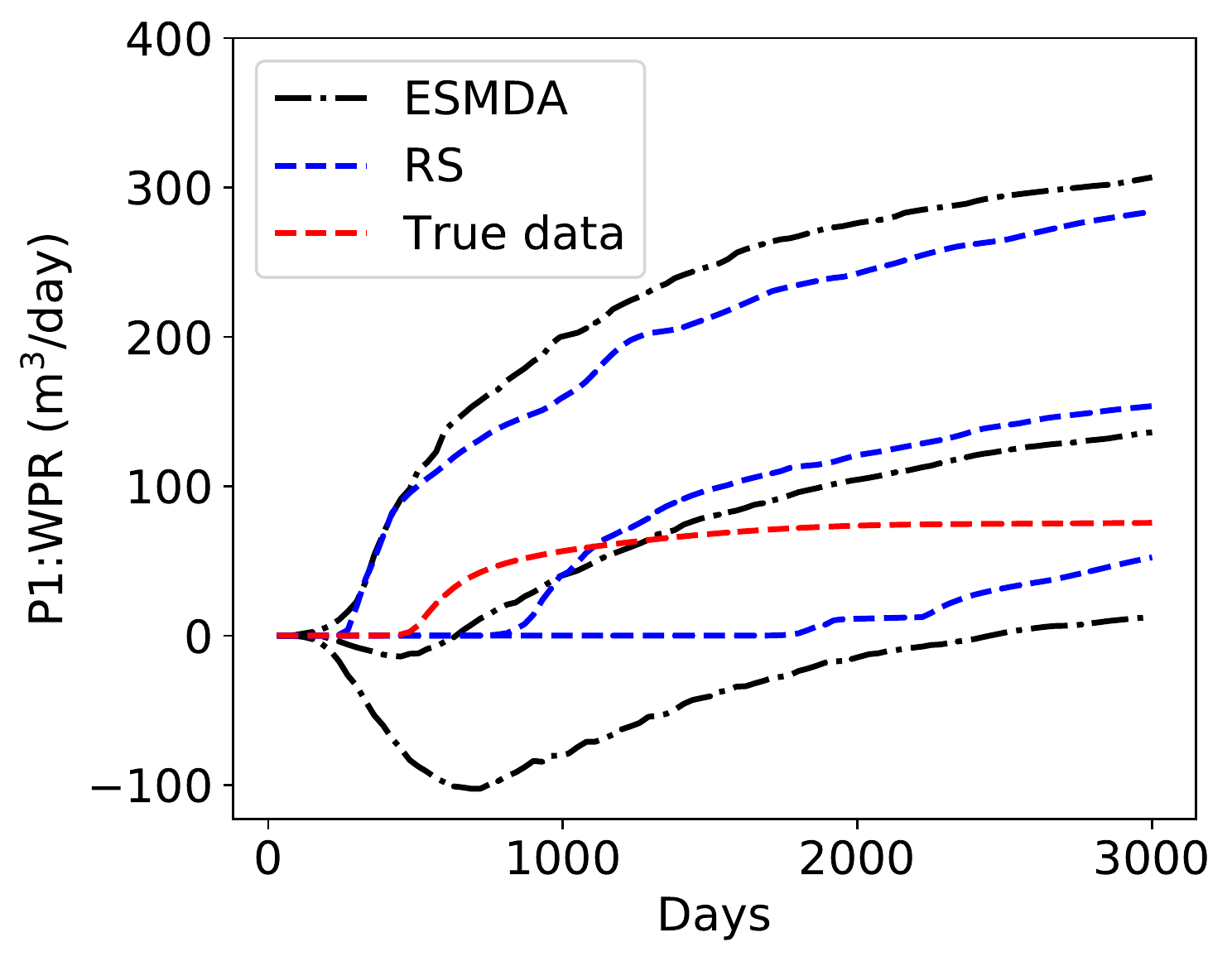}
\subcaption{P1 WPR}
\end{minipage}
\hspace{.05\linewidth}
\begin{minipage}{.4\linewidth}\centering
\includegraphics[width=\linewidth]{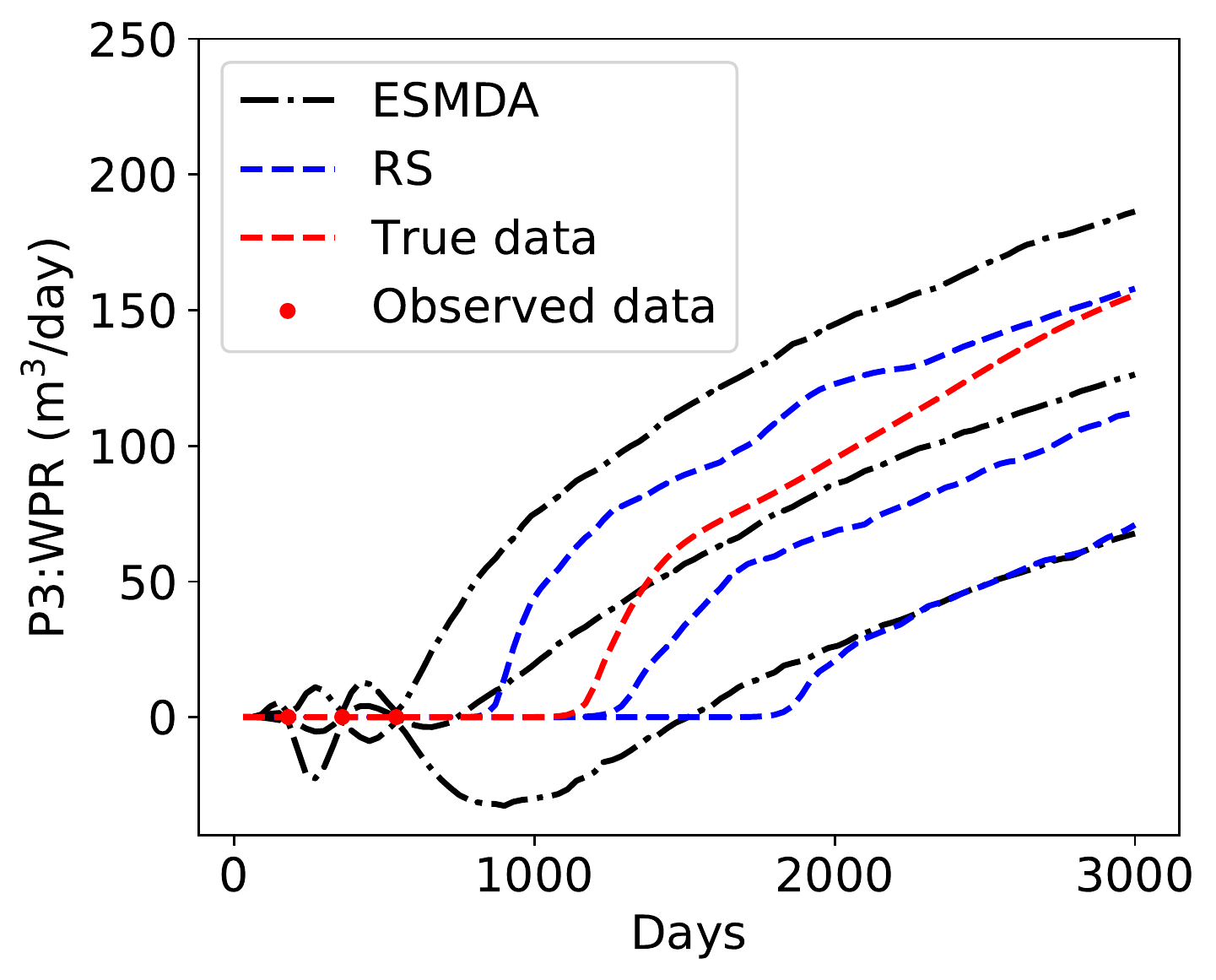}
\subcaption{P3 WPR}
\end{minipage}
\caption{ESMDA (in DSI) posterior forecasting results without truncation}\label{fig:post_esmda}
\end{figure*}

We now assess the performance of the four different DSI implementations. We first consider the use of ESMDA without data parameterization. We use $\Nr = 800$ prior realizations and assimilate data four times ($\Na = 4$). The black dash-dot lines in Fig.~\ref{fig:post_esmda} indicate the P$_{10}$, P$_{50}$ and P$_{90}$ posterior results using ESMDA without any post-processing (i.e., without truncation). The reference RS results appear as the blue dashed lines. Note that the prior results are not shown in this figure, and the data ranges are different than those in Fig.~\ref{fig:prior_rs_2d}. It is evident that the ESMDA results are unphysical in some cases -- e.g., the P$_{10}$ curves for both wells indicate negative production rates over much of the simulation. If we `truncate' negative rates to zero, as suggested by Lima et al.~\cite{lima2019data}, then the ESMDA results agree reasonably well with the RS results (though there are still some unphysical fluctuations at early time in the P3 WPR results).

\begin{figure*}[!ht]
\centering
\begin{minipage}{.4\linewidth}\centering
\includegraphics[width=\linewidth]{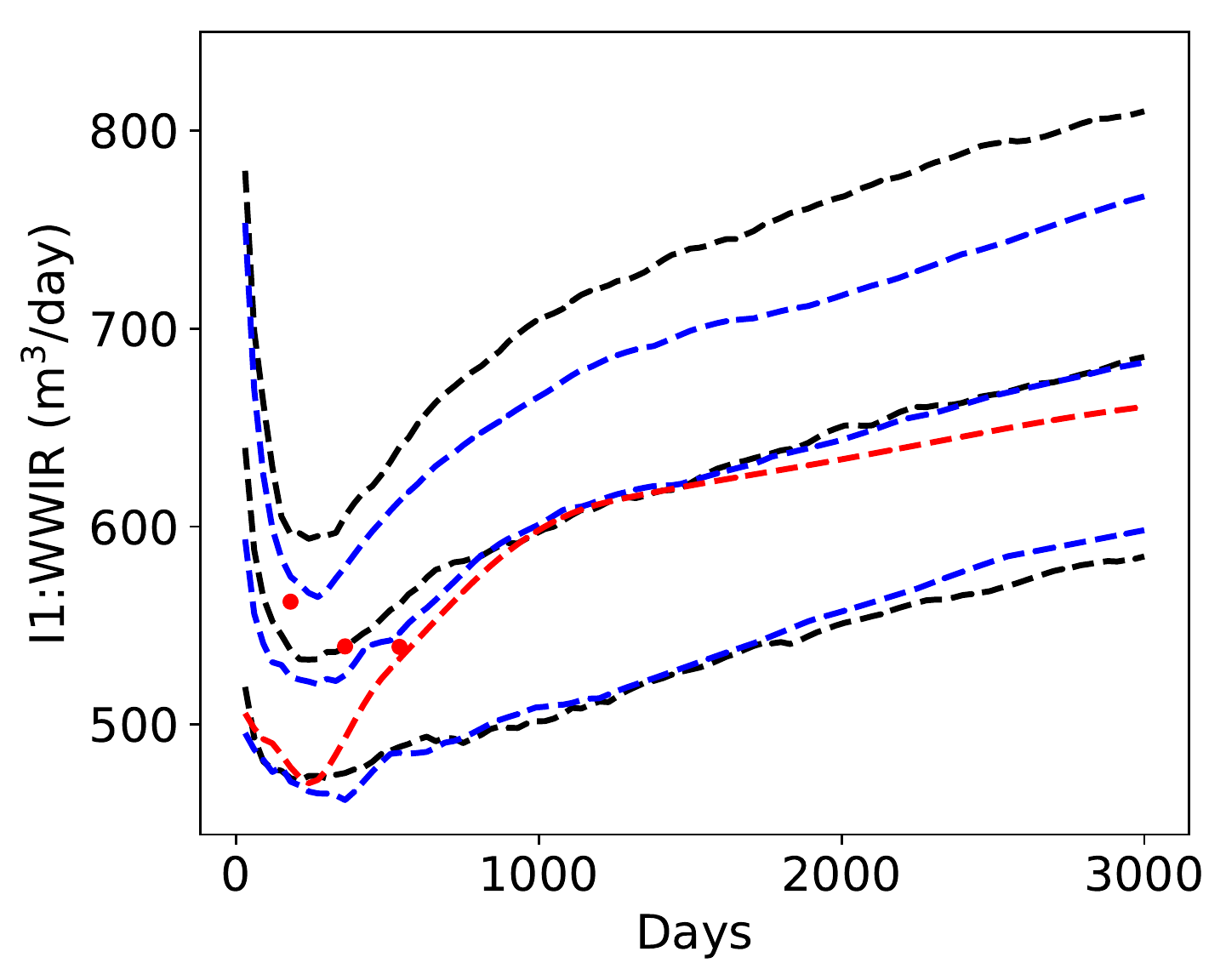}
\subcaption{I1 WIR}
\end{minipage}
\hspace{.05\linewidth}
\begin{minipage}{.4\linewidth}\centering
\includegraphics[width=\linewidth]{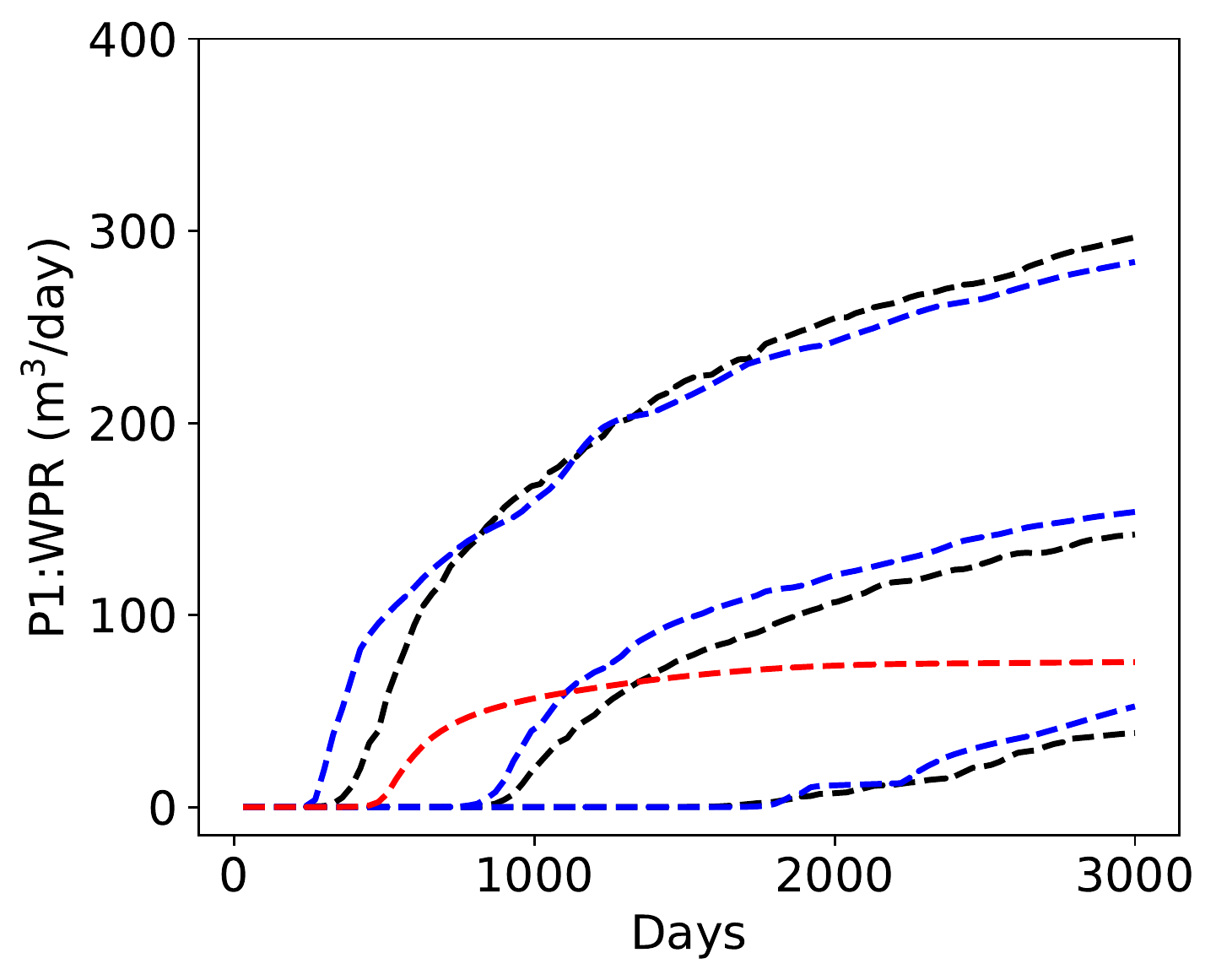}
\subcaption{P1 WPR}
\end{minipage}
\begin{minipage}{.4\linewidth}\centering
\includegraphics[width=\linewidth]{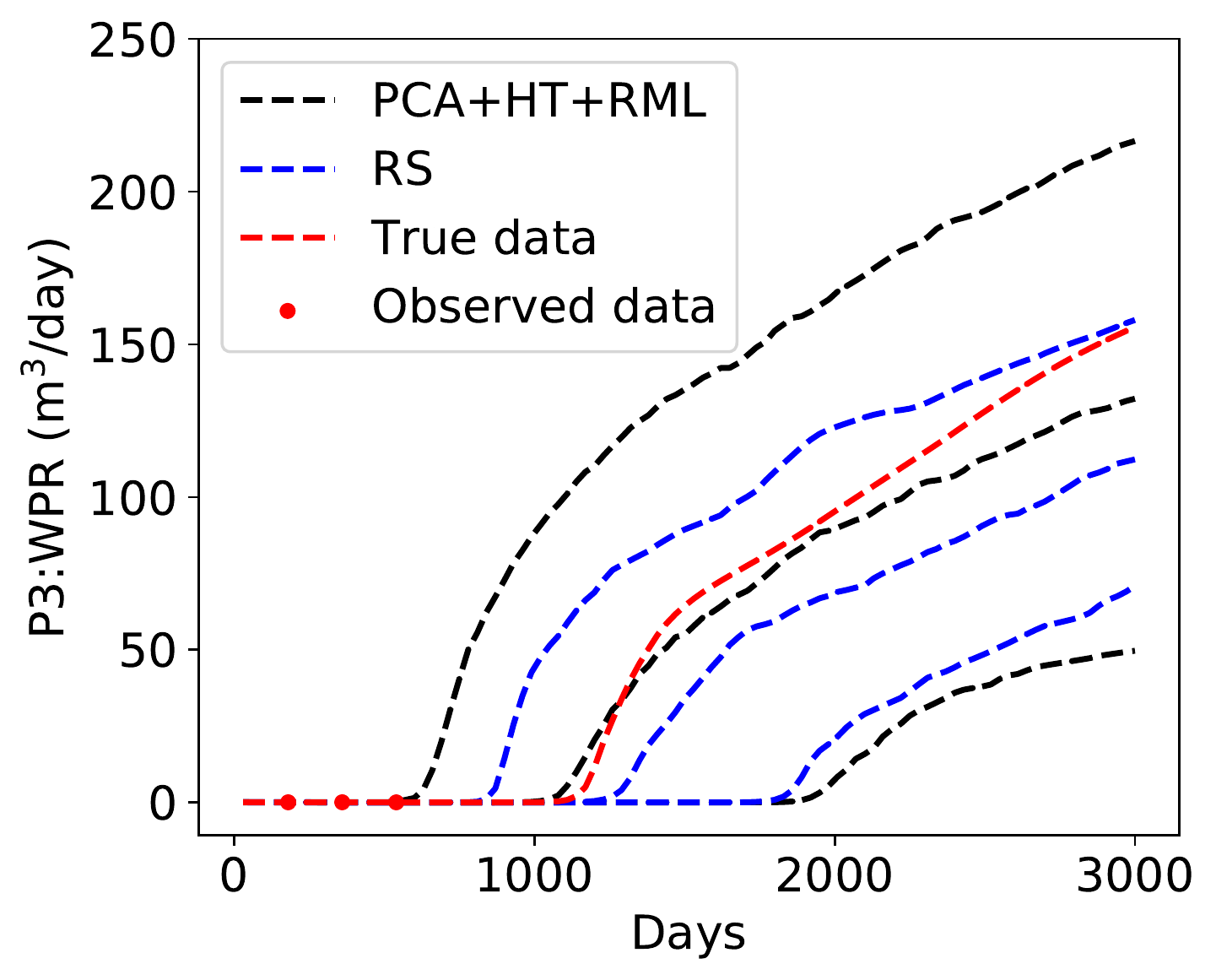}
\subcaption{P3 WPR}
\end{minipage}
\hspace{.05\linewidth}
\begin{minipage}{.4\linewidth}\centering
\includegraphics[width=\linewidth]{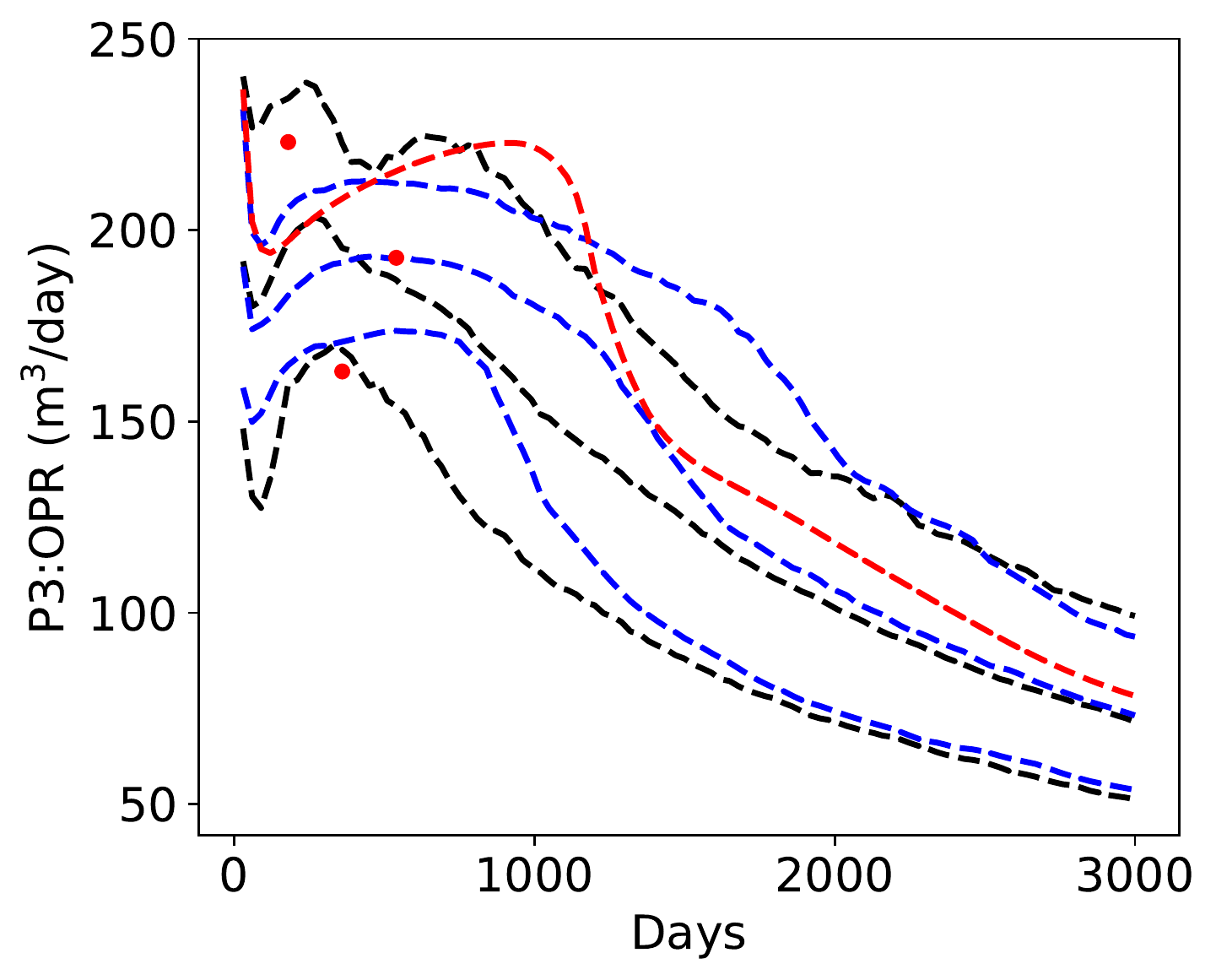}
\subcaption{P3 OPR}
\end{minipage}
\caption{PCA+HT+RML DSI posterior results. Legend in (c) applies to all subplots}\label{fig:post_DSI_only}
\end{figure*}

The next DSI formulation considered entails the use of PCA with histogram transformation for parameterization combined with RML for sampling (PCA+HT+RML).
We use $\Nl=31$ latent variables, based on application of an energy criterion. A total of 800 posterior samples are again generated, and results are shown in Fig.~\ref{fig:post_DSI_only}. The DSI results in this case are consistently physical, and they are again in reasonable agreement with the RS results. There are some discrepancies, however, particularly in the P$_{90}$ results for I1 WIR and P3 WPR. 

\begin{figure*}[!ht]
\centering
\begin{minipage}{.4\linewidth}\centering
\includegraphics[width=\linewidth]{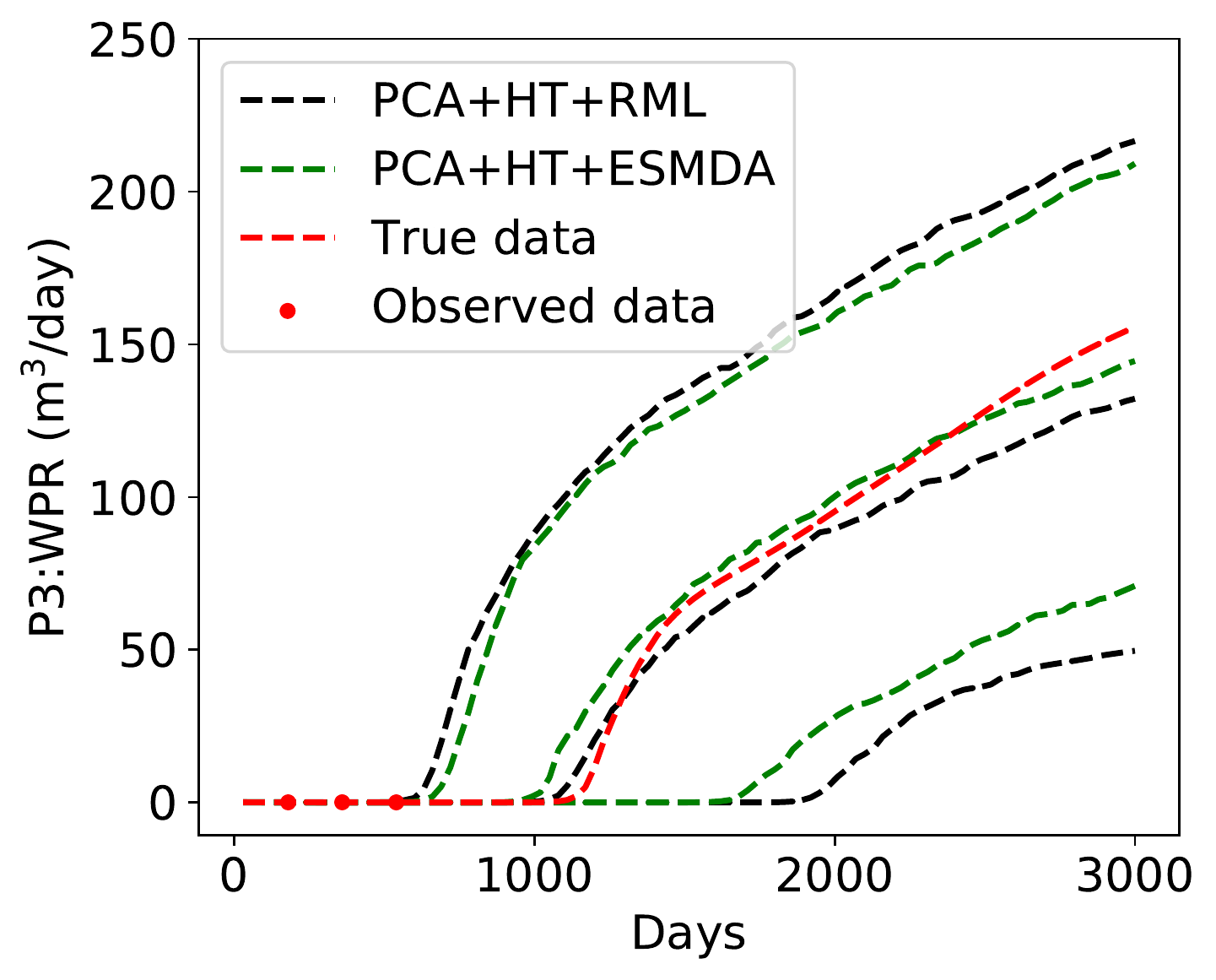}
\subcaption{P3 WPR}
\end{minipage}
\hspace{.05\linewidth}
\begin{minipage}{.4\linewidth}\centering
\includegraphics[width=\linewidth]{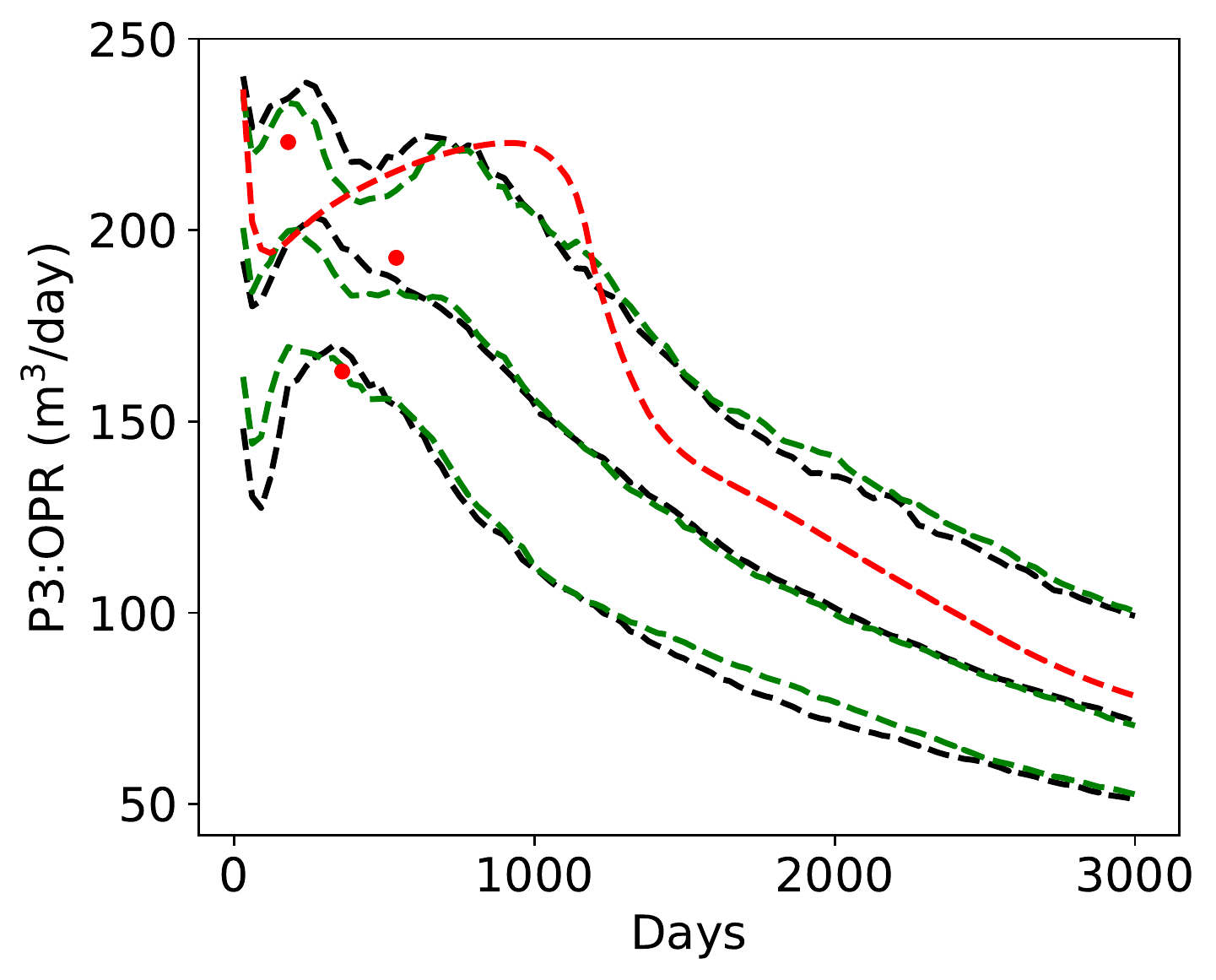}
\subcaption{P3 OPR}
\end{minipage}
\caption{PCA+HT+RML and PCA+HT+ESMDA posterior DSI results. Legend in (a) applies to both subplots}\label{fig:post_rml_esmda}
\end{figure*}

We next compare DSI results using PCA+HT+ESMDA to those using PCA+HT+RML. Thus the methods differ only in how the posterior sampling is accomplished. From the comparisons shown in Fig.~\ref{fig:post_rml_esmda}, for P3 water and oil production rates, we see that the P$_{10}$, P$_{50}$ and P$_{90}$ posterior predictions from the two sampling methods agree closely. This suggests that either method could be used for this case. Note that we have not observed ensemble collapse, here or for the 3D example in Section~\ref{sec:3D_case}, using ESMDA. This may be related to the very limited amount of data considered in these cases.

\begin{figure*}[!ht]
\centering
\begin{minipage}{.4\linewidth}\centering
\includegraphics[width=\linewidth]{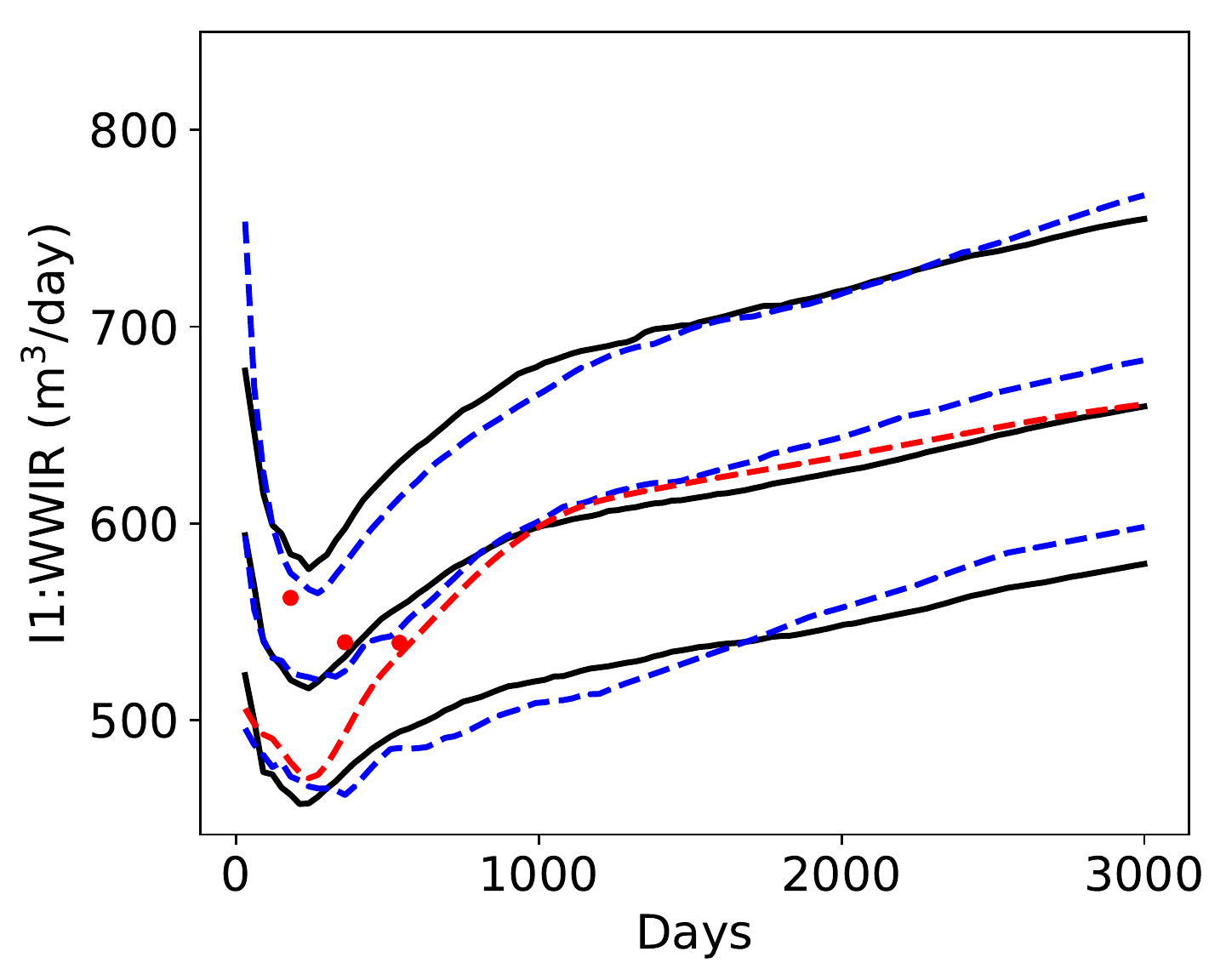}
\subcaption{I1 WIR}
\end{minipage}
\hspace{.05\linewidth}
\begin{minipage}{.4\linewidth}\centering
\includegraphics[width=\linewidth]{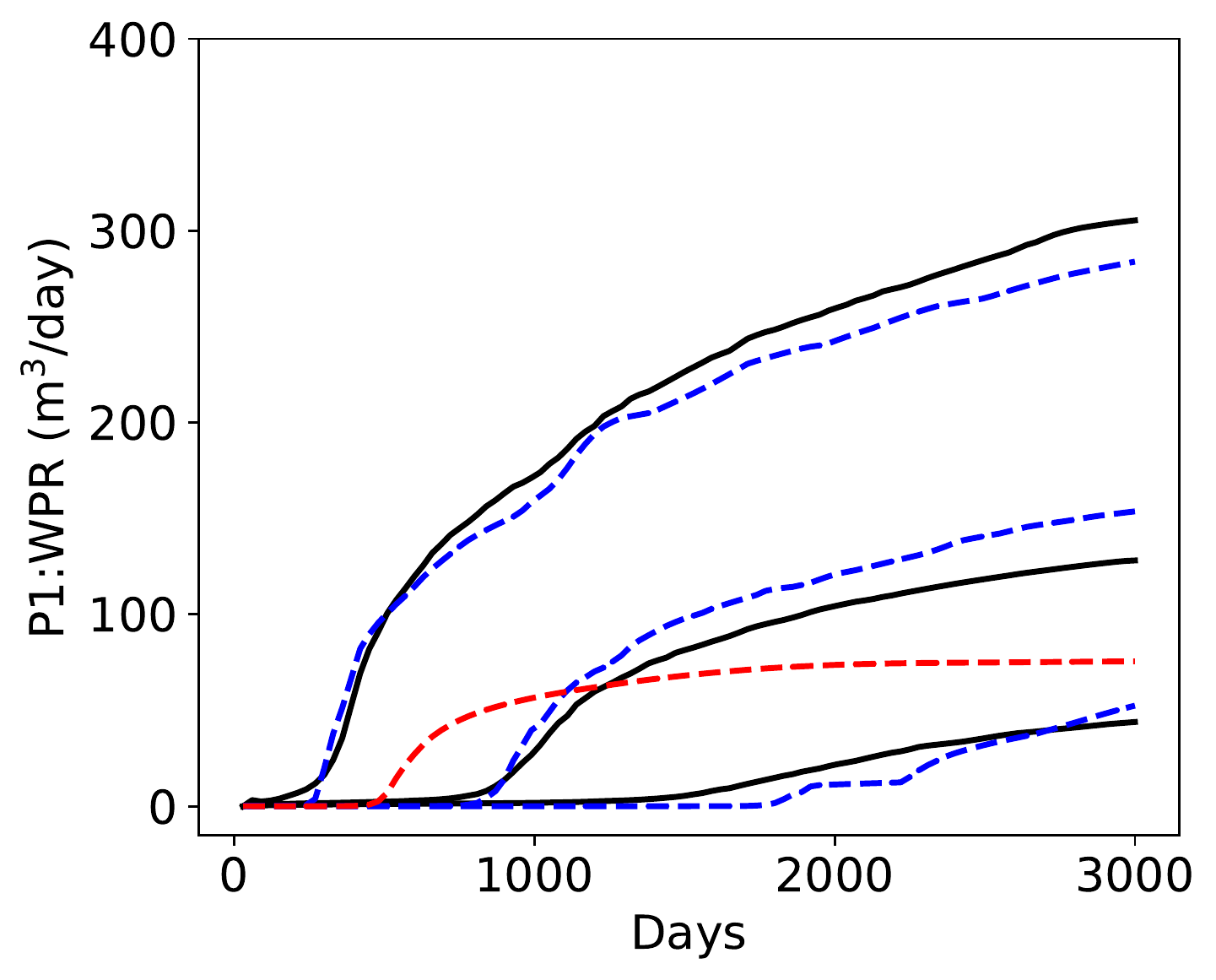}
\subcaption{P1 WPR}
\end{minipage}
\begin{minipage}{.4\linewidth}\centering
\includegraphics[width=\linewidth]{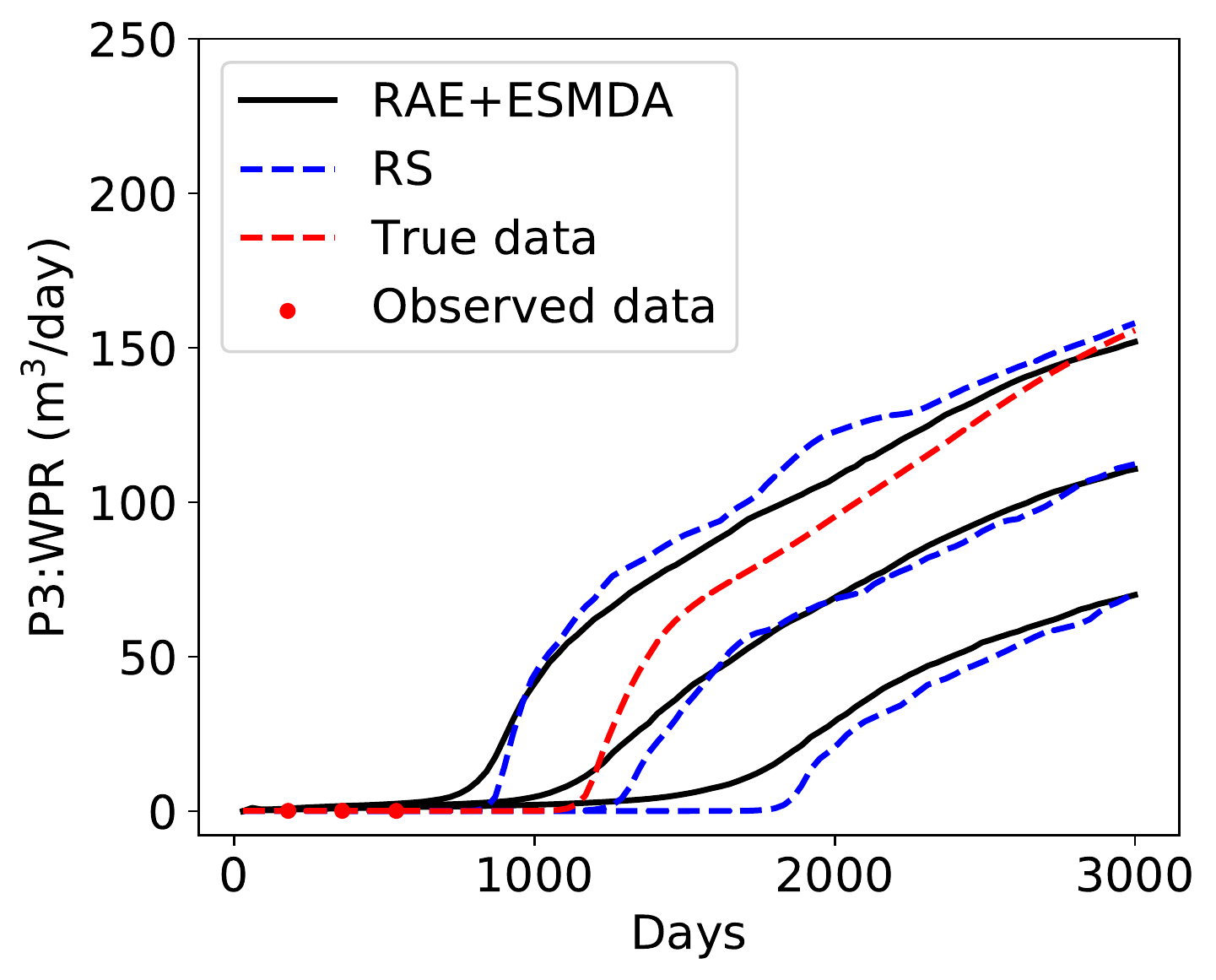}
\subcaption{P3 WPR}
\end{minipage}
\hspace{.05\linewidth}
\begin{minipage}{.4\linewidth}\centering
\includegraphics[width=\linewidth]{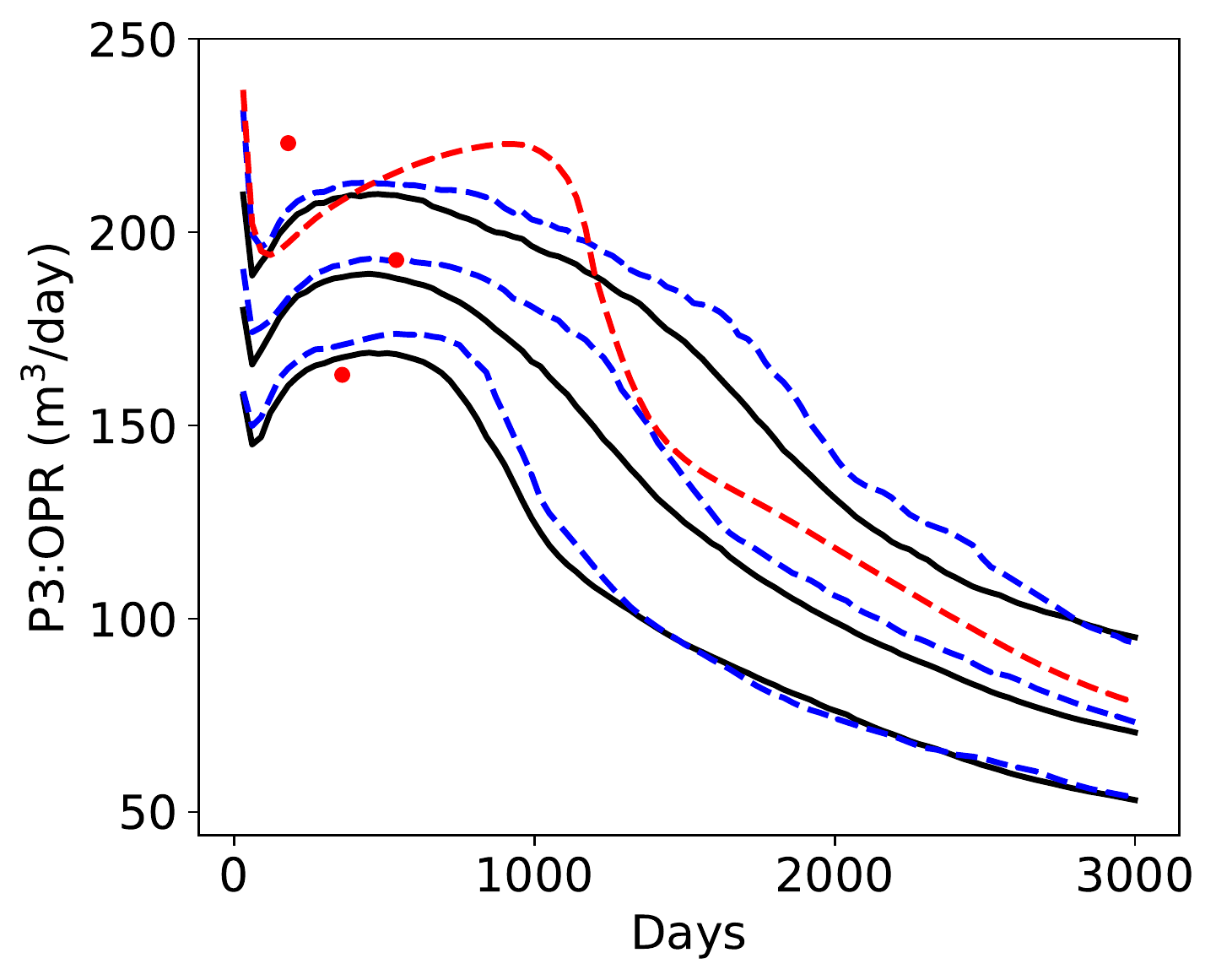}
\subcaption{P3 OPR}
\end{minipage}
\caption{RAE+ESMDA posterior DSI results. Legend in (c) applies to all subplots}\label{fig:post_rae}
\end{figure*}

Finally, we apply the DSI framework with the RAE data parameterization method and ESMDA (RAE+ESMDA). In this case the training set includes $\Nr = 800$ realizations and we specify $\Nl = 31$; i.e., the dimension of the latent variable is equal that used with PCA. We again use $\Na=4$ for ESMDA. Figure~\ref{fig:post_rae} presents posterior DSI results using the RAE+ESMDA procedure. The P$_{10}$, P$_{50}$ and P$_{90}$ posterior forecasts agree reasonably closely with the reference RS results. In fact, the level of agreement in Fig.~\ref{fig:post_rae} is better than that observed for the other three DSI formulations considered. This highlights the impact of accurate data parameterization in DSI settings.

Thus far we have considered DSI predictions for `primary' quantities of interest. By this we mean that the quantity is included in the prior data vector. For such variables, RAE+ESMDA outperforms the other methods in terms of error relative to the RS results, indicating the benefit of RAE compared to PCA+HT for parameterization. These advantages stem from the fact that RAE can represent nonlinear features that are not captured by PCA. All of the DSI procedures, however, were able to provide results of acceptable accuracy in the sense that error was relatively small compared to the amount of uncertainty reduction achieved. In the next section we will consider the prediction of more challenging (derived) quantities, and the differences in performance between the parameterizations will be larger.

\subsection{DSI Results for Derived Quantities and Correlations}

In this section we consider predictions of derived quantities and correlations between quantities. Derived quantities, defined as results computed from two or more primary quantities, do not appear directly in the DSI data variable $\Bd$ or as observed data. Many of the time-varying quantities of interest in practical settings are, indeed, derived. These include the total liquid production rate for a producer (sum of water and oil production rates), the water `cut' for a producer (water rate divided by total liquid rate), field-wide injection rate (sum of injection rates over all injectors), along with analogous field-wide quantities for water and oil production. Reliable prediction of derived quantities requires that the correlations between primary quantities be captured correctly -- otherwise these combined variables may be inaccurate or unphysical.

\begin{figure*}[!htb]
\centering
\begin{minipage}{.4\linewidth}\centering
\includegraphics[width=\linewidth]{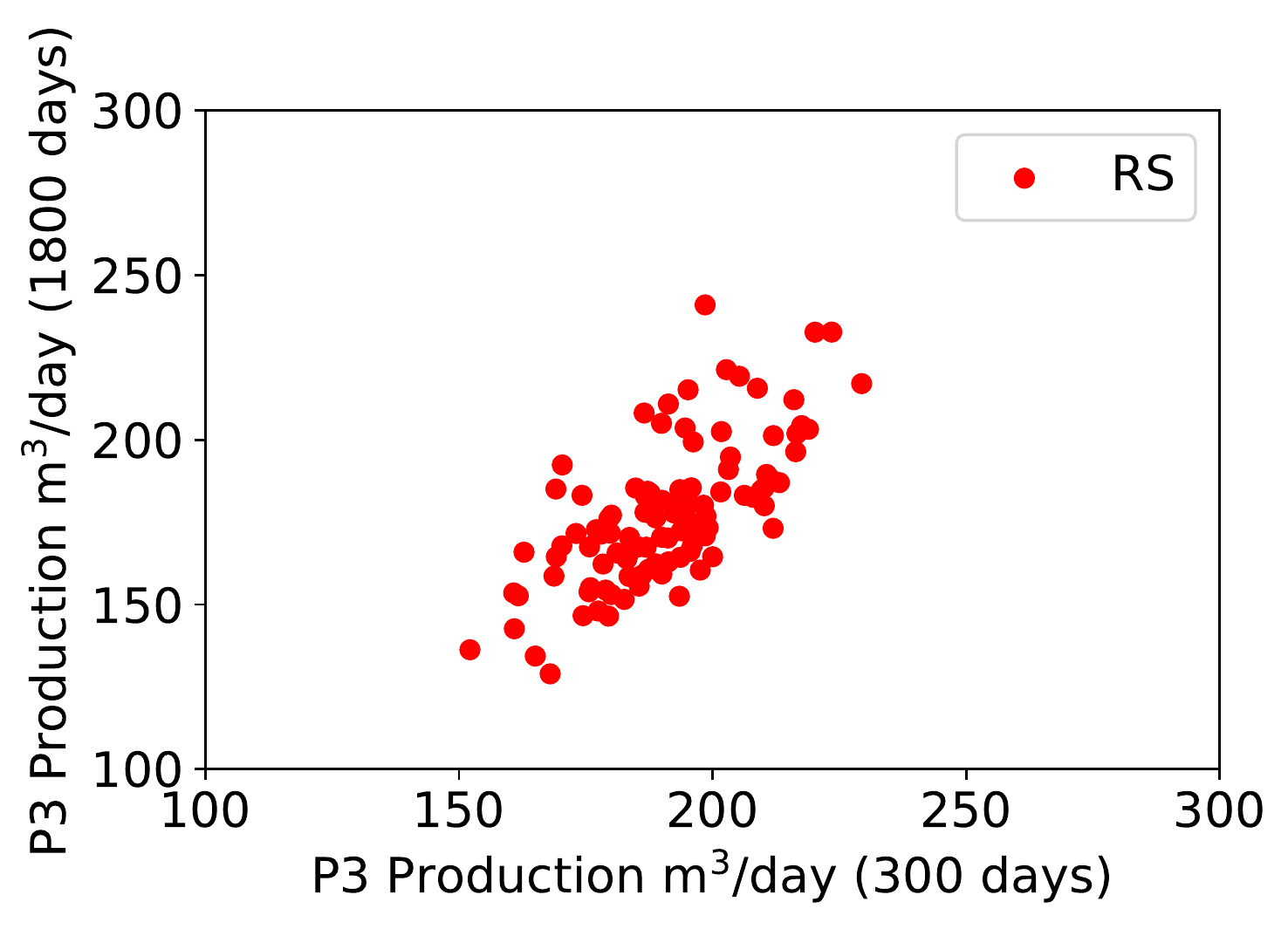}
\subcaption{RS}
\end{minipage}
\hspace{.05\linewidth}
\begin{minipage}{.4\linewidth}\centering
\includegraphics[width=\linewidth]{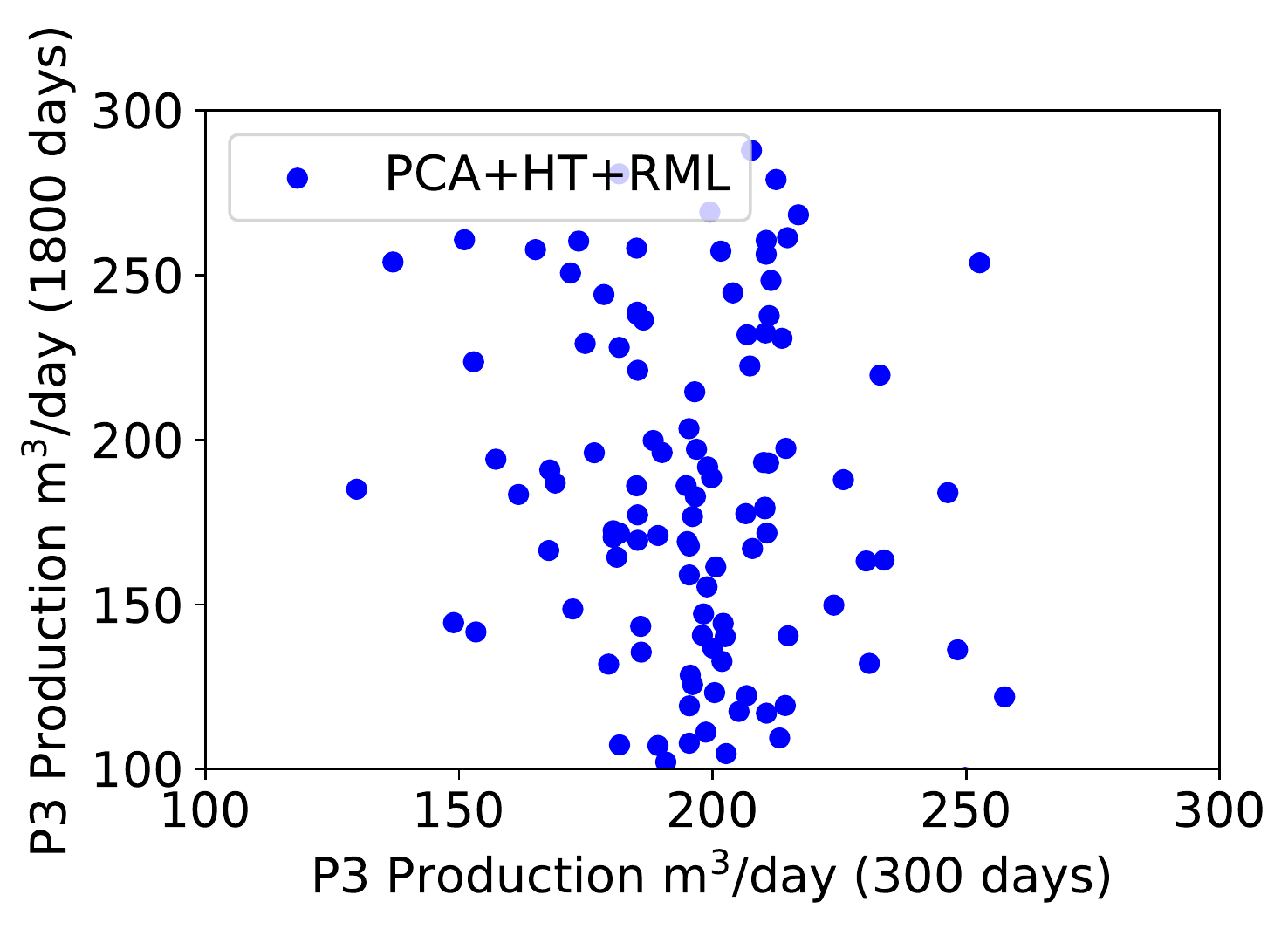}
\subcaption{PCA+HT+RML}
\end{minipage}
\begin{minipage}{.4\linewidth}\centering
\includegraphics[width=\linewidth]{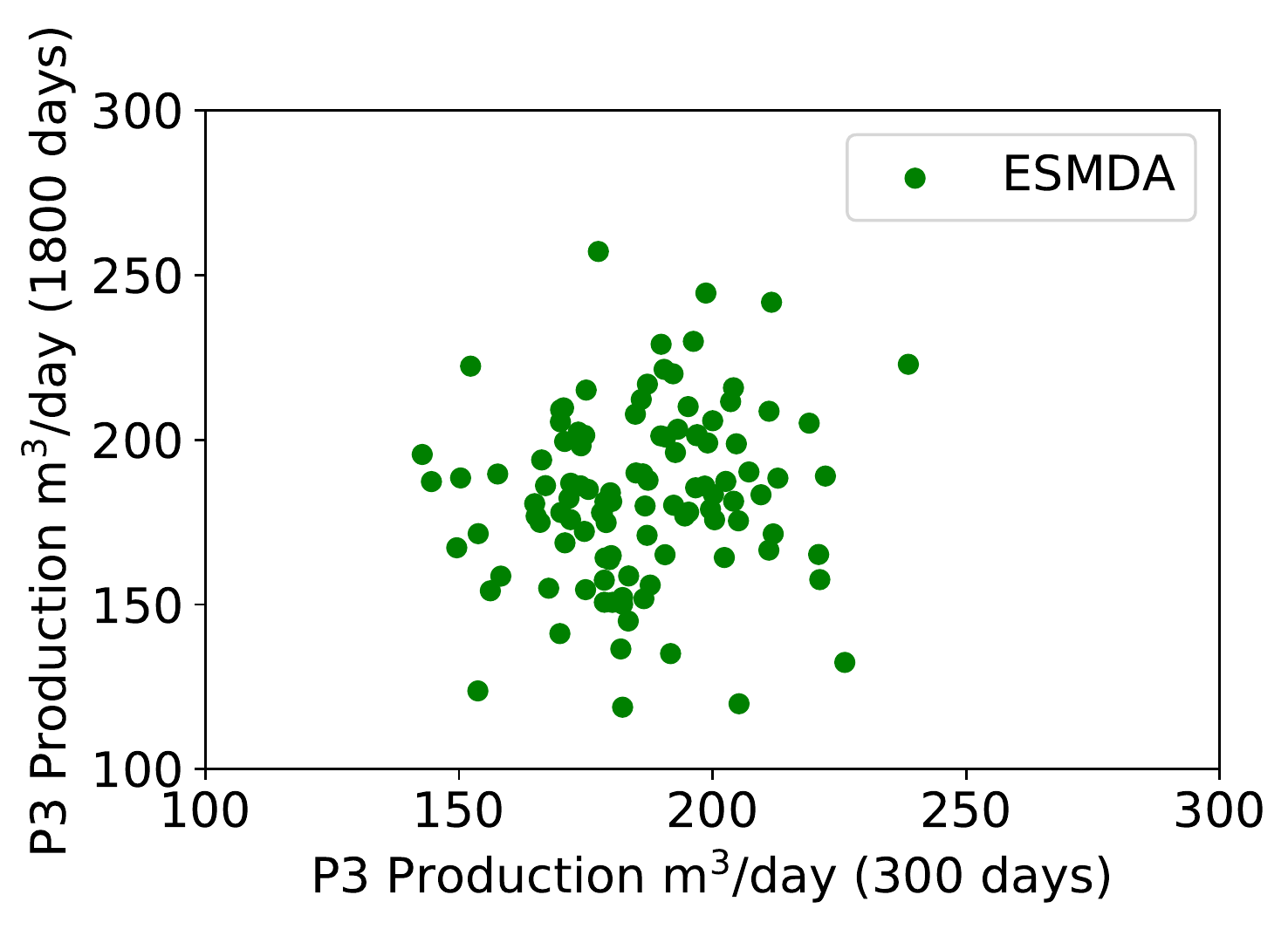}
\subcaption{ESMDA (with truncation)}
\end{minipage}
\hspace{.05\linewidth}
\begin{minipage}{.4\linewidth}\centering
\includegraphics[width=\linewidth]{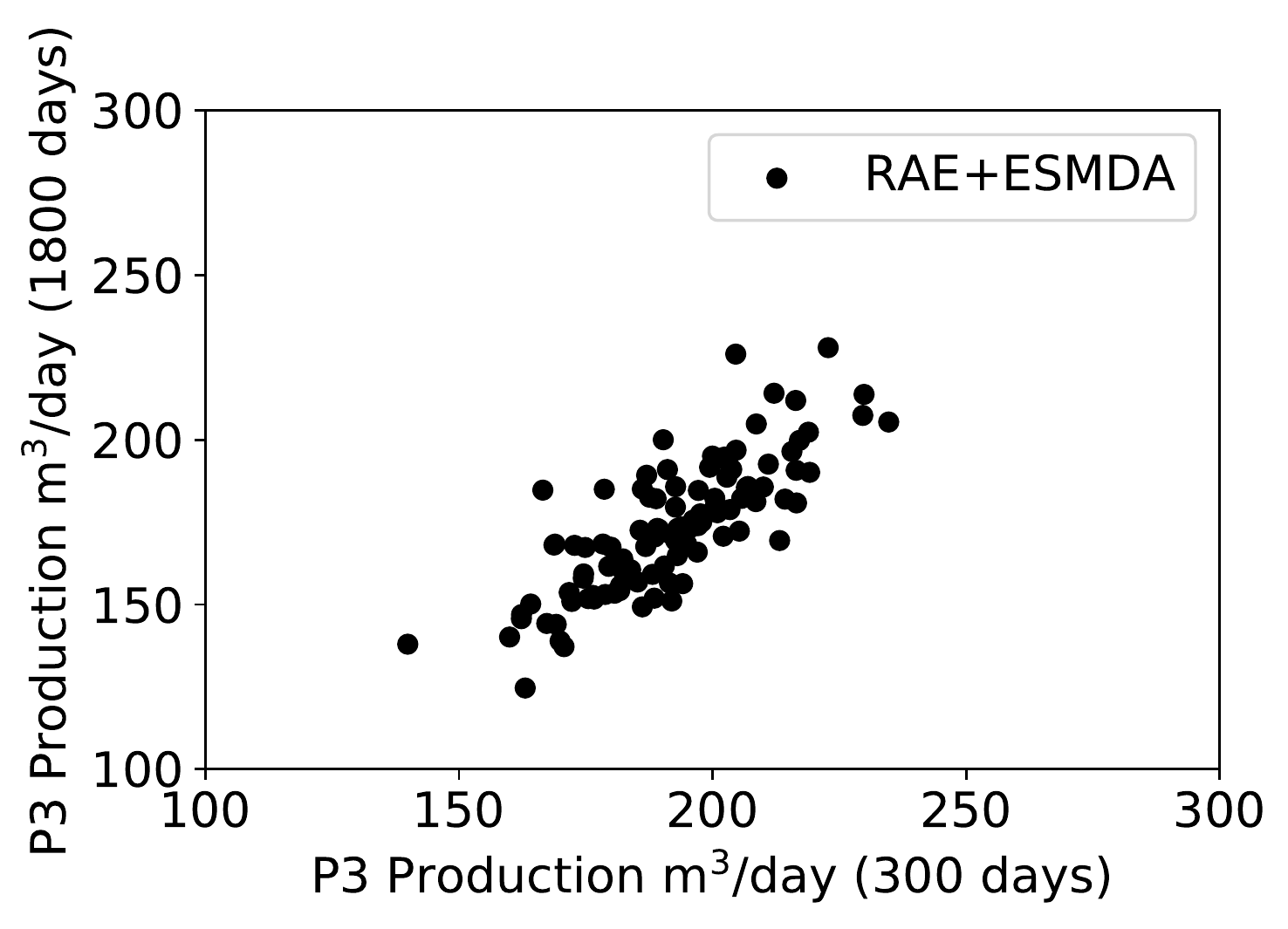}
\subcaption{RAE+ESMDA}
\end{minipage}
\caption{Cross-plots for P3 total liquid (water plus oil) production rate at 300~days and 1800~days}\label{fig:corr_2d_p3_production}
\end{figure*}

We first assess the ability of the various procedures to represent correlations in the posterior predictions. Figure~\ref{fig:corr_2d_p3_production} displays  cross-plots of total liquid production rate in well P3 at 300~days and 1800~days. The red points in Fig.~\ref{fig:corr_2d_p3_production}(a) represent the rejection sampling results (recall that 117 data vectors were accepted). Posterior results for the other three methods are shown in Fig.~\ref{fig:corr_2d_p3_production}(b)-(d). We randomly selected 117 points from the 800 DSI posterior results for these plots to enable clearer comparisons with RS. The positive correlation in the RS results is not reproduced by PCA+HT, and is only partially captured by ESMDA with truncation. RAE, by contrast, provides results that closely resemble the RS results. This suggests that RS does indeed capture correlations between these variables. 

In order to more quantitatively assess relationships in DSI posterior time series, we compute covariance and correlation. The covariance between two variables $x$ and $y$ is given by
\begin{equation}
    \text{Cov}(x, y) = E[(x - \mu_x)(y - \mu_y)],
\end{equation}
where $\mu_x$and $\mu_y$ are the mean values of $x$ and $y$. The correlation coefficient is given by
\begin{equation}
    \text{Corr}(x, y) = \frac{\text{Cov}(x, y)}{\sigma_x \sigma_y},
\end{equation}
where $\sigma_x$ and $\sigma_y$ denote the standard deviations of variables $x$ and $y$. Note that $\text{Corr}(x, y)$ is in the range of $[-1, 1]$.

\begin{figure*}[!ht]
\centering
\begin{minipage}{.4\linewidth}\centering
\includegraphics[width=\linewidth]{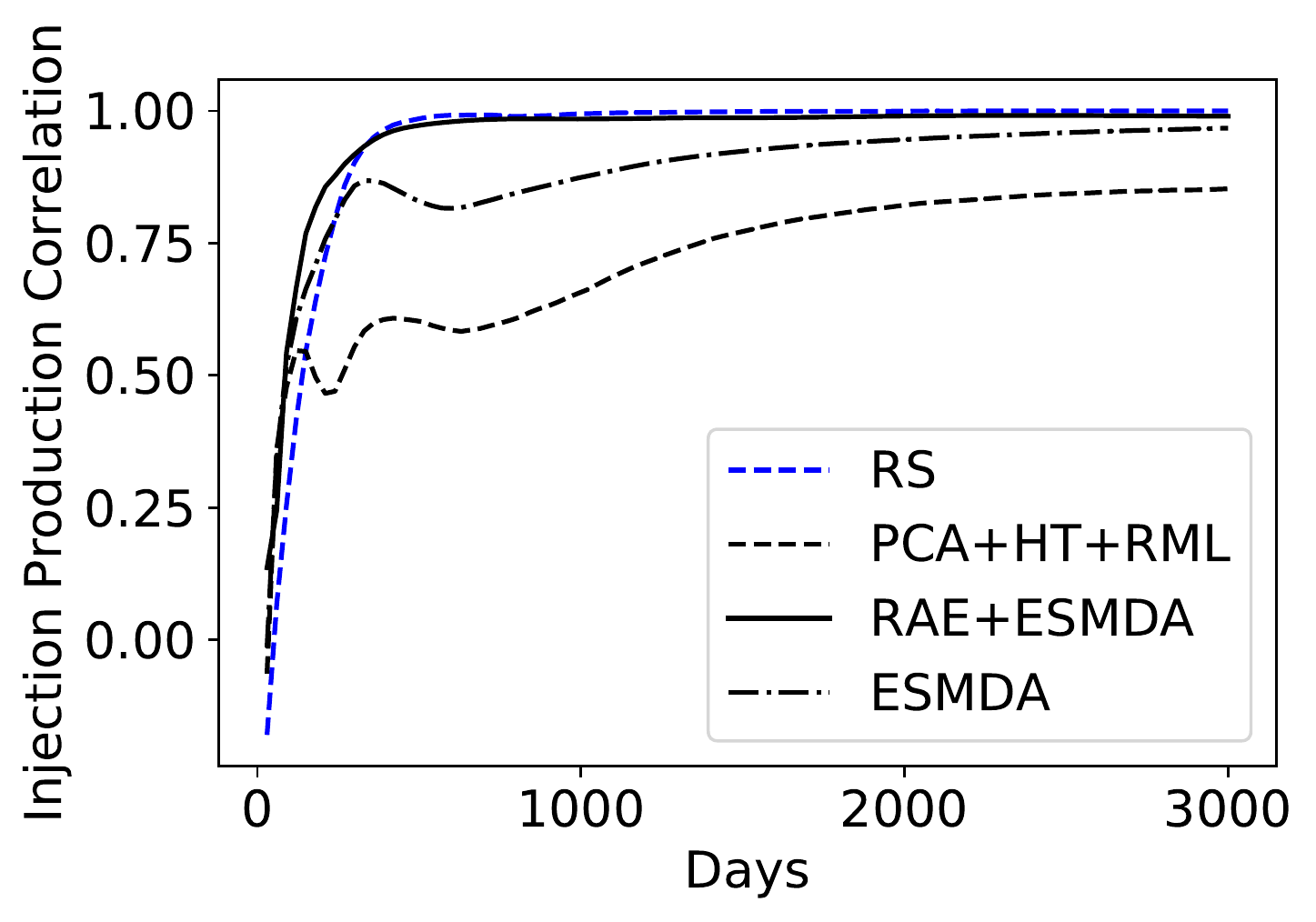}
\subcaption{Correlation between field-wide injection and production rates}
\end{minipage}
\hspace{.05\linewidth}
\begin{minipage}{.4\linewidth}\centering
\includegraphics[width=\linewidth]{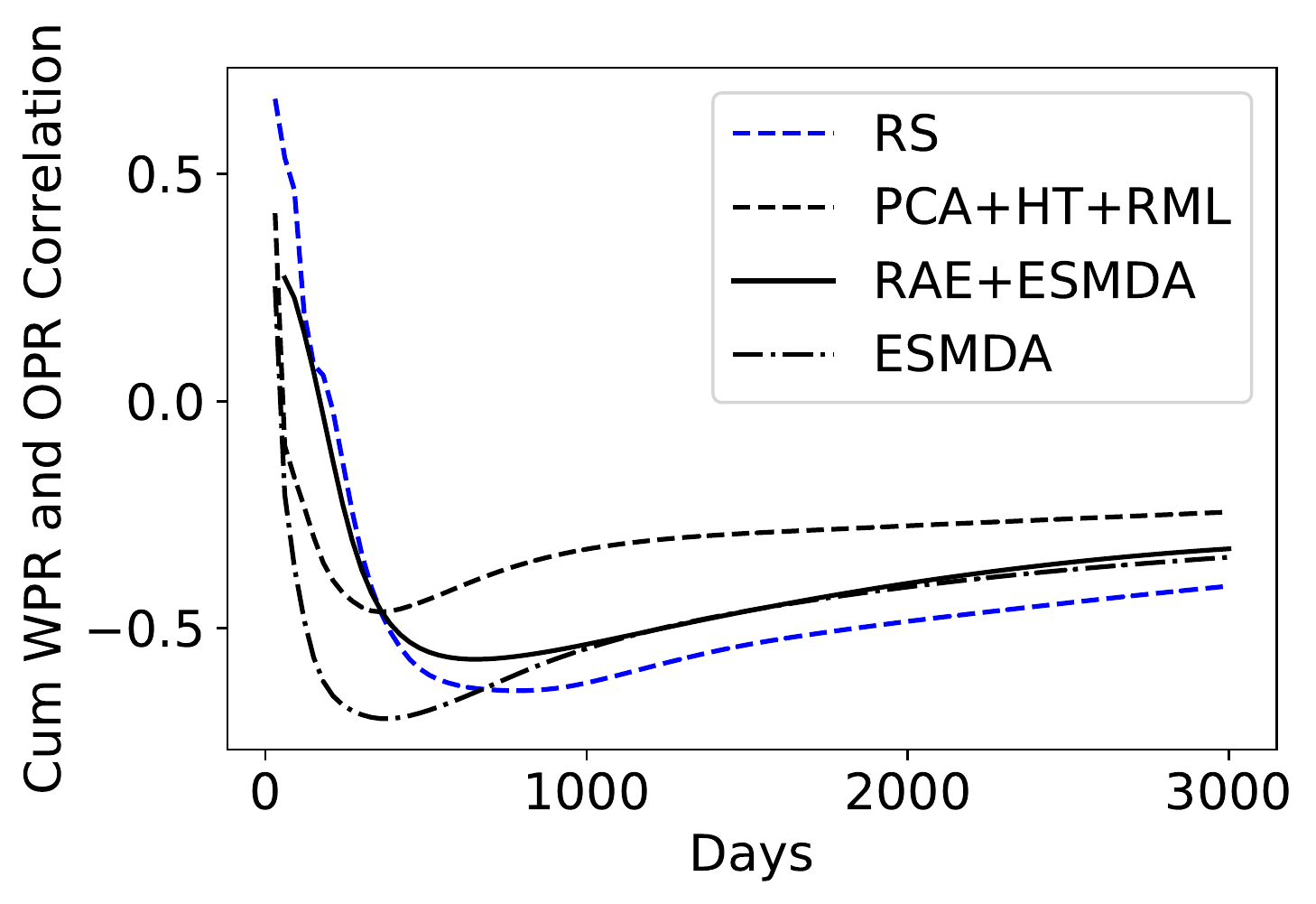}
\subcaption{Correlation between field-wide WPR and OPR}
\end{minipage}
\begin{minipage}{.4\linewidth}\centering
\includegraphics[width=\linewidth]{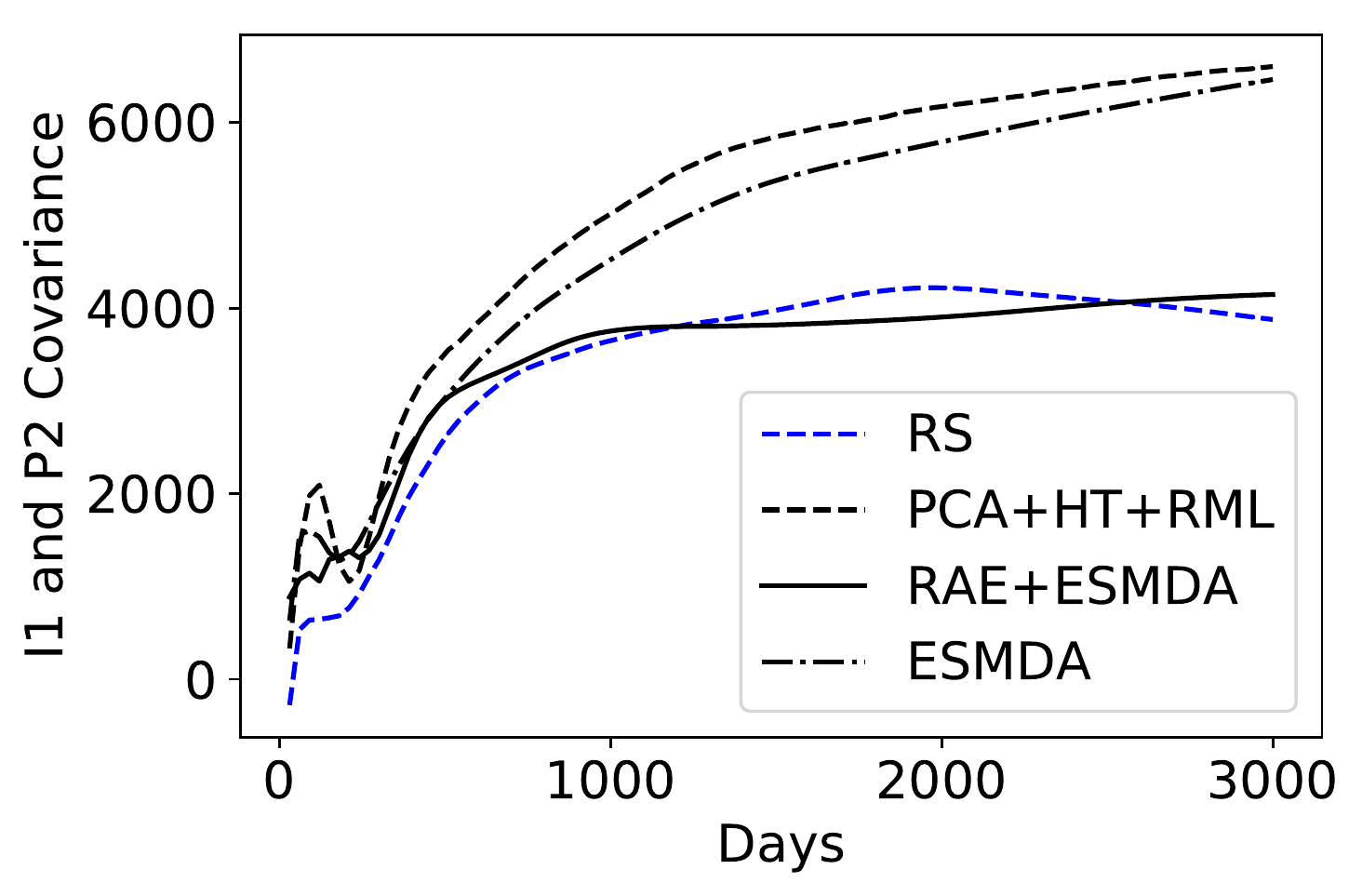}
\subcaption{Covariance between I1 WIR and P2 total liquid production}
\end{minipage}
\hspace{.05\linewidth}
\begin{minipage}{.4\linewidth}\centering
\includegraphics[width=\linewidth]{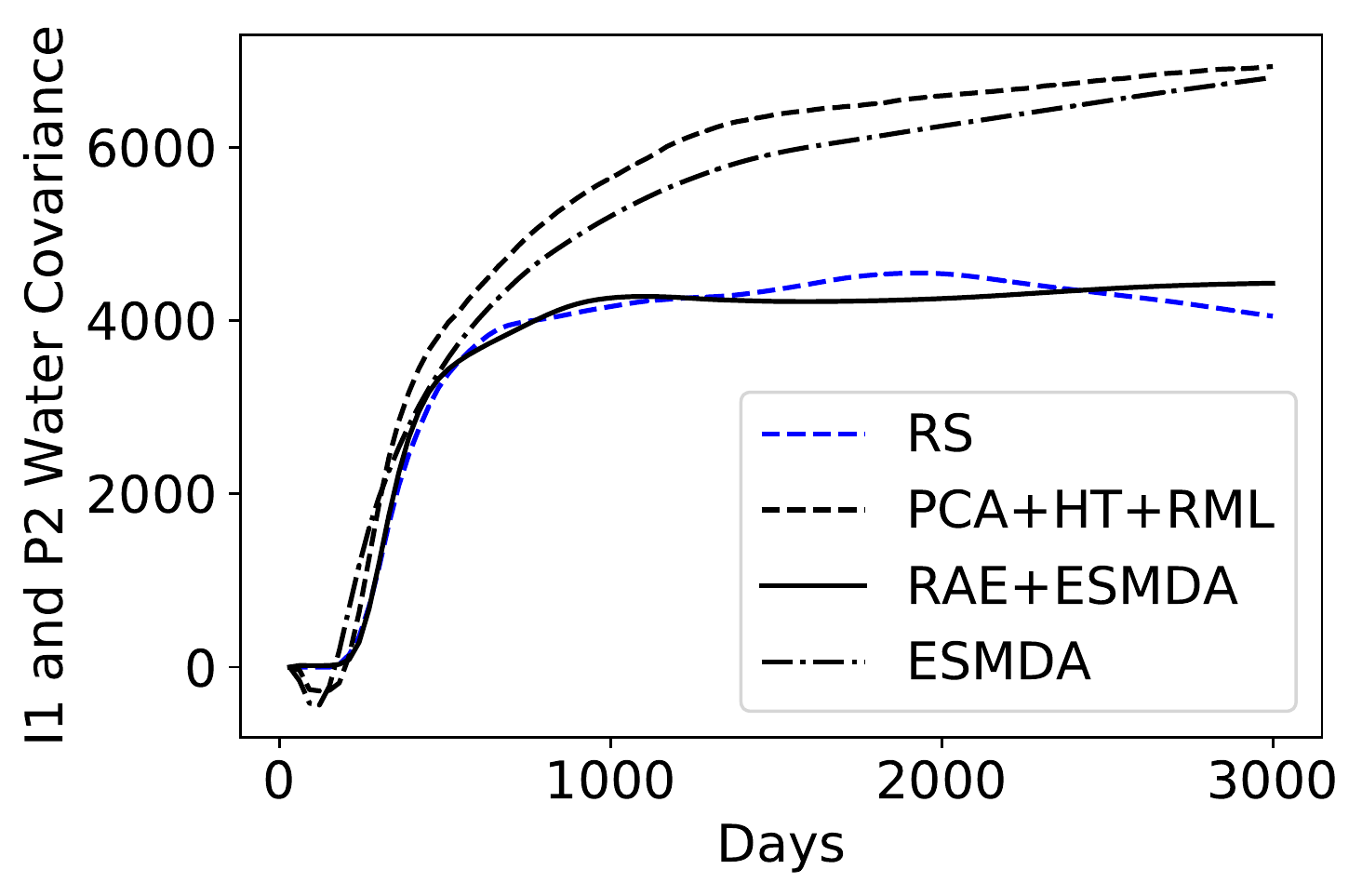}
\subcaption{Covariance between I1 WIR and P2 WPR}
\end{minipage}
\caption{Correlation and covariance for various quantities using different DSI treatments} \label{fig:corr_cha_time}
\end{figure*}

We can calculate the covariance and/or correlation between different quantities at each time step over the full simulation period. Figure~\ref{fig:corr_cha_time} displays results for the various DSI procedures. In Fig.~\ref{fig:corr_cha_time}(a) we show the correlation between field-wide injection and production rates. Because the system is nearly incompressible, after the initial transient period, the injection and production rates should balance. This means the correlation coefficient should be close to 1. DSI with the RAE parameterization (solid black curve) clearly captures the RS result (blue dashed curve), while the other two treatments show differences. Figure~\ref{fig:corr_cha_time}(b) presents the correlation between field-wide water production rate and oil production rate, which is seen to be negative over most of the simulation time frame. The RAE+ESMDA treatment captures the general trend in these results. ESMDA with truncation also tracks the RS result over most of the simulation time frame.

Figure~\ref{fig:corr_cha_time}(c) displays the covariance between I1 injection rate and P2 total liquid production rate (note that the latter is itself a derived quantity), and Fig.~\ref{fig:corr_cha_time}(d) shows the covariance between I1 injection rate and P2 water production rate. Here we see that the RAE+ESMDA procedure clearly provides results in close agreement with RS, while the other two treatments significantly over-predict covariance in both cases. We note finally that covariance between primary quantities appears directly in the data assimilation process (equations~\eqref{eq:esmda} and~\eqref{eq:esmda_ksi}), so it is important that this be represented accurately in the parameterization.

\begin{figure*}[!ht]
\centering
\begin{minipage}{.4\linewidth}\centering
\includegraphics[width=\linewidth]{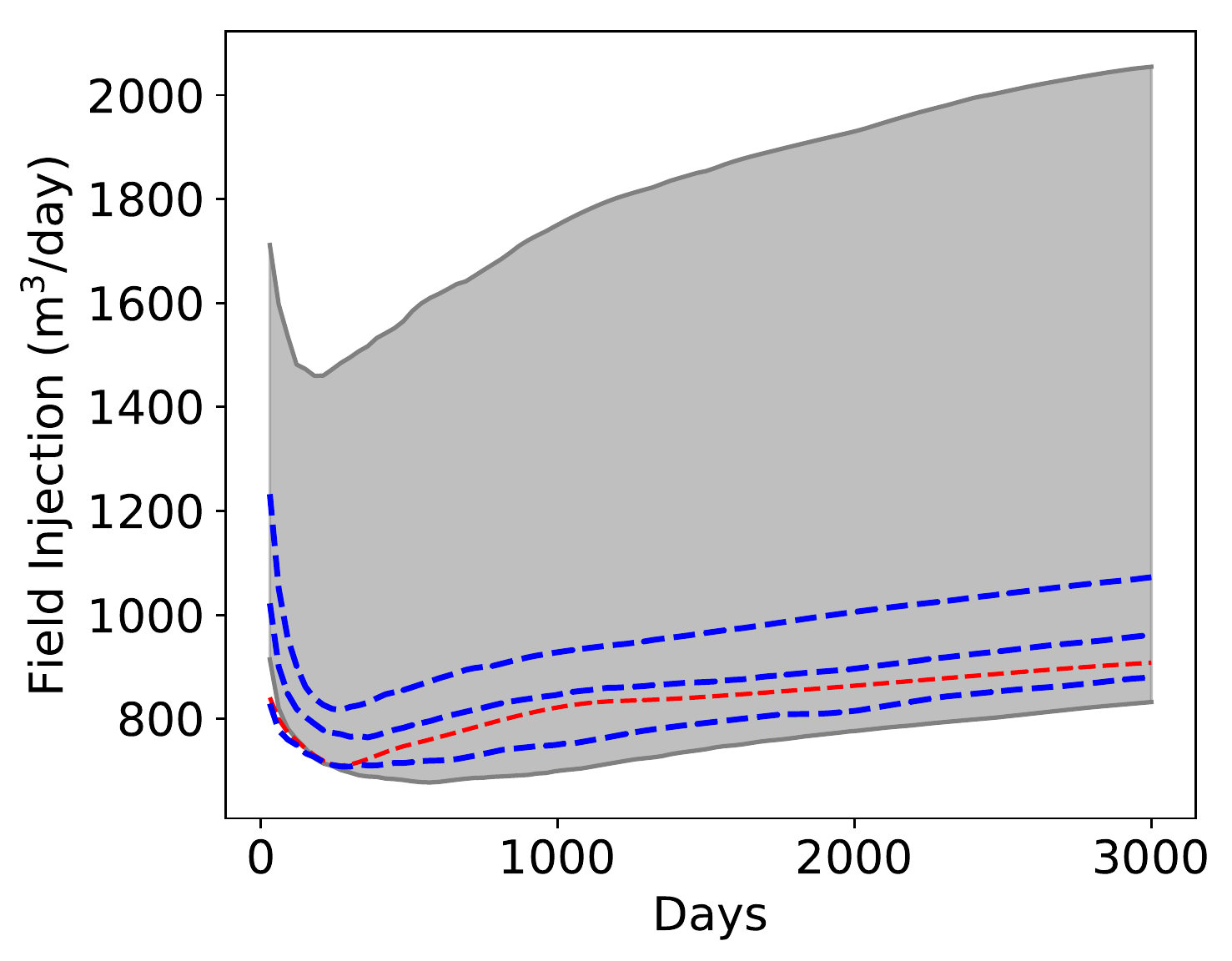}
\subcaption{Field-wide injection rate}
\end{minipage}
\hspace{.05\linewidth}
\begin{minipage}{.4\linewidth}\centering
\includegraphics[width=\linewidth]{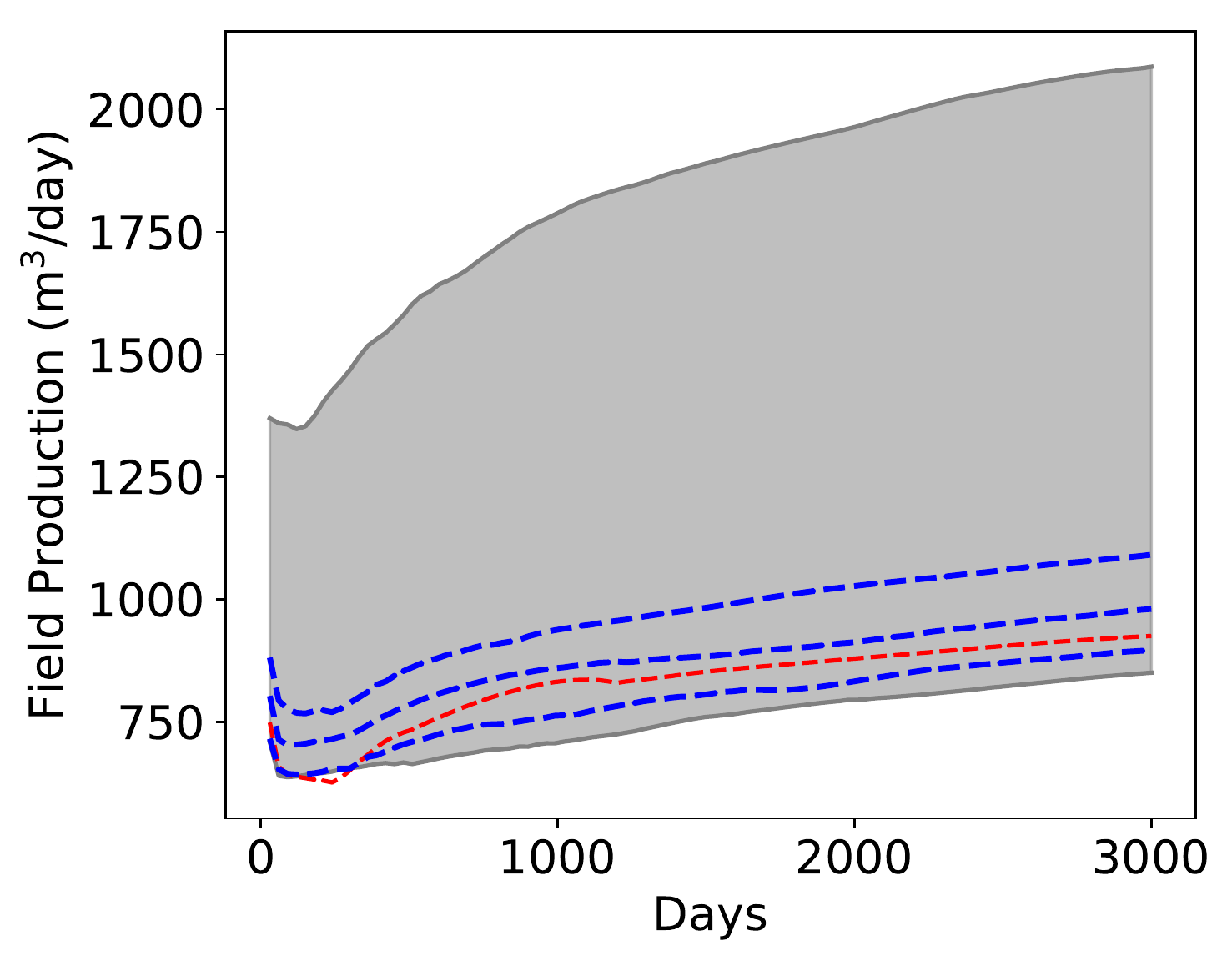}
\subcaption{Field-wide total liquid production rate}
\end{minipage}
\begin{minipage}{.4\linewidth}\centering
\includegraphics[width=\linewidth]{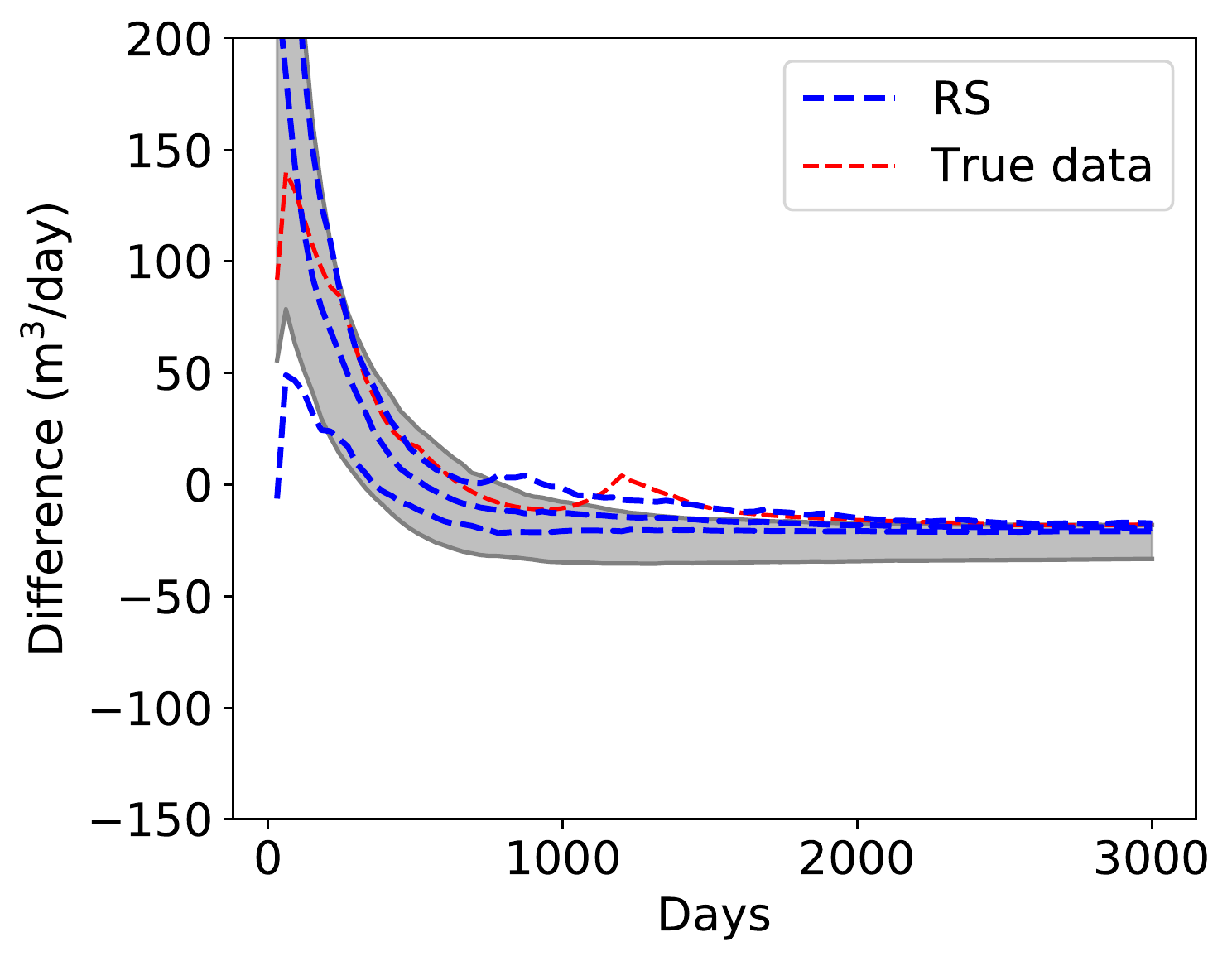}
\subcaption{Difference in field-wide injection and total liquid production rates}
\end{minipage}
\hspace{.05\linewidth}
\begin{minipage}{.4\linewidth}\centering
\includegraphics[width=\linewidth]{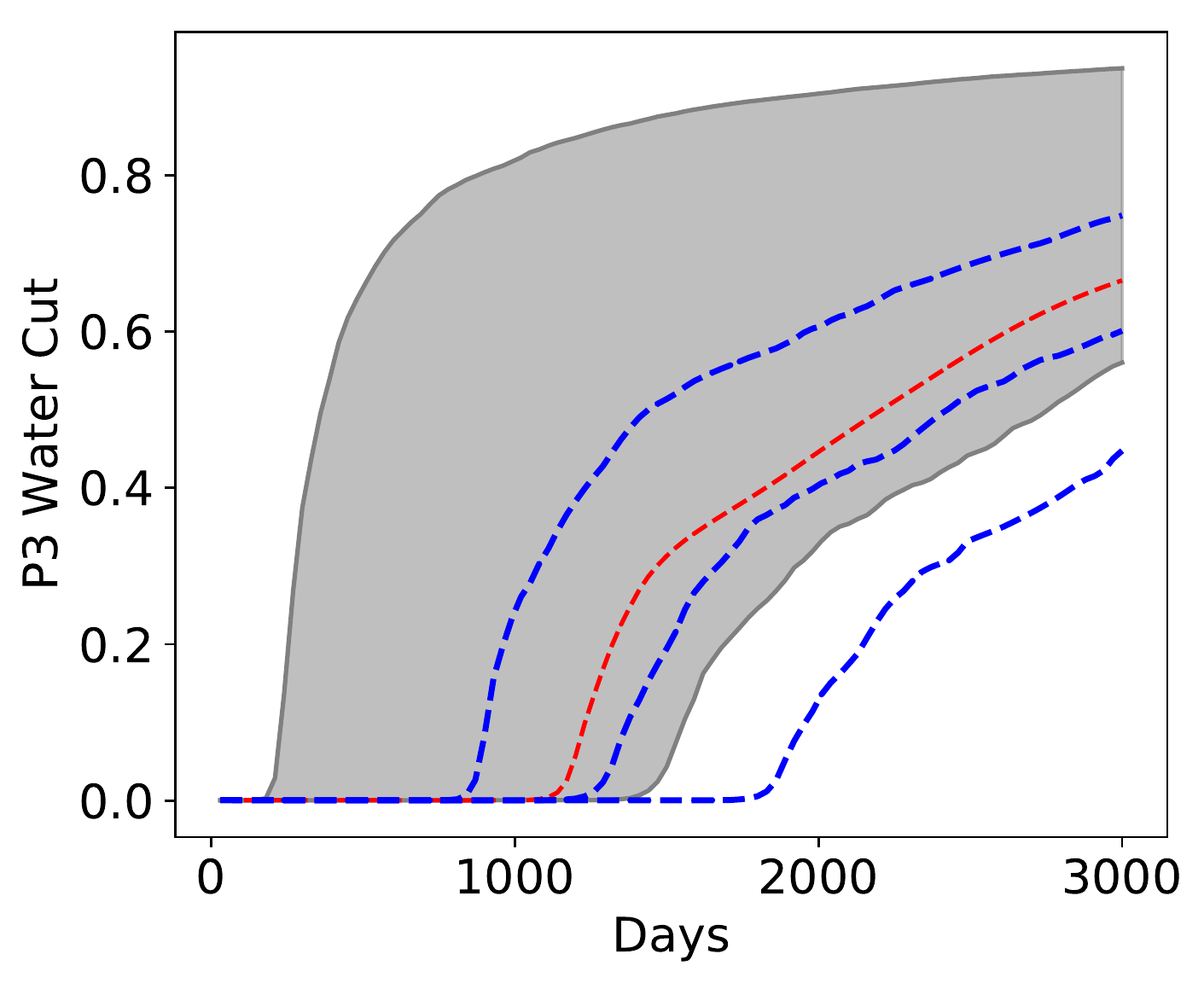}
\subcaption{P3 water cut}
\end{minipage}
\caption{Prior simulation results and RS results for derived quantities. Legend in (c) applies to all subplots. The gray shaded regions show the P$_{10}$-P$_{90}$ range of prior simulation results}\label{fig:der_2d_prior}
\end{figure*}

We next consider DSI predictions for time series of derived quantities. Figure~\ref{fig:der_2d_prior} presents prior simulation results and RS posterior results for field-wide injection rate, field-wide total liquid production rate, P3 water cut, and the difference between field-wide injection and total liquid production rates. The P$_{10}$-P$_{90}$ ranges for the prior simulation results are indicated by the gray regions. The true data are denoted by the red curves and the P$_{10}$, P$_{50}$ and P$_{90}$ RS results by the blue dashed curves. The uncertainty reduction in the posterior results relative to the prior results is quite substantial in Fig.~\ref{fig:der_2d_prior}(a), (b) and (d).

\begin{figure*}[!ht]
\centering
\begin{minipage}{.4\linewidth}\centering
\includegraphics[width=\linewidth]{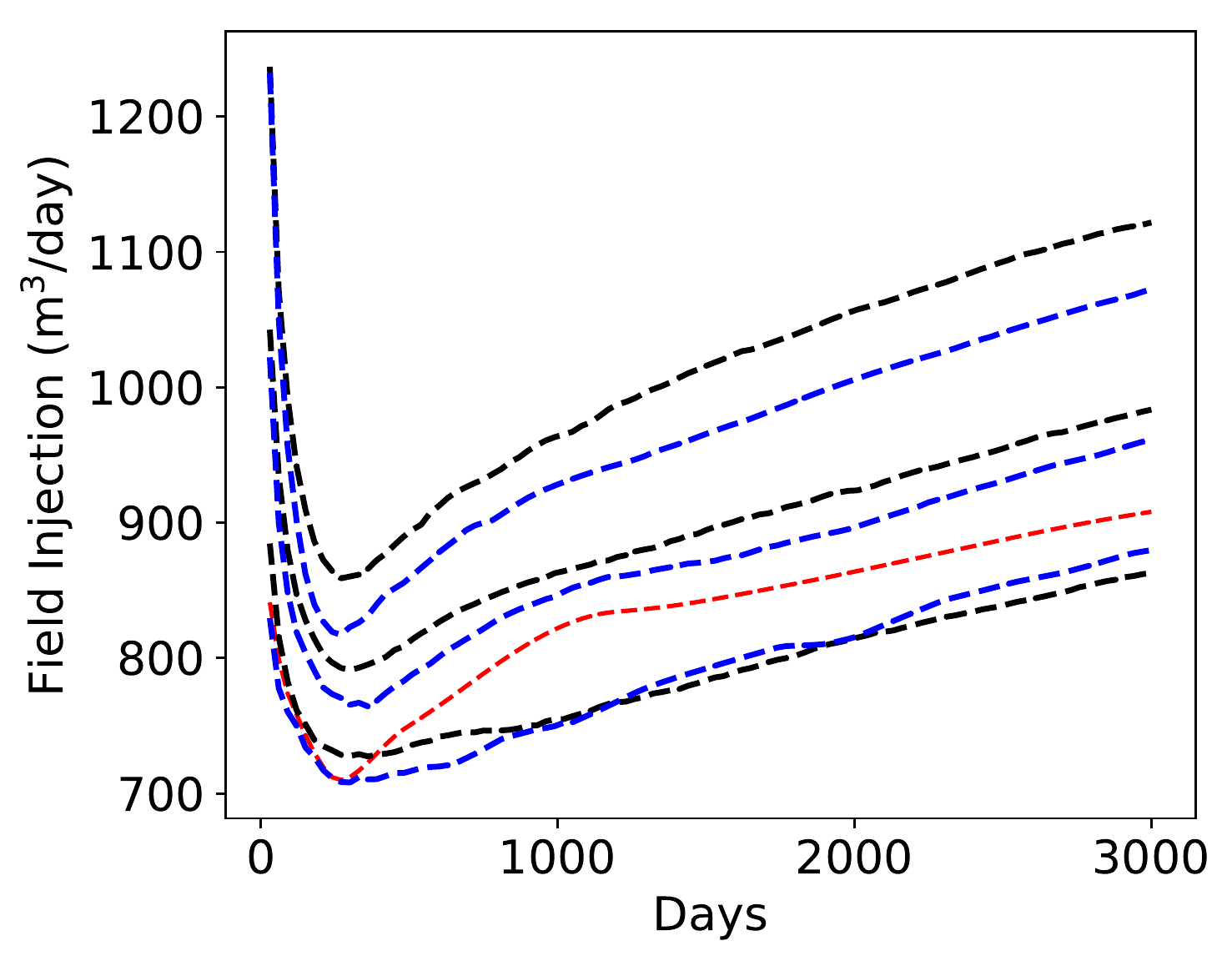}
\subcaption{Field-wide injection rate}
\end{minipage}
\hspace{.05\linewidth}
\begin{minipage}{.4\linewidth}\centering
\includegraphics[width=\linewidth]{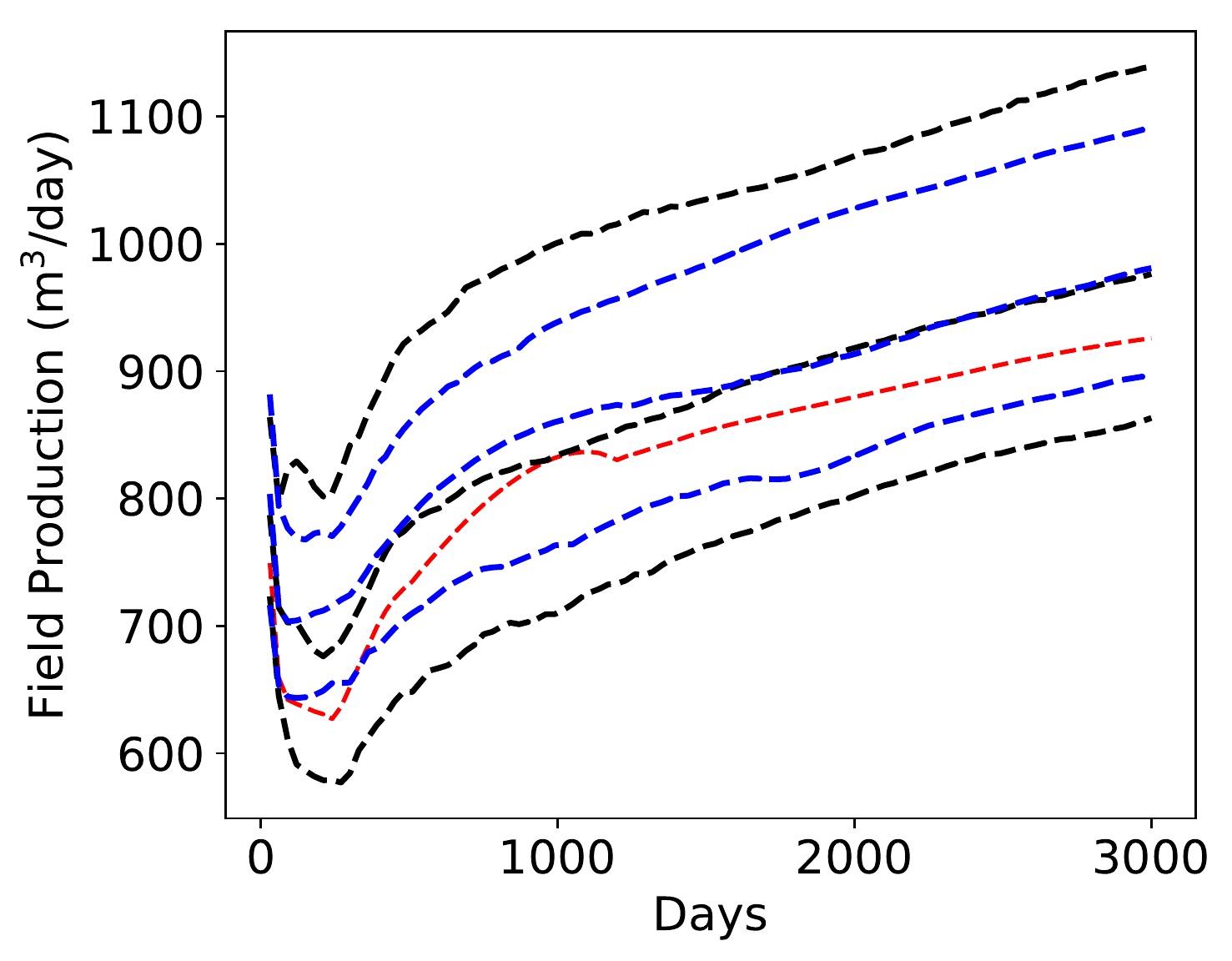}
\subcaption{Field-wide total liquid production rate}
\end{minipage}
\begin{minipage}{.4\linewidth}\centering
\includegraphics[width=\linewidth]{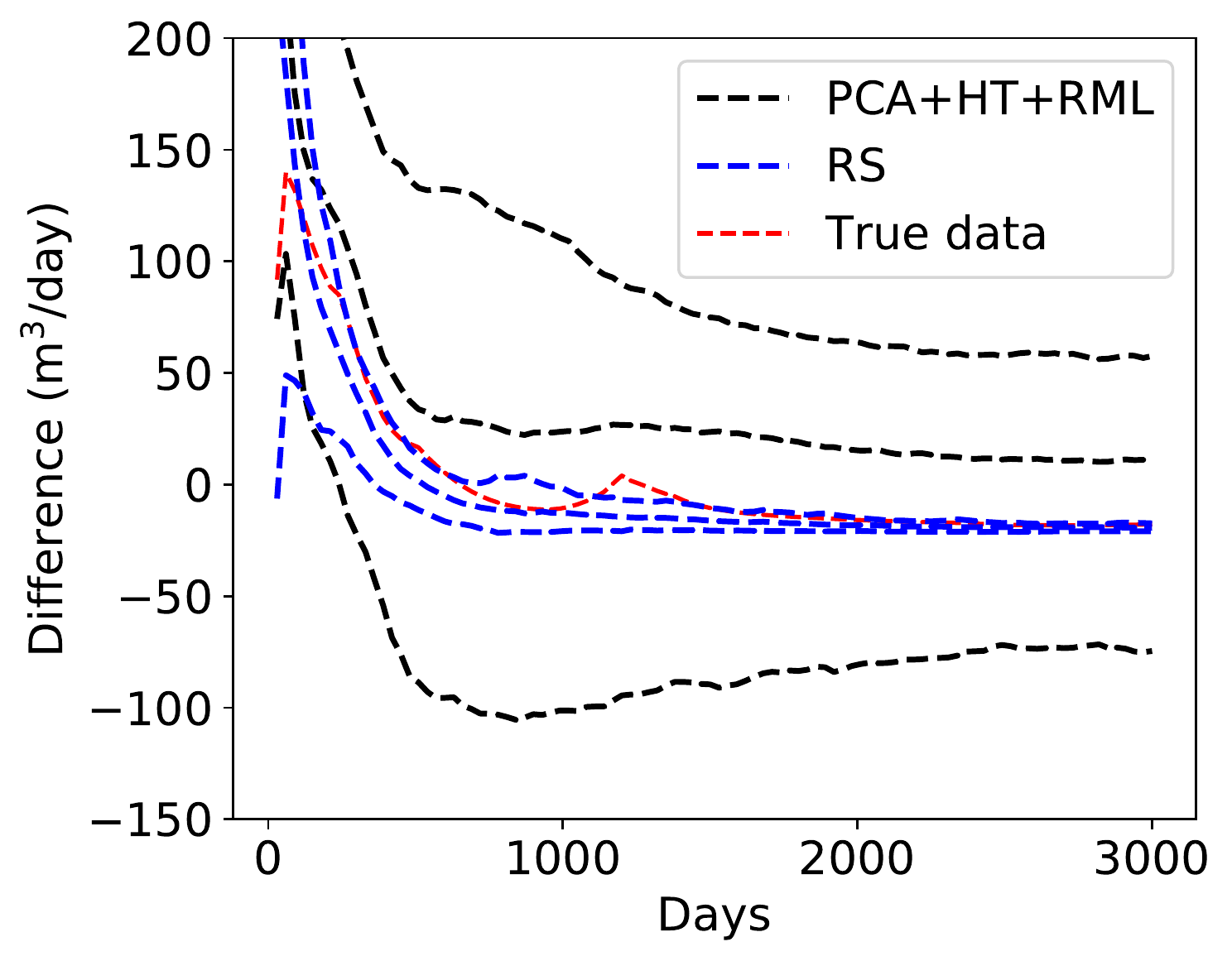}
\subcaption{Difference in field-wide injection and total liquid production rates}
\end{minipage}
\hspace{.05\linewidth}
\begin{minipage}{.4\linewidth}\centering
\includegraphics[width=\linewidth]{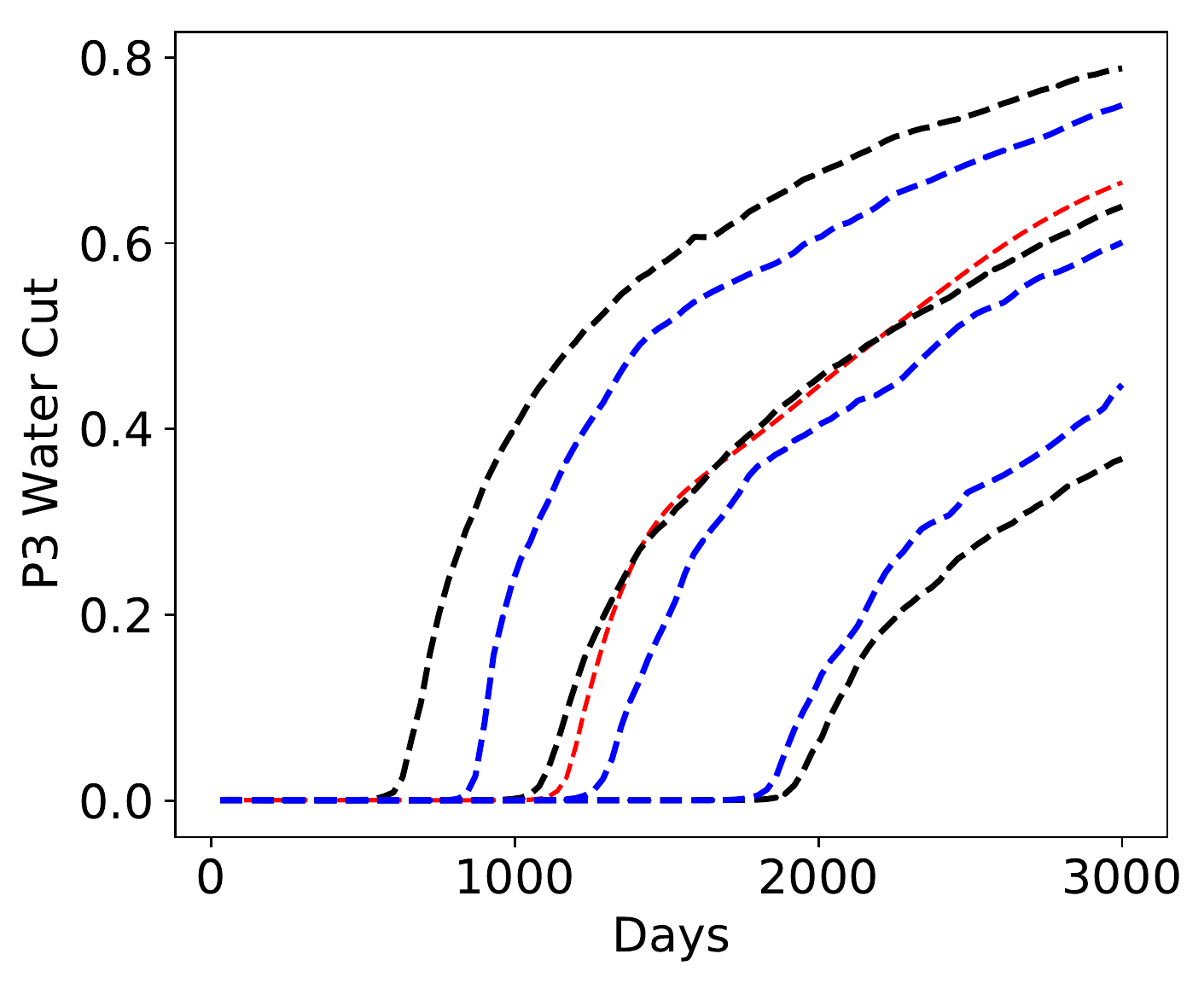}
\subcaption{P3 water cut}
\end{minipage}
\caption{PCA+HT+RML DSI results for derived quantities. Legend in (c) applies to all subplots}\label{fig:der_2d_pca}
\end{figure*}

Figure~\ref{fig:der_2d_pca} compares the P$_{10}$, P$_{50}$ and P$_{90}$ posterior results from PCA+HT+RML to those from RS for these derived quantities. The PCA+HT+RML results are reasonably close to those from RS for three of the four derived quantities considered (Fig.~\ref{fig:der_2d_pca}(a), (b) and (d)). They do, however, display a larger uncertainty range relative to the RS results. The PCA+HT+RML results are quite inaccurate in their predictions for the difference between injection and total liquid production rates, shown in Fig.~\ref{fig:der_2d_pca}(c). This quantity should be near zero since total injection and production essentially balance after the initial transient. This difference is a particularly challenging quantity for PCA+HT to resolve, as it involves a combination of eight primary quantities at each time step. The PCA+HT treatment, which strictly captures only marginal distributions, does not maintain 
correlations accurately enough to represent this quantity.

\begin{figure*}[!ht]
\centering
\begin{minipage}{.4\linewidth}\centering
\includegraphics[width=\linewidth]{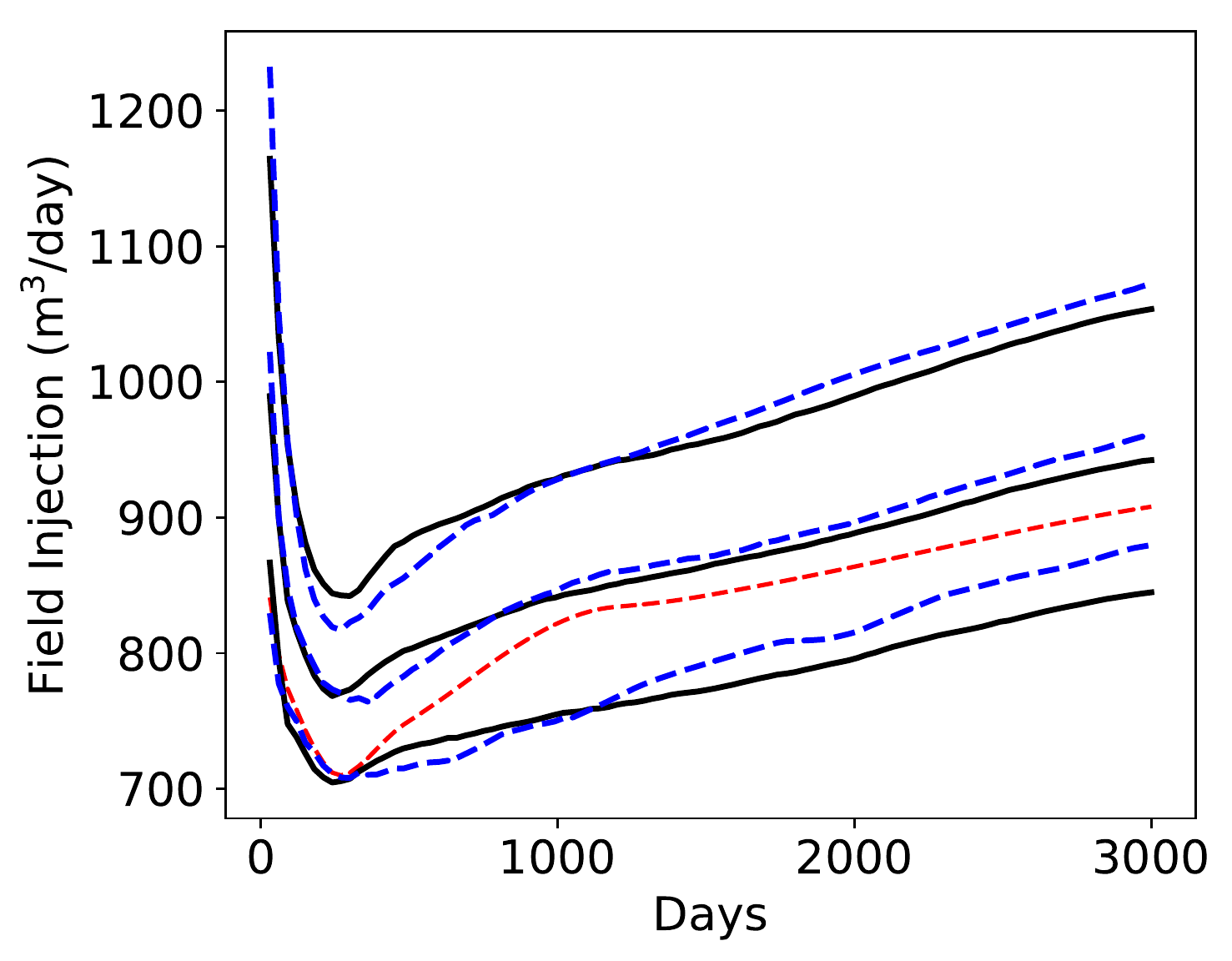}
\subcaption{Field-wide injection rate}
\end{minipage}
\hspace{.05\linewidth}
\begin{minipage}{.4\linewidth}\centering
\includegraphics[width=\linewidth]{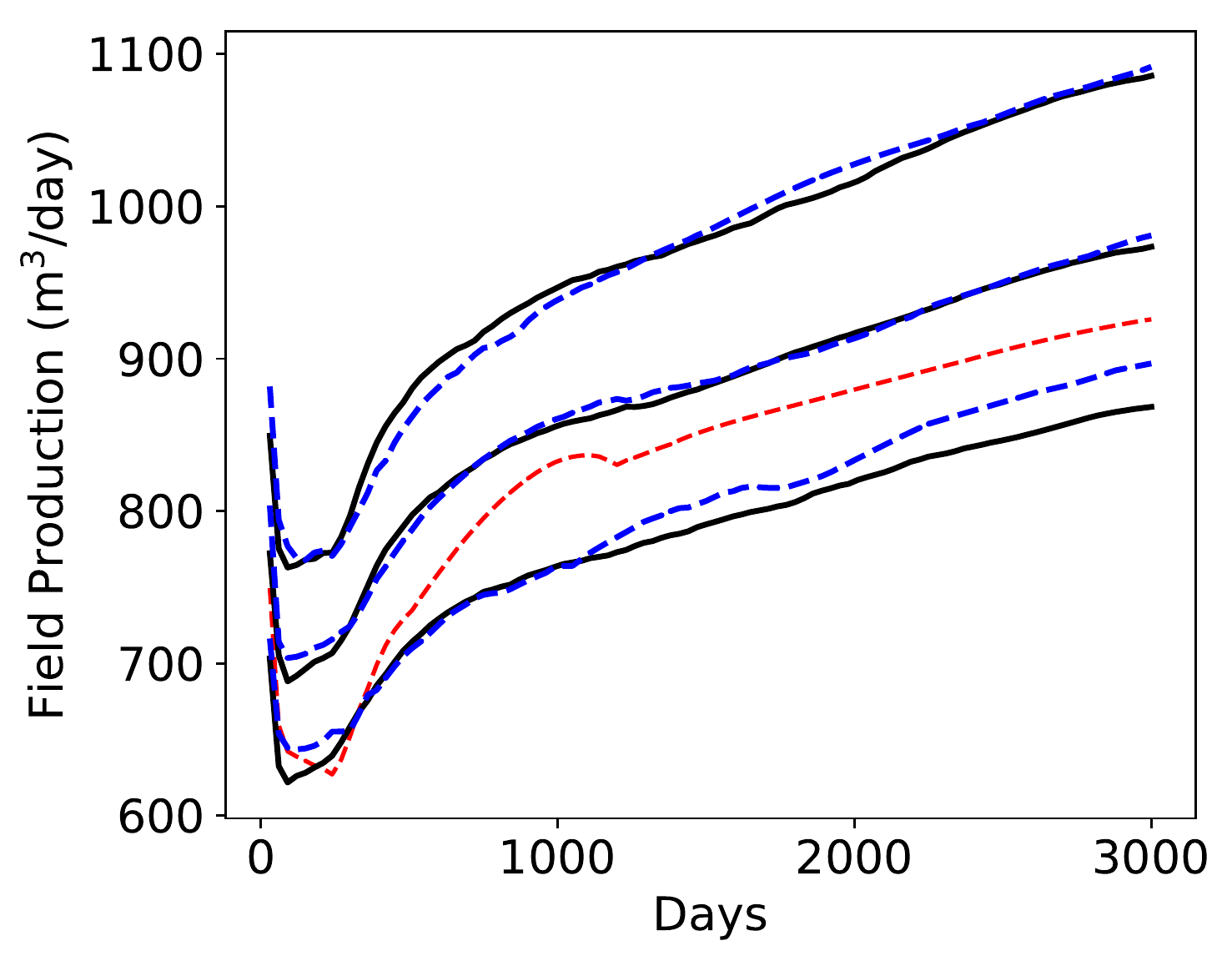}
\subcaption{Field-wide total liquid production rate}
\end{minipage}
\begin{minipage}{.4\linewidth}\centering
\includegraphics[width=\linewidth]{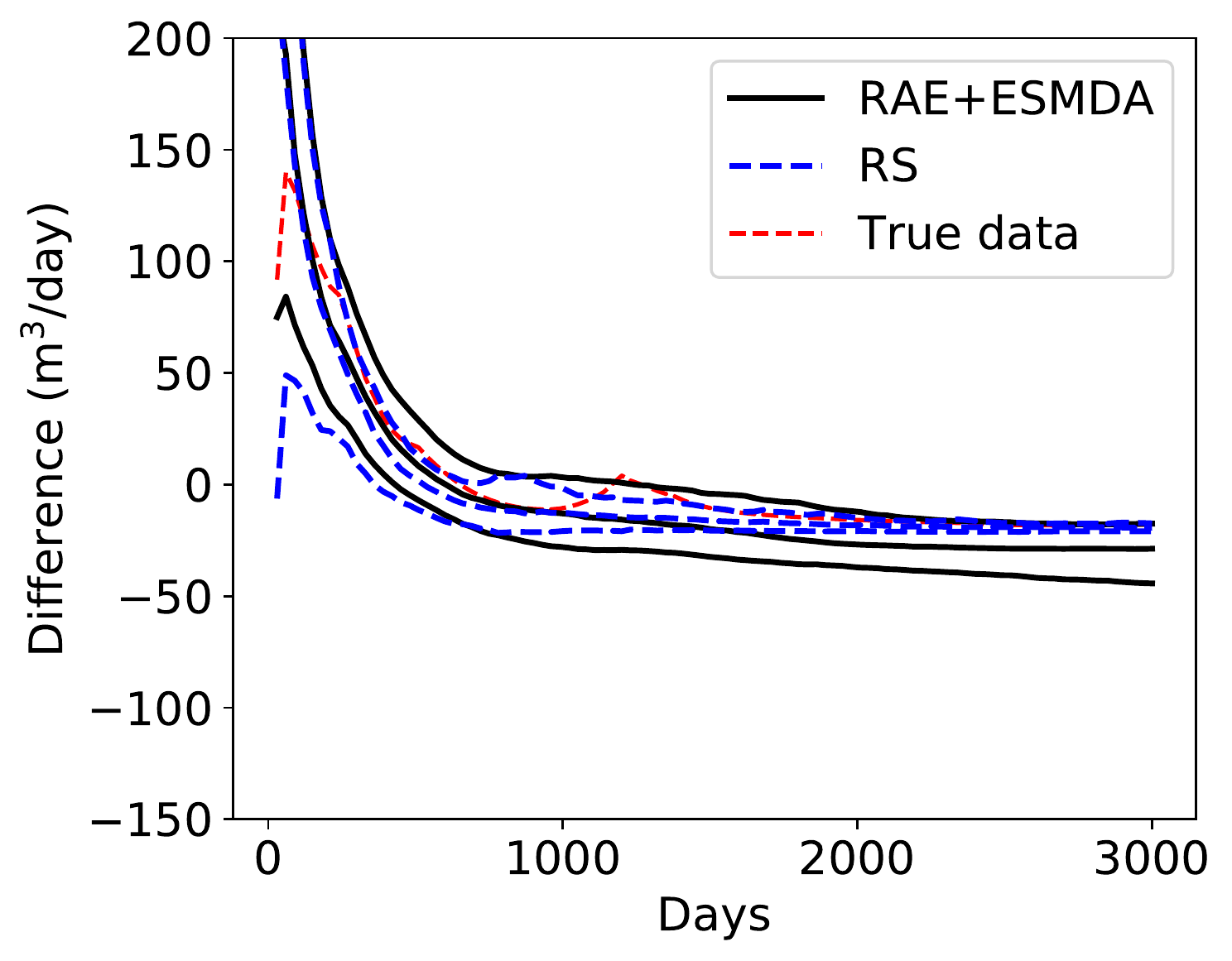}
\subcaption{Difference in field-wide injection and total liquid production rates}
\end{minipage}
\hspace{.05\linewidth}
\begin{minipage}{.4\linewidth}\centering
\includegraphics[width=\linewidth]{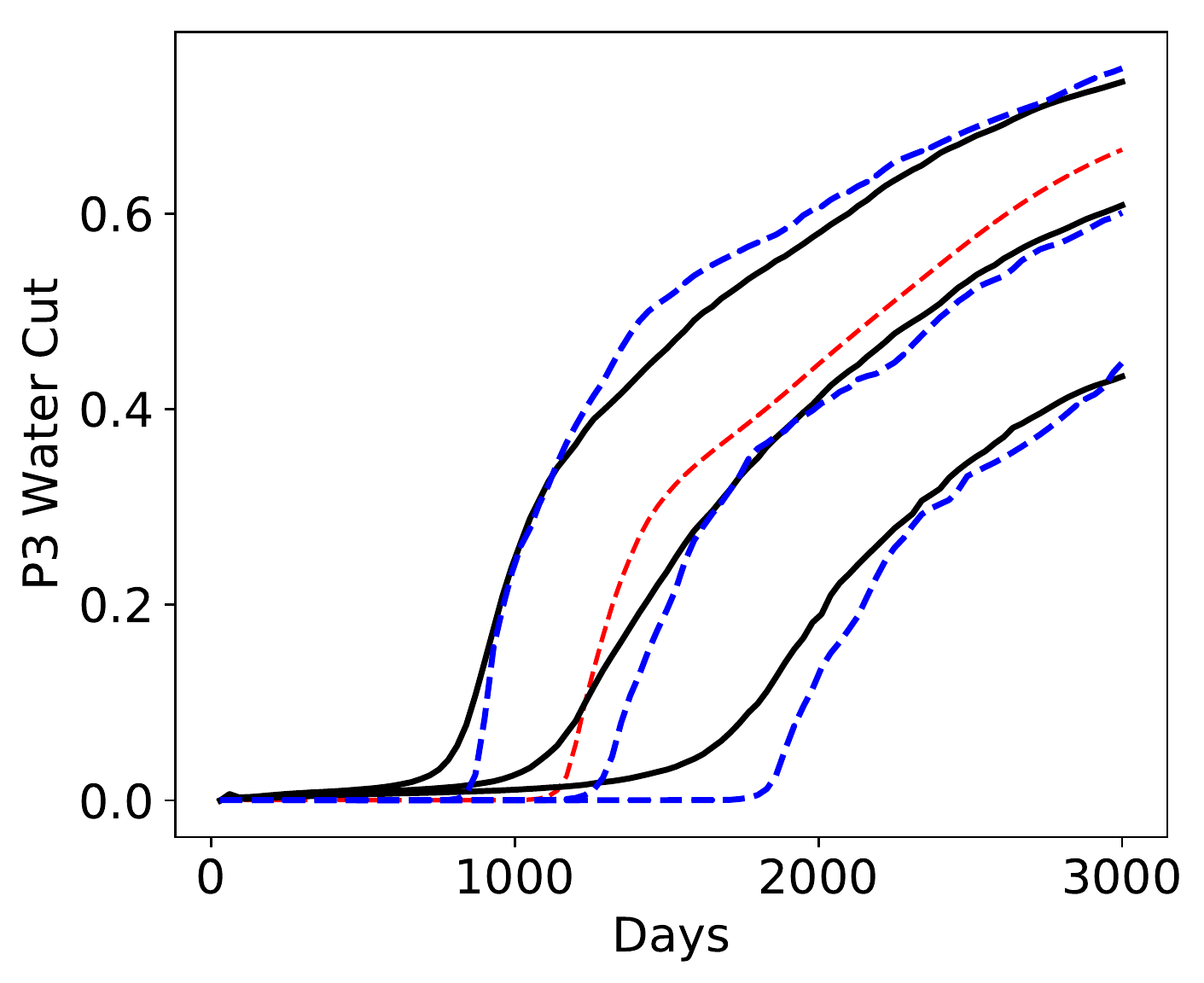}
\subcaption{P3 water cut}
\end{minipage}
\caption{RAE+ESMDA DSI results for derived quantities. Legend in (c) applies to all subplots}\label{fig:der_2d_rae}
\end{figure*}

Posterior results for these derived quantities using the RAE+ESMDA DSI procedure are shown in Fig.~\ref{fig:der_2d_rae}. Overall, the P$_{10}$, P$_{50}$ and P$_{90}$ RAE+ESMDA posterior results match RS reasonably well. The improvement in the difference between field-wide injection and total liquid production rates, relative to the PCA+HT results, is immediately apparent (compare Figs.~\ref{fig:der_2d_rae}(c) and~\ref{fig:der_2d_pca}(c)). The RAE+ESMDA results in Fig.~\ref{fig:der_2d_rae}(c) do slightly overestimate posterior uncertainty at late time, but the general trend is clearly captured. The P3 water cut results in Fig.~\ref{fig:der_2d_rae}(d) also display some discrepancy relative to the RS results, particularly in breakthrough time in the P$_{10}$ and P$_{50}$ curves. The general level of agreement is, however, clearly better than that observed with the PCA+HT procedure.

\begin{figure}[!htpb]
\centering
\includegraphics[width = 0.47\textwidth]{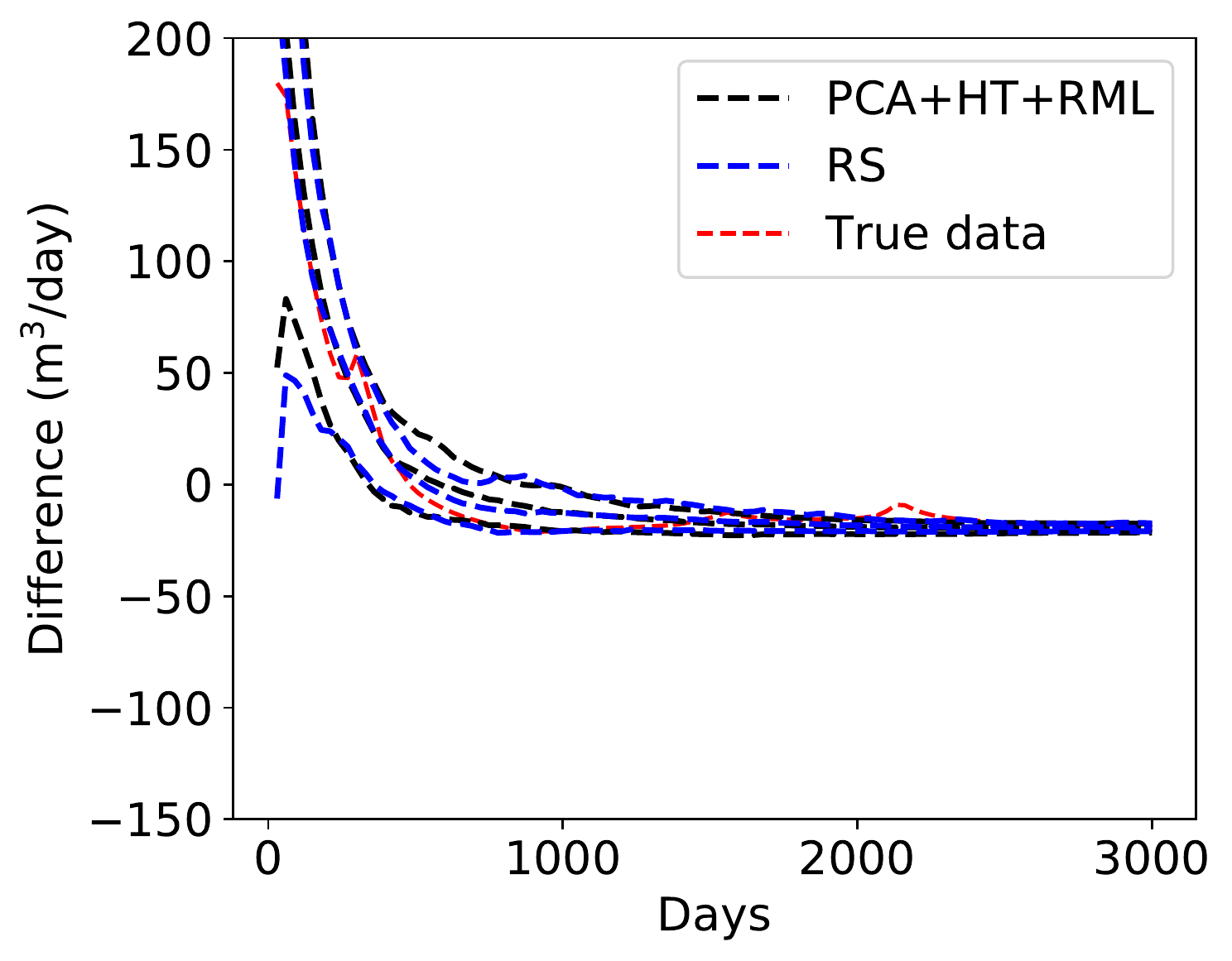}
\caption{PCA+HT+RML DSI results for difference between field-wide injection and total liquid production rates, with this difference included as a primary quantity in $\Bd$}\label{fig:2d_diff_pca}
\end{figure}

We now demonstrate that we are able to achieve improved results for particular quantities of interest, using the PCA+HT parameterization, if we include these variables directly in the data vector $\Bd$ (i.e., treat them as primary rather than derived quantities). To accomplish this we run the PCA+ HT+RML DSI method with the difference between field-wide injection and total liquid production rates, at each time step, included in $\Bd$. Posterior results using this treatment are shown in Fig.~\ref{fig:2d_diff_pca}. Agreement with the RS results is clearly much better here than when this difference is treated as a derived quantity (see Fig.~\ref{fig:der_2d_pca}(c)). Although it would not be practical to include all possible quantities of interest directly in the data vector, this indicates that posterior predictions for particular variables may be captured with acceptable accuracy using the PCA+HT parameterization.

\subsection{DSI Results for Additional `True' Models}
\label{sec:multi_true}

In this section we present results, in a very compact form, for two additional `true' models. These will be referred to as True Models~B and C (the model considered thus far will be called True Model~A). We will evaluate the posterior results for different DSI treatments in terms of a measure of consistency -- specifically the Mahalanobis distance -- relative to the reference RS results.

The Mahalanobis distance considered here quantifies the difference between each DSI posterior data realization $(\Bd_{\text{post}})_i$, $i = 1, \ldots, N_{\text{post}}$, where $N_{\text{post}}$ is the total number of posterior DSI data realizations, and the reference posterior distribution generated by RS. As in~\cite{sun2019data}, the Mahalanobis distance $D_M$ is defined as 
\begin{equation} \label{eq:DM}
\begin{split}
D_M((\Bd_{\text{post}})_i) = &\left( [(\Bd_{\text{post}})_i - \bar{
\Bd}_{\text{RS}}]^T C^{-1}_{\text{d}_\text{RS}} [(\Bd_{\text{post}})_i  - \bar{
\Bd}_{\text{RS}}] \right)^{\frac{1}{2}}, \\ 
&i = 1, \ldots, N_{\text{post}},
\end{split}
\end{equation}
where $\bar{\Bd}_{\text{RS}}$ and $C_{\text{d}_\text{RS}}$ denote the mean and covariance of the RS posterior results. Thus $D_M$ provides a measure of distance between each posterior data vector $(\Bd_{\text{post}})_i$ and the reference distribution $N(\bar{\Bd}_{\text{RS}}, C_{\text{d}_\text{RS}})$, where the mean and covariance are calculated from the $O(100)$ RS samples accepted. 

The covariance matrix in equation~\eqref{eq:DM} is, however, generally not invertible. We thus perform singular value decomposition on the data matrix containing, as its columns, the data vectors accepted by RS (the procedure is analogous to that described in Section~\ref{sec:PCA_HT}). We can now represent $\Bd_\text{RS}$ as $\Bd_\text{RS} = U \Sigma \boldsymbol{\omega} + \bar{\Bd}_{\text{RS}}$, where $U \in \R^{\Nf \times k}$ denotes the matrix containing the left singular vectors, $\Sigma \in \R^{k \times k}$ indicates a diagonal matrix of singular values, and $\boldsymbol{\omega} \in \R^{k \times 1}$ is the latent space variable, with $k$ the dimension of this latent space. With this representation, the covariance matrix is expressed as $C_{\text{d}_\text{RS}} = U \Sigma^2 U^T$. Thus, we now have
\begin{equation} \label{eq:DMw}
\begin{split}
&D_M((\Bd_{\text{post}})_i) = \left( \boldsymbol{\omega}_i^T \boldsymbol{\omega}_i \right)^{\frac{1}{2}}, \\
&\boldsymbol{\omega}_i =  \Sigma^{-1}U^T\left((\Bd_{\text{post}})_i- \bar{\Bd}_{\text{RS}} \right), \ \ i = 1, \ldots, N_{\text{post}}.  
\end{split}
\end{equation}
We apply an energy criterion to determine the value of $k$ for each true model. It is important to avoid over-fitting in this representation, as discussed by Sun and Durlofsky~\cite{sun2019data}. Here we preserve 99\% of the energy, which corresponds to $k=16$ or 17 for the three true models considered.

We compute a value of $D_M$ for each posterior data vector $(\Bd_{\text{post}})_i$, $i = 1, \ldots, N_{\text{post}}$. The $D_M$ values thus computed are displayed in terms of CDFs. To generate the CDF for the RS results, we take each accepted RS data vector in turn as $(\Bd_{\text{post}})_i$, and then apply equation~\eqref{eq:DMw} to compute $D_M((\Bd_{\text{post}})_i)$. The CDF is then constructed from the full set of RS posterior results. By comparing the CDFs for the various DSI methods with the RS CDF we can, very concisely, assess the degree of consistency in the posterior results.

\begin{figure*}[!ht]
\centering
\begin{minipage}{.4\linewidth}\centering
\includegraphics[width=\linewidth]{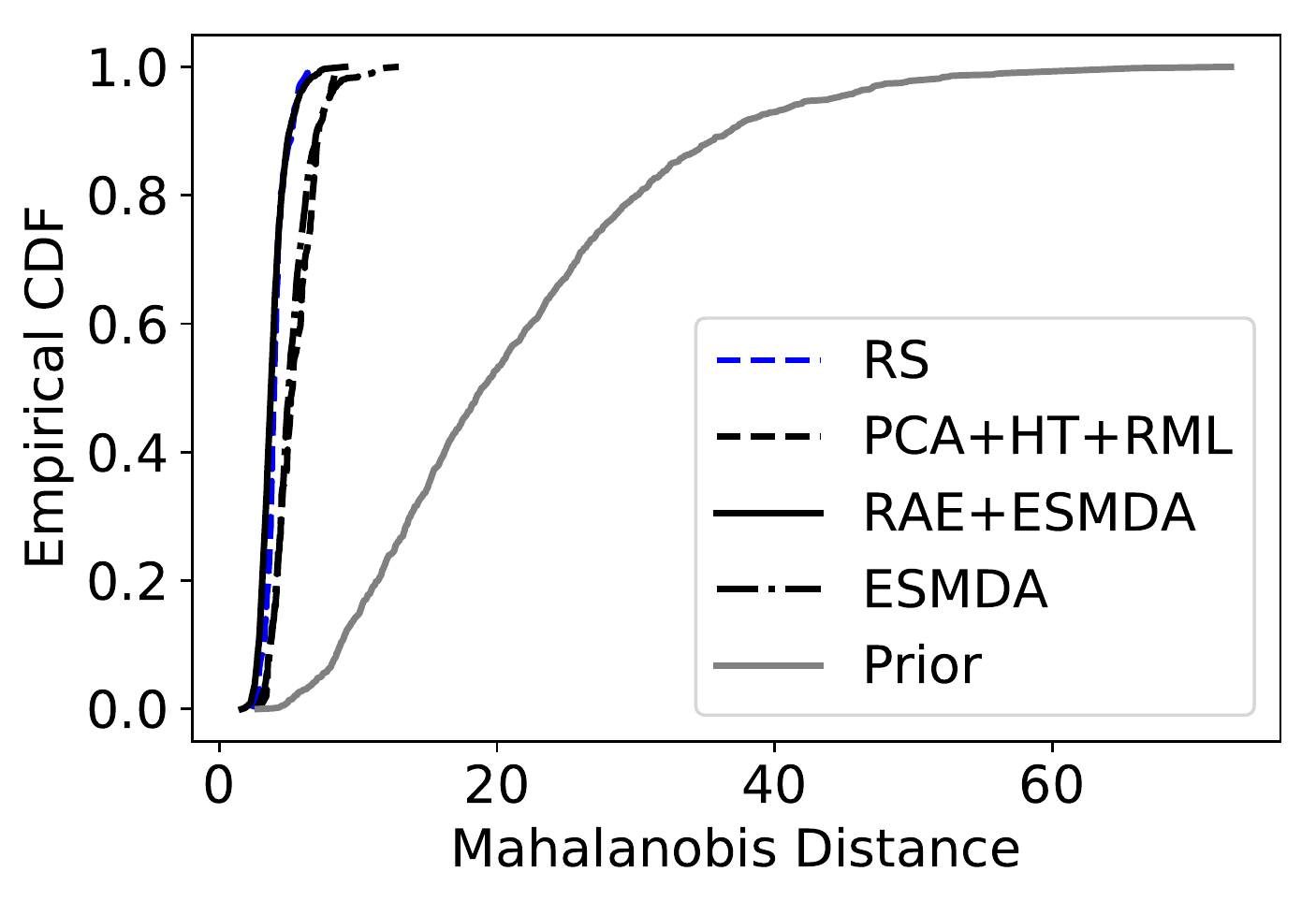}
\subcaption{True Model~A (with prior)}
\end{minipage}
\hspace{.05\linewidth}
\begin{minipage}{.4\linewidth}\centering
\includegraphics[width=\linewidth]{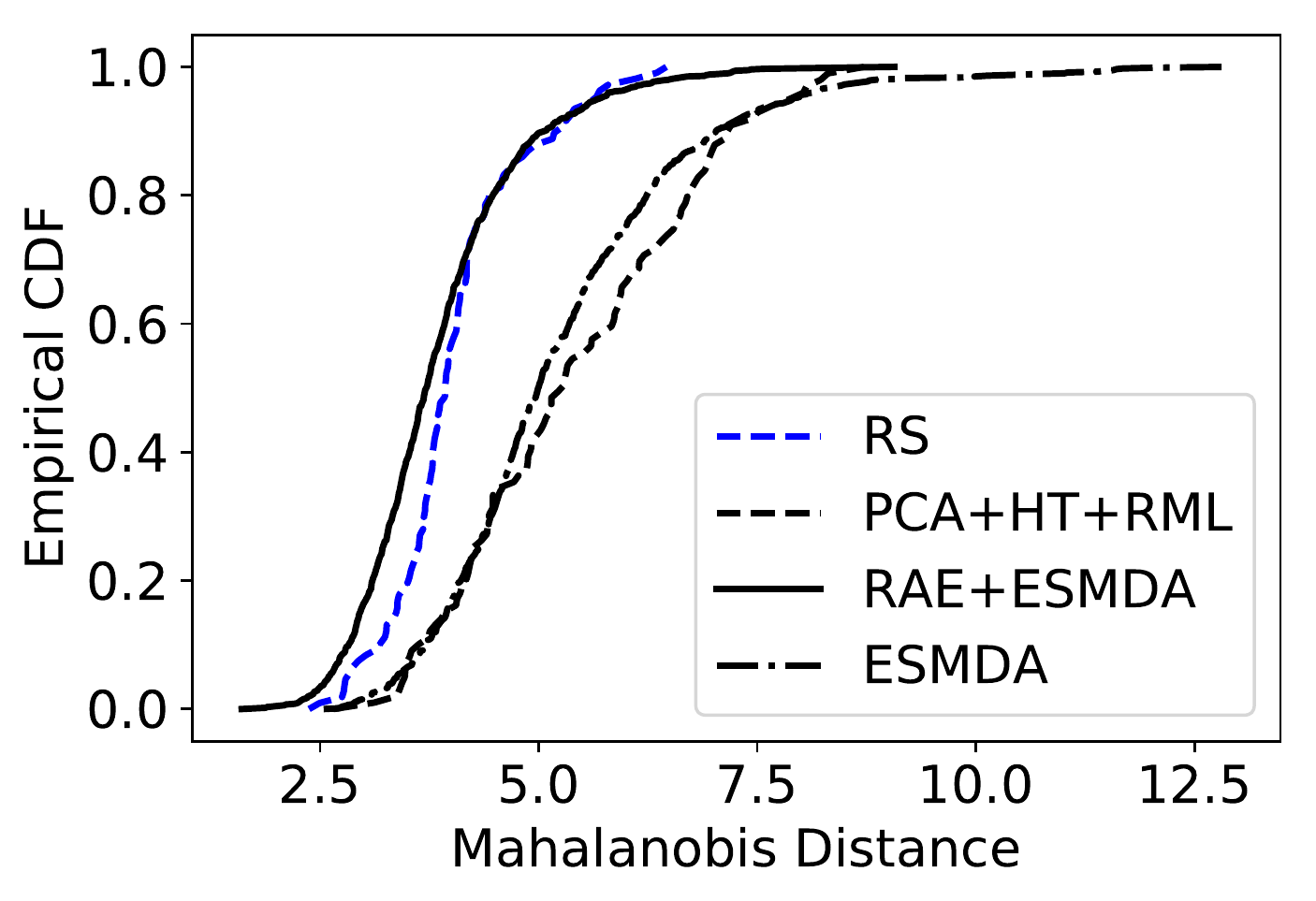}
\subcaption{True Model~A}
\end{minipage}
\begin{minipage}{.4\linewidth}\centering
\includegraphics[width=\linewidth]{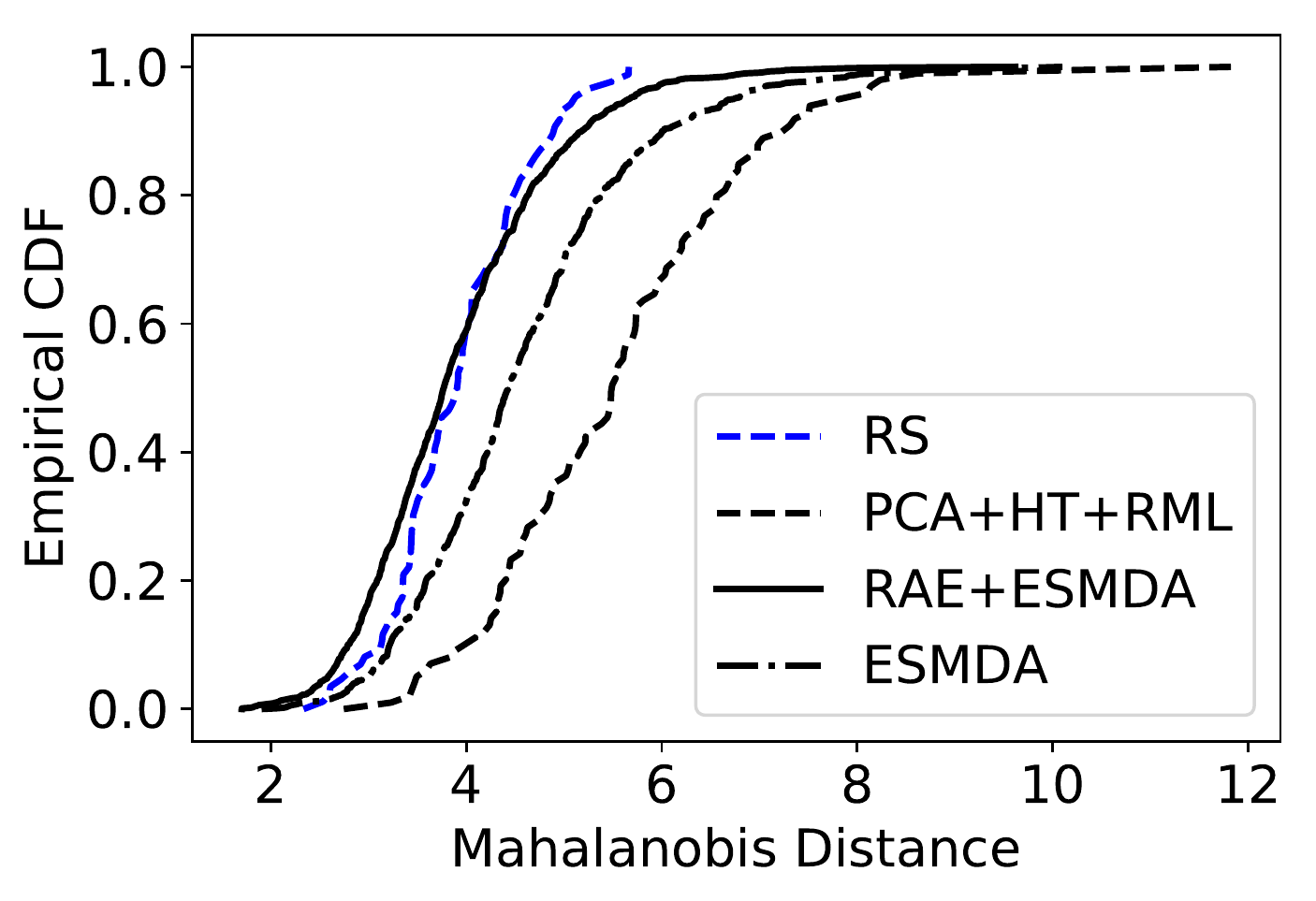}
\subcaption{True Model~B}
\end{minipage}
\hspace{.05\linewidth}
\begin{minipage}{.4\linewidth}\centering
\includegraphics[width=\linewidth]{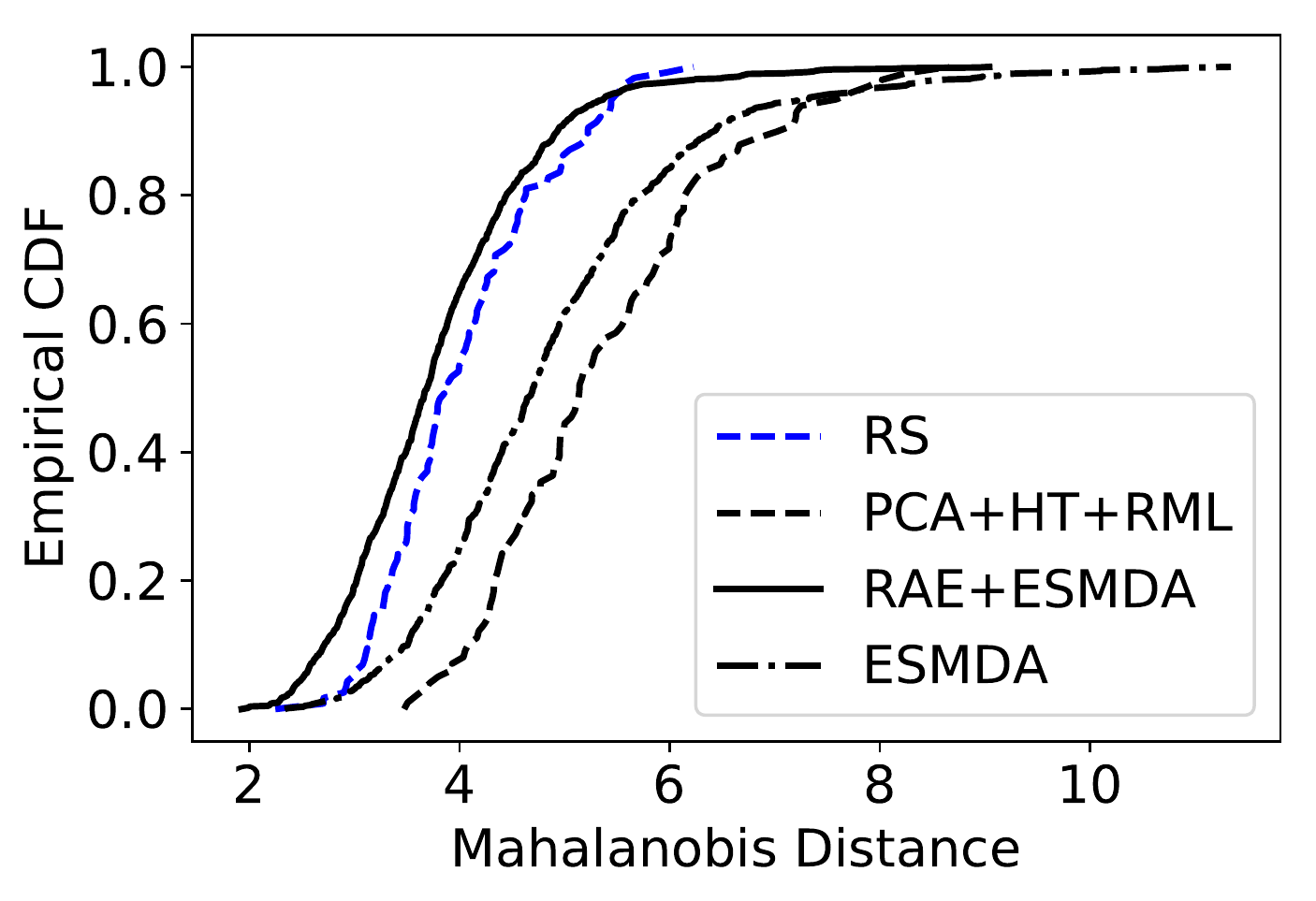}
\subcaption{True Model~C}
\end{minipage}
\caption{CDFs of Mahalanobis distance relative to RS results for True Models~A, B and C. Standalone ESMDA method includes truncation}\label{fig:dist_multiple}
\end{figure*}

Results are shown in Fig.~\ref{fig:dist_multiple} for DSI procedures based on PCA+HT+RML, ESMDA with truncation, and RAE+ ESMDA. In Fig.~\ref{fig:dist_multiple}(a) we consider True Model~A. The CDF of $D_M$ for the prior models is included in this figure. We see that the posterior CDFs from RS and all DSI treatments differ considerably from the prior CDF, and that the differences between the posterior CDFs are small compared to the difference between the prior and (any of the) posterior results. This demonstrates the impact of data assimilation. The same posterior results are displayed in Fig.~\ref{fig:dist_multiple}(b), though now the prior is not included and the $x$-axis scale is very different. Here we see that the CDF for DSI using RAE+ESMDA is quite close to that for RS, while those for the other two DSI methods show significant differences. 

True Models~B and C were newly generated and not included in the prior ensemble. RS was performed separately for each of these models, with the same type and amount of data as used for True Model~A. In the case of True Model~B, RS accepted 87 models (and corresponding data vectors), while for True Model~C, 108 models and data vectors were accepted. Posterior DSI results for True Models~B and C are shown in Fig.~\ref{fig:dist_multiple}(c) and (d). We again see that the RAE+ESMDA results are the closest to RS, and that the PCA+HT+RML and ESMDA (with truncation) results clearly differ from the RS CDFs. Although the RAE+ESMDA treatment clearly provides the best results, it is also of interest to observe that ESMDA (with truncation) outperforms PCA+HT+RML, in this metric, for True Models~B and C.

Taken in total, the results for the 2D channelized system considered here demonstrate that the DSI procedure with RAE for time-series parameterization and ESMDA for posterior sampling consistently outperforms alternate DSI treatments. This DSI methodology was shown to provide results for a range of quantities (both primary and derived) in reasonable agreement with reference RS solutions. In the next section, we consider the performance of DSI for a 3D case.

\subsection{DSI Results for a 3D Model}
\label{sec:3D_case}

In this section, we will assess two DSI formulations for a 3D oil field example that involves oil, water and gas phases. The model is exactly that used by Sun and Durlofsky~\cite{Sun2017} -- please refer to that paper for the details on the geological model and fluid properties. The anticlinal system is defined on a $31 \times 11 \times 40$ grid, with grid blocks of dimensions $50~\text{m} \times 50~\text{m} \times 2~\text{m}$. The center of the top layer is 2400~m in depth, and the flank is 2450~m in depth. The initial pressure is 234~bar.

\begin{figure*}[!htpb]
\centering
\includegraphics[width = 0.8\textwidth]{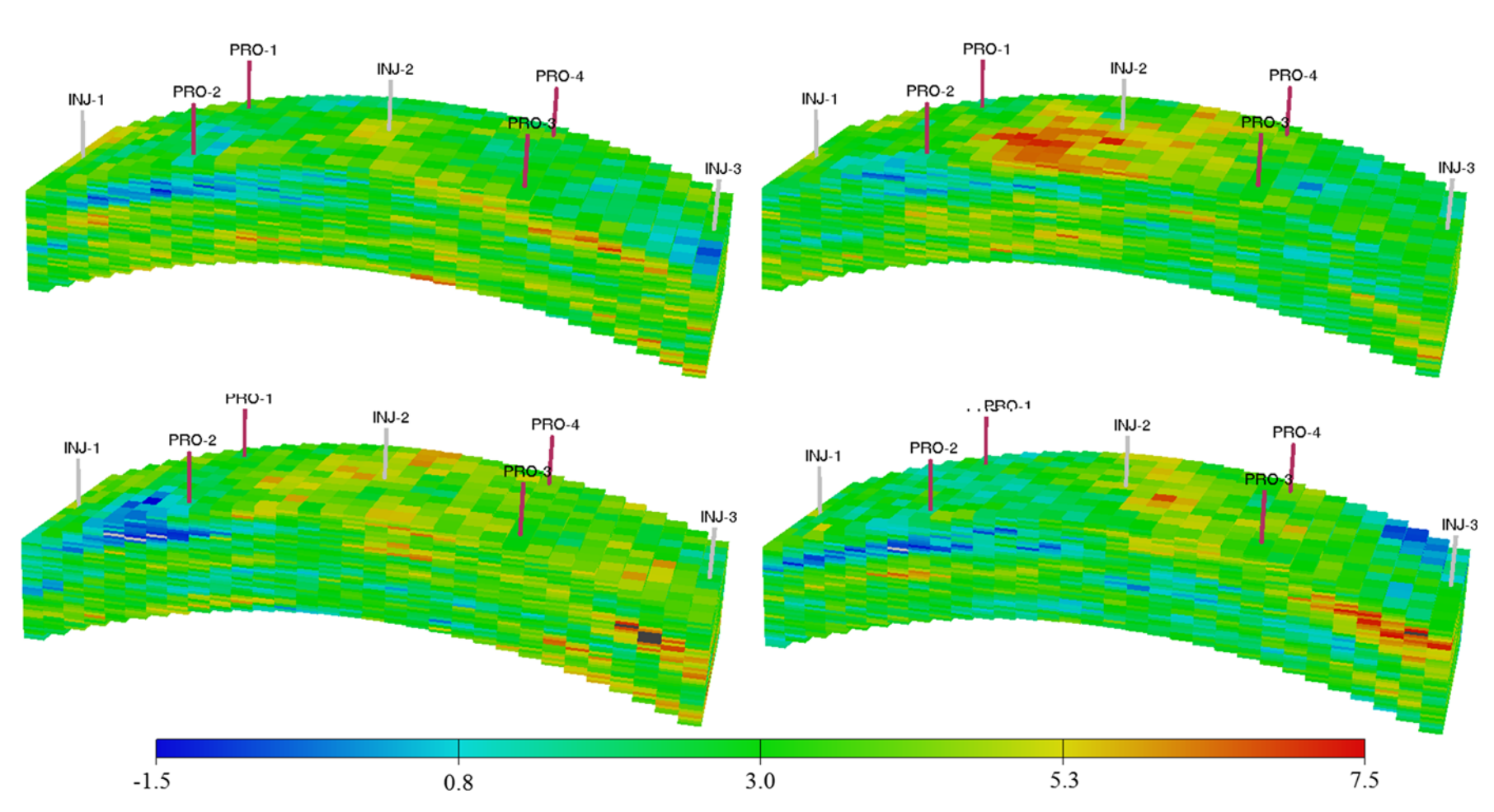}
\caption{Prior realizations of $\log k_x$ for multi-Gaussian 3D model}\label{fig:3d_perm}
\end{figure*}

Figure~\ref{fig:3d_perm} shows four prior permeability realizations. The multi-Gaussian permeability fields were conditioned to the permeability in well blocks. The model contains three injectors and four producers, with all wells penetrating all layers. The mean of the log-permeability is 3 and the standard deviation is 1.5. The components of permeability in the $x$ and $y$ directions ($k_x$ and $k_y$) are equal; permeability in $z$ is given by $k_z = 0.3 k_x$. Porosity is constant at 0.2.

The water-oil contact is 2510~m in depth. Below the contact there is only water; above the contact, the initial oil and water saturations are 0.8 and 0.2. Gas is dissolved in the oil phase and can appear as reservoir pressure decreases. All simulations were performed by Sun and Durlofsky~\cite{Sun2017}. The simulations were run for 3000~days, with data reported every 30~days (thus we have $\Nt = 100$ time steps). The simulation is divided into two periods -- the primary production period and the water injection period. The primary production period, which extends for the first 900~days, entails production without any water injection. During this period all producers operate at a fixed oil rate of 200~$\text{m}^3$/day, subject to a minimum BHP of 100~bar. Water is injected from 900~days to 3000~days. During this period all injectors operate at a constant BHP of 500~bar and the producers operate at an oil rate of 250~$\text{m}^3$/day. The minimum BHP for producers remains at 100~bar. 

In this case, the data in $\Bd$ include the injection rates of the three injectors and oil production rate (OPR), water production rate (WPR), and BHP for the four producers (thus we have $\Nqoi = 15$ and $\Bd \in \R^{1500 \times 1}$). In order to enable comparisons with RS we again consider a small amount of data. Specifically, we use observed data at five wells (I1, I2, P1, P2 and P3) at only a single time step -- 1440~days, which is during the water injection period. The measurement errors again follow Gaussian distributions with mean value zero and standard deviation of $10\%$ of the simulated true data. We use 500 prior realizations for DSI. A total of 500,000 models were simulated to provide RS samples. Of these, 102 samples were accepted in the RS process.

\begin{figure*}[!ht]
\centering
\begin{minipage}{.4\linewidth}\centering
\includegraphics[width=\linewidth]{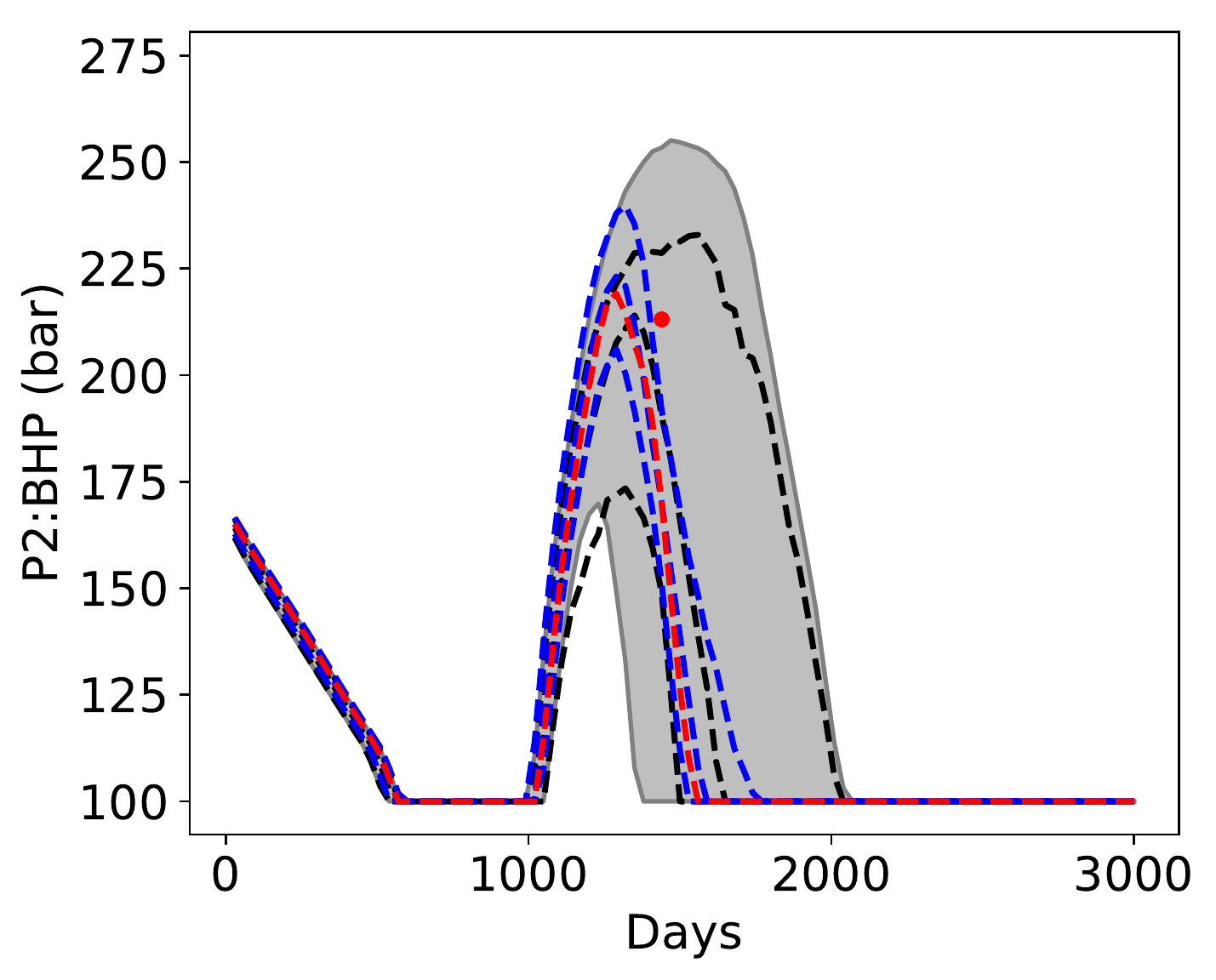}
\subcaption{P2: BHP (PCA+HT+RML)}
\end{minipage}
\hspace{.05\linewidth}
\begin{minipage}{.4\linewidth}\centering
\includegraphics[width=\linewidth]{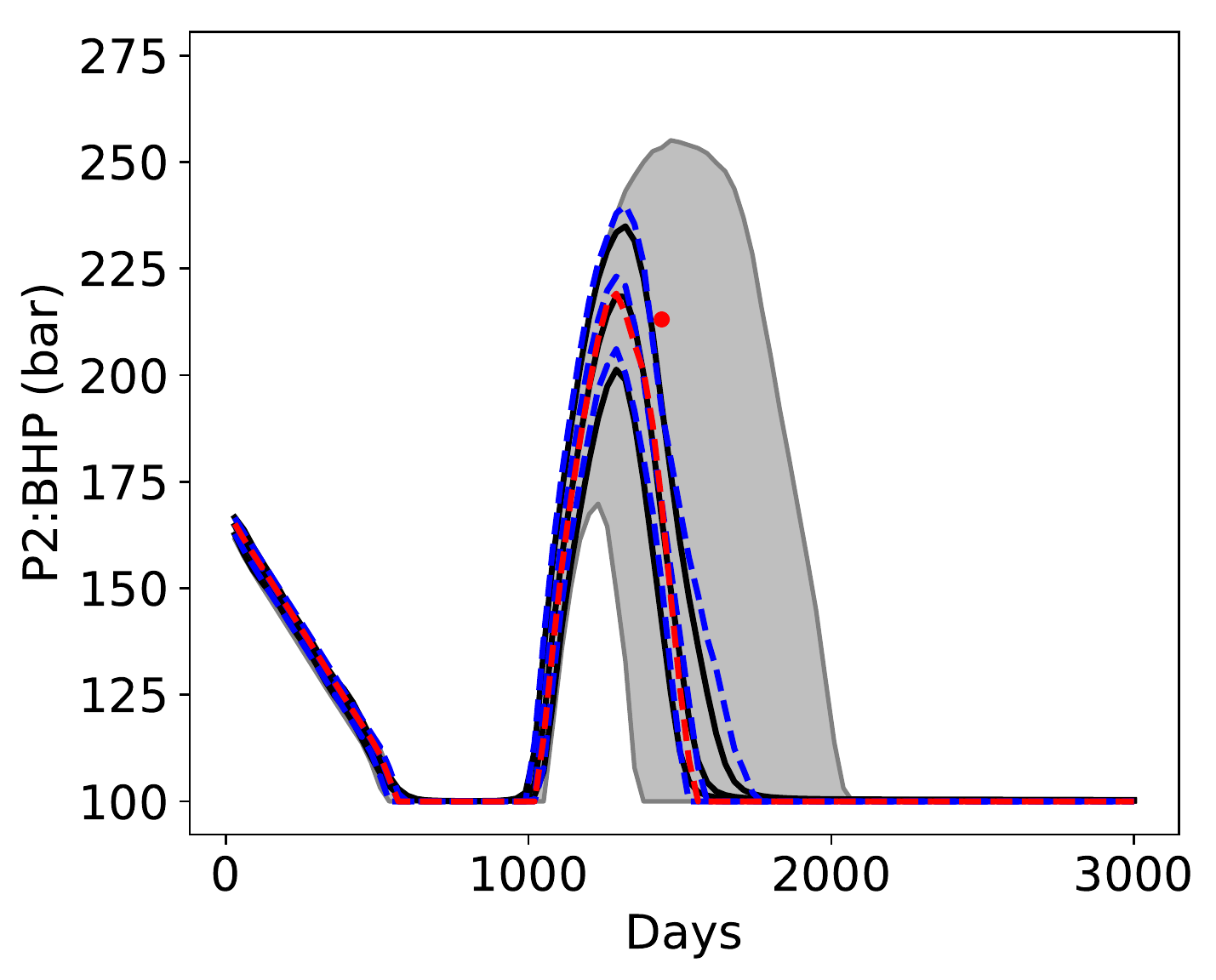}
\subcaption{P2: BHP (RAE+ESMDA)}
\end{minipage}
\begin{minipage}{.4\linewidth}\centering
\includegraphics[width=\linewidth]{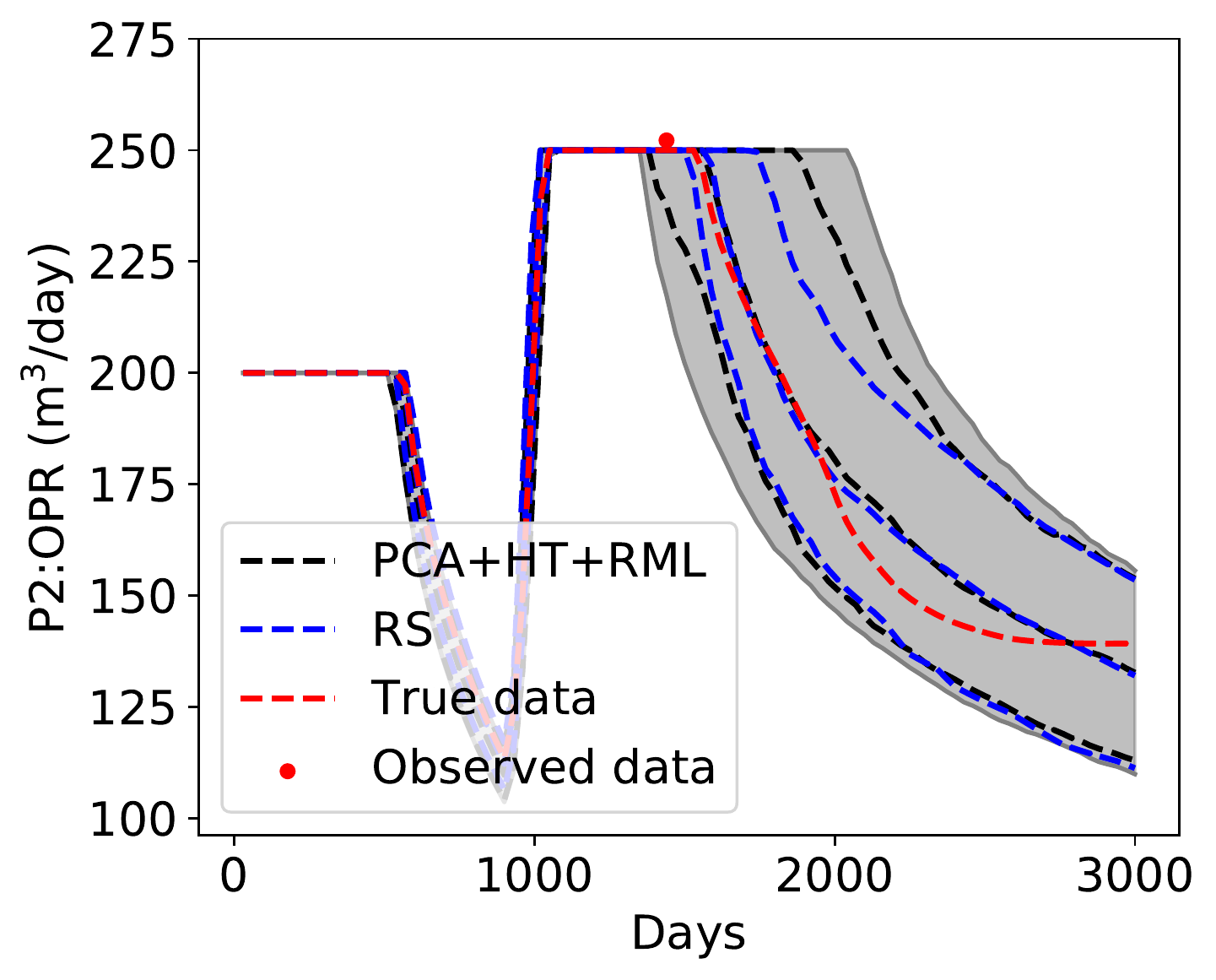}
\subcaption{P2: OPR (PCA+HT+RML)}
\end{minipage}
\hspace{.05\linewidth}
\begin{minipage}{.4\linewidth}\centering
\includegraphics[width=\linewidth]{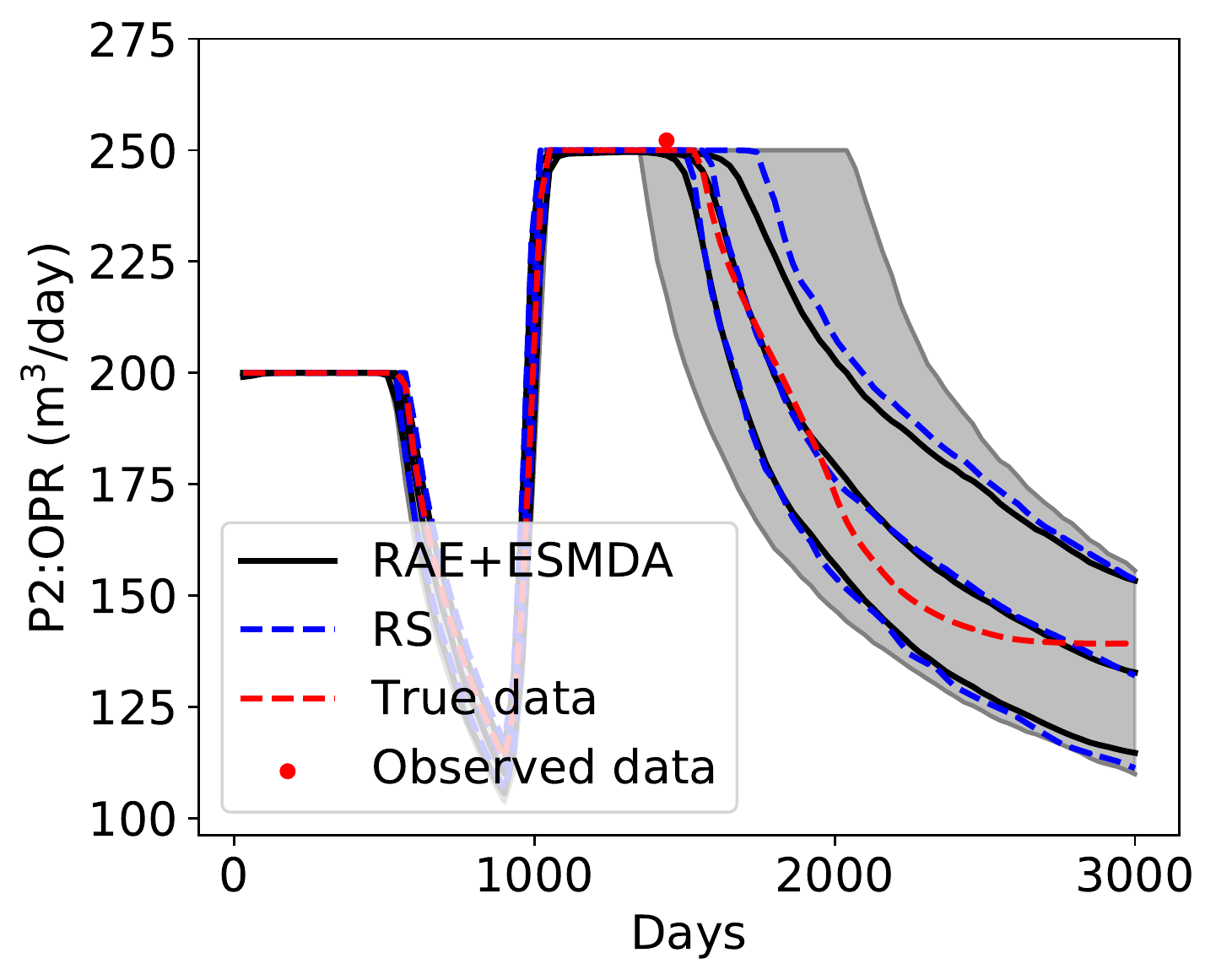}
\subcaption{P2: OPR (RAE+ESMDA)}
\end{minipage}
\caption{PCA+HT+RML and RAE+ESMDA posterior results along with reference RS results. Gray regions indicate P$_{10}$-P$_{90}$ prior uncertainty range} \label{fig:post_3d_pca_rae}
\end{figure*}

We now present posterior DSI results using the PCA+HT +RML and RAE+ESMDA treatments. In both cases we use $\Nl=15$, and a total of 500 posterior predictions are constructed. Results for BHP and oil production rate for well P2 are shown in Fig.~\ref{fig:post_3d_pca_rae}. The results in Fig.~\ref{fig:post_3d_pca_rae}(a) and (c) use PCA+HT+RML, while those in Fig.~\ref{fig:post_3d_pca_rae}(b) and (d) apply RAE+ESMDA. Note that the P$_{10}$-P$_{90}$ prior uncertainty range is also shown on all plots.

In this case, because we have only one observation for each data type, the prediction accuracy depends on properly capturing the correlations between different variables. In Fig.~\ref{fig:post_3d_pca_rae}(a), we see that the PCA+HT+RML DSI results display error in the P$_{10}$ and P$_{90}$ predictions compared to the reference RS solution. In fact, the P$_{10}$ and P$_{90}$ curves show relatively little uncertainty reduction relative to the prior, in contrast to the RS results. The RAE+ESMDA predictions in Fig.~\ref{fig:post_3d_pca_rae}(b), by contrast, show close agreement with RS. Accuracy is observed in the P$_{10}$, P$_{50}$ and P$_{90}$ curves, and the appropriate degree of uncertainty reduction is achieved in this case. 

Enhanced accuracy using the RAE+ESMDA treatment is also apparent in the results for P2 OPR in Fig.~\ref{fig:post_3d_pca_rae}(c) and (d). The PCA+HT+RML DSI results in Fig.~\ref{fig:post_3d_pca_rae}(c) are reasonably accurate, but some error is evident in the P$_{10}$ and P$_{90}$ curves around the time that they decrease from the production plateau. For the RAE+ESMDA results in Fig.~\ref{fig:post_3d_pca_rae}(d), the P$_{10}$ and P$_{50}$ curves track the RS results closely. There is still some error in the P$_{90}$ prediction, but it is less than that in the PCA+HT+RML results.

\begin{figure*}[!ht]
\centering
\begin{minipage}{.4\linewidth}\centering
\includegraphics[width=\linewidth]{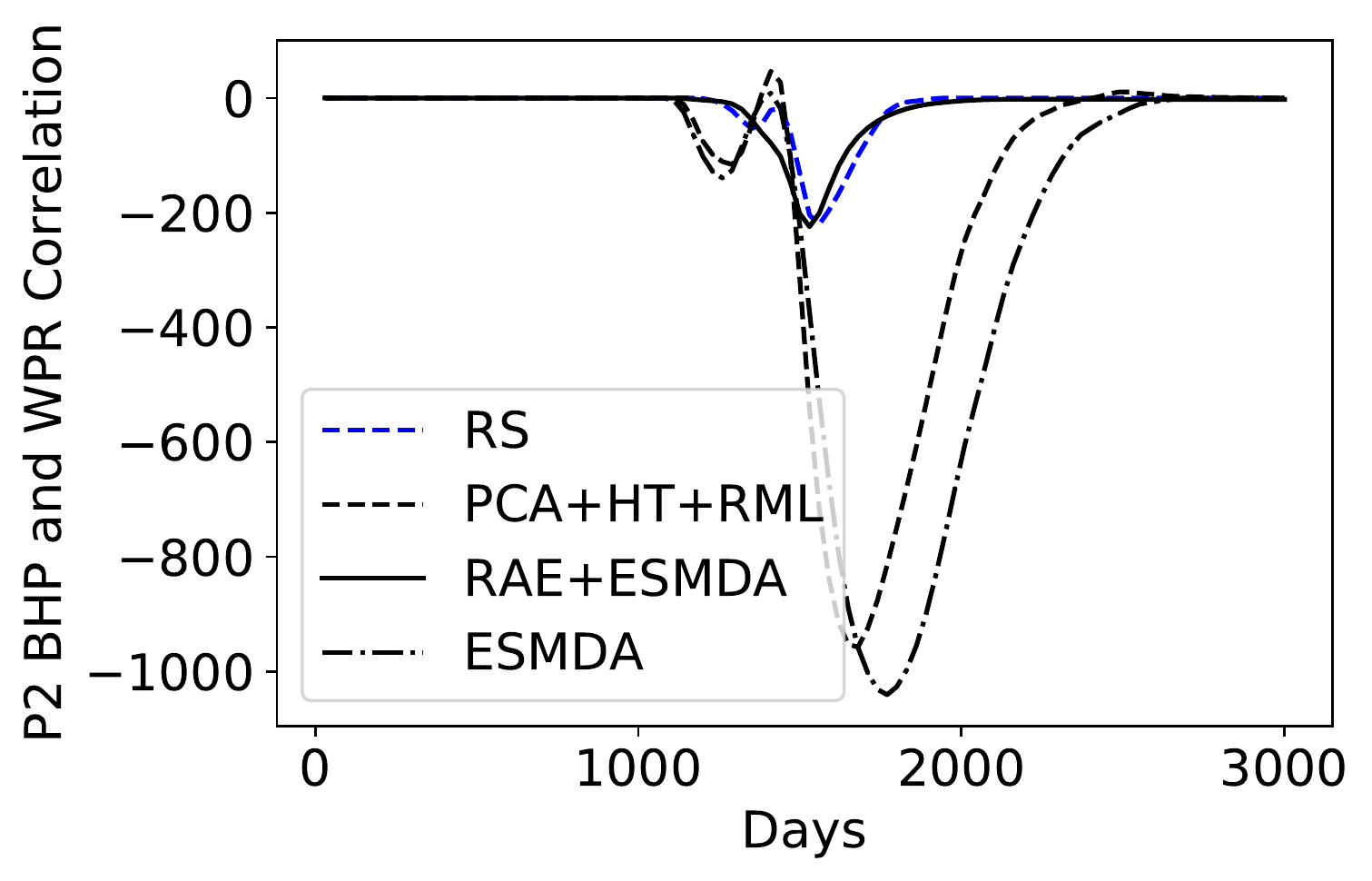}
\subcaption{Covariance for P2 BHP and WPR}
\end{minipage}
\hspace{.05\linewidth}
\begin{minipage}{.4\linewidth}\centering
\includegraphics[width=\linewidth]{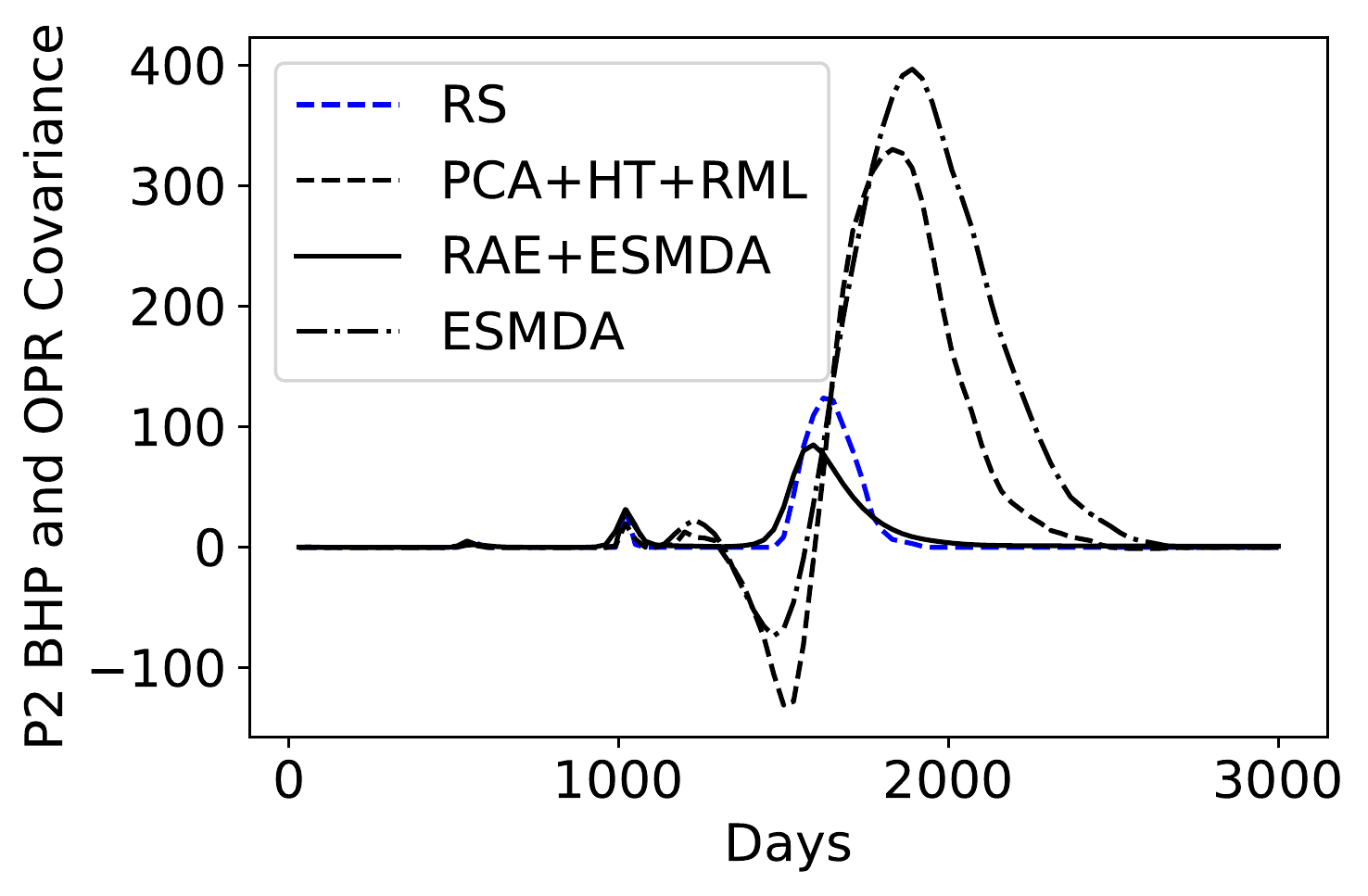}
\subcaption{Covariance for P2 BHP and OPR}
\end{minipage}
\begin{minipage}{.4\linewidth}\centering
\includegraphics[width=\linewidth]{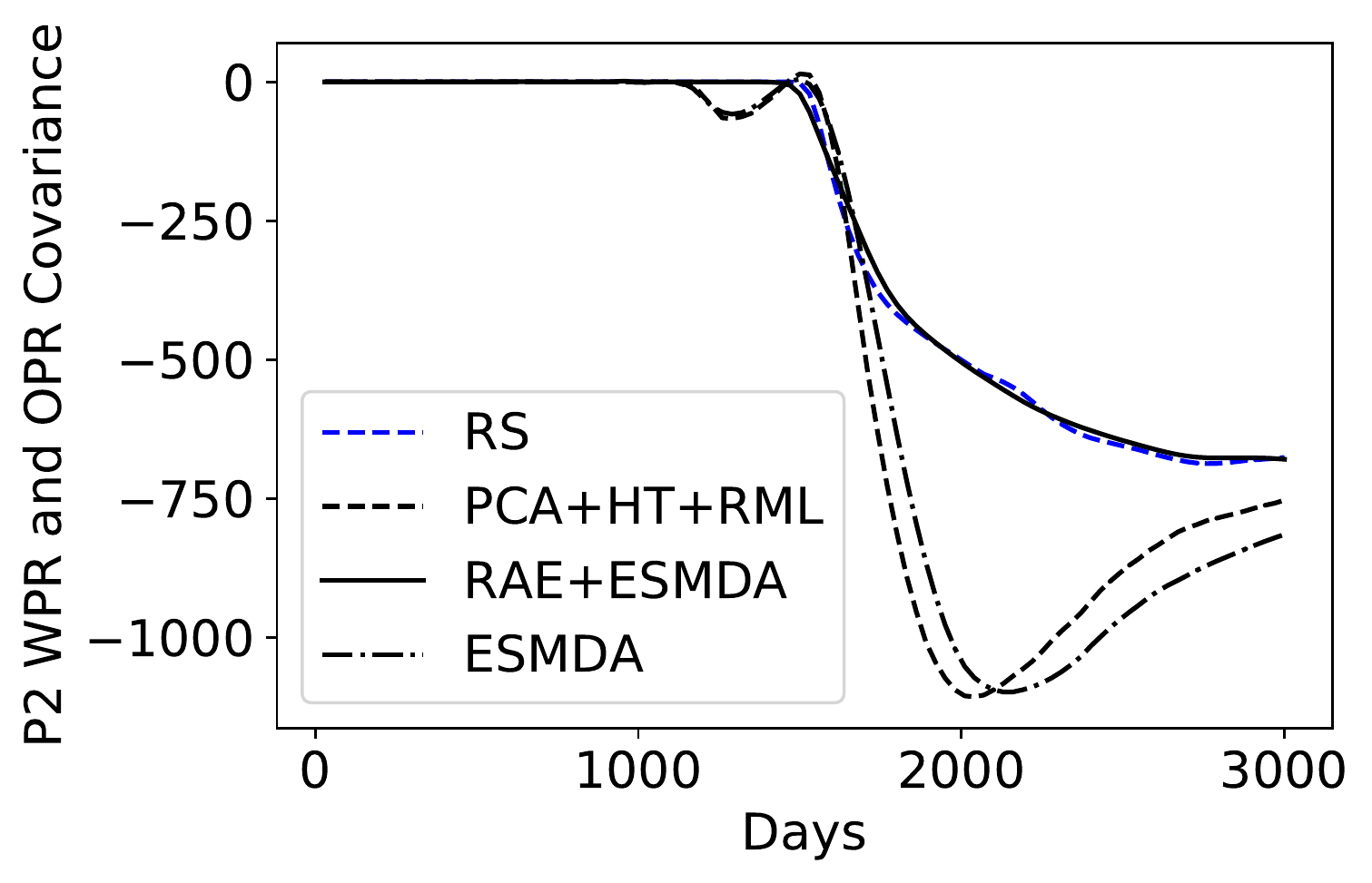}
\subcaption{Covariance for P2 WPR and OPR}
\end{minipage}
\caption{Covariance for P2 WPR, OPR and BHP for the full period. Standalone ESMDA method includes truncation}\label{fig:corr_post_3d}
\end{figure*}

As noted above, because we use a small amount of data in this case, reliable posterior predictions require that correlations be captured accurately. In
Fig.~\ref{fig:corr_post_3d}, we display the covariance between three production quantities -- WPR, OPR and BHP -- for well P2 over the full simulation time frame. Results for PCA+HT+ESMDA, ESMDA with truncation, and RAE+ESMDA are compared to those from RS. We see that the covariance for DSI using RAE+ESMDA is consistently accurate, while that for DSI using PCA+HT+ESMDA and ESMDA with truncation display large discrepancies. These inaccuracies in covariance result in less accurate posterior predictions for primary quantities in this case, consistent with the results in Fig.~\ref{fig:post_3d_pca_rae}. Thus this example clearly highlights the importance of accurately representing covariance in the DSI parameterization in cases with small amounts of observed data.

\section{Concluding Remarks}\label{sec:conclusion}

In this paper, we introduced a new data parameterization method based on a recurrent autoencoder (RAE), and implemented this procedure into the data-space inversion methodology. DSI operates within a Bayesian framework and provides posterior data variables/vectors (but not posterior models). The RAE-based treatment enables the parameterization of DSI data vectors, which include time-series data corresponding to the phase flow rates at injection and production wells. The RAE consists of an autoencoder to perform the low-dimensional parameterization and LSTM networks to capture the correlations in the time-series data. Predictions are generated from the decoder with stacked LSTM layers. The RAE parameterization represents a nonlinear generalization of PCA-based methods. As such, it mitigates the unphysical behavior that can result from the use of PCA combined with histogram transformation (HT). Within the DSI framework, an ensemble smoother with multiple data assimilation (ESMDA) was used to efficiently generate posterior samples of data vectors conditioned to observations. This sampling procedure was shown to be compatible with the new RAE-based parameterization. 

We evaluated the performance of DSI with RAE and ESMDA, along with other DSI variants, for a bimodal 2D channelized geological system and a 3D multi-Gaussian system. These problems involved two-phase and three-phase flow, respectively. DSI posterior predictions were compared, in all cases, to reference (computationally expensive) rejection sampling results. DSI performance using RAE and ESMDA was shown to be consistently better than that from the other treatments, which included the use of PCA and HT, and the application of standalone ESMDA with truncation. More specifically, DSI with RAE and ESMDA consistently outperformed the other methods in terms of the accuracy of P$_{10}$, P$_{50}$, and P$_{90}$ posterior predictions for primary and derived data variables, and for correlation and covariance between data variables, for both the 2D and 3D cases. For the 2D channelized case, we considered two additional `true' models. Performance was assessed for these cases by comparing the Mahalanobis distance CDFs for the DSI methods to that from RS. For this metric, we found DSI with RAE and ESMDA to again provide the best accuracy for the three `true' cases considered.

There are a number of topics in this general area that should be addressed in future work. Our current RAE method requires $O(500-1000)$ training data realizations for cases with $O(1000-2000)$ data variables. Additional treatments or enhancements may be required to handle cases with significantly more data (as may occur with larger numbers of wells). It will also be of interest to evaluate the performance of RAE with data variables from different geological scenarios. The parameterization of data with a large amount of noise or abrupt variations (due to, e.g., wells opening and closing), which is often observed in practice, should be considered. The use of more flexible network architectures (such as 1D CNNs) in the autoencoder may enable the treatment of different types of data variables with different time spans, which would extend the applicability of the DSI methodology. Finally, in common with all inversion methodologies, DSI relies on the appropriateness of the prior ensemble of geomodels and corresponding data vectors. Because these models are, in practice, imperfect, it will be very useful to incorporate model-error effects into the DSI framework. Work along these lines is currently underway.

\begin{acknowledgements}
We thank Chevron ETC and the industrial affiliates of the Stanford Smart Fields Consortium for financial support. We are grateful to Wenyue Sun for providing the prior geological models and flow simulation results used in this work, and to Yimin Liu and Robin Hui for useful discussions and suggestions.
\end{acknowledgements}

\bibliographystyle{spmpscit}

\end{document}